\documentclass[11pt, letterpaper]{arxiv}
\usepackage[all]{hypcap}
\usepackage[comma,numbers,sort,compress]{natbib}
\usepackage{hyperref}[citecolor=magenta]
\usepackage[capitalize,noabbrev]{cleveref}
\hypersetup{
    colorlinks = true,
    citecolor = {magenta},
}

\usepackage{microtype}
\usepackage{graphicx}
\usepackage{subfigure}
\usepackage{booktabs}

\usepackage{xcolor}
\usepackage{colortbl}
\usepackage{algorithm}
\usepackage{algpseudocode}
\usepackage{setspace}

\usepackage{amsmath}
\usepackage{amssymb}
\usepackage{mathtools}
\usepackage{amsthm}
\usepackage{bbm}
\usepackage{multirow}

\usepackage{placeins}

\usepackage{fleqn, tabularx}

\usepackage{float}

\usepackage[most,skins,theorems]{tcolorbox}
\tcbset{
  aibox/.style={
    width=\linewidth,
    top=7pt,
    bottom=2pt,
    colback=blue!6!white,
    colframe=black,
    colbacktitle=black,
    enhanced,
    center,
    attach boxed title to top left={yshift=-0.1in,xshift=0.15in},
    boxed title style={boxrule=0pt,colframe=white,},
  }
}
\newtcolorbox{AIbox}[2][]{aibox,title=#2,#1}

\usepackage{wrapfig}
\definecolor{lightblue}{rgb}{0.22,0.45,0.70}
\usepackage{fleqn, tabularx}

\definecolor{rliableolive}{HTML}{BBCC33}
\definecolor{rliableblue}{HTML}{77AADD}
\definecolor{rliablered}{HTML}{EE8866}

\usepackage{crossreftools}
\usepackage[suppress]{color-edits}

\usepackage{dsfont}
\usepackage{nicefrac}
\newtcolorbox{analysisbox}[1][]{
    enhanced jigsaw,
    colback=white,
    colframe=blue!75!black,
    fonttitle=\bfseries,
    boxsep=5pt,
    left=5pt,
    right=5pt,
    top=5pt,
    bottom=5pt,
    title=#1,
}
\definecolor{editInitialResponse}{RGB}{255, 235, 156} 
\definecolor{editBacktrack}{RGB}{0, 0, 139}
\definecolor{editRevisedResponse}{RGB}{255, 182, 193}

\usepackage{pifont}
\usepackage{soul}
\definecolor{highlightmistake}{RGB}{255, 179, 179} 
\definecolor{highlightcorrect}{RGB}{179, 255, 179}

\usepackage[capitalize,noabbrev]{cleveref}

\theoremstyle{plain}
\newtheorem{theorem}{Theorem}[section]

\theoremstyle{definition}
\newtheorem{definition}[theorem]{Definition}
\newtheorem{assumption}[theorem]{Assumption}
\theoremstyle{remark}

\usepackage[textsize=tiny]{todonotes}

\usepackage{enumitem}

\usepackage{amsmath,amsfonts,bm}

\def\eqref#1{Eq.~\ref{#1}}

\def\1{\bm{1}}

\DeclareMathAlphabet{\mathsfit}{\encodingdefault}{\sfdefault}{m}{sl}
\SetMathAlphabet{\mathsfit}{bold}{\encodingdefault}{\sfdefault}{bx}{n}

\title{On the Universality of Self-Supervised Learning}

\author[1,2]{Wenwen Qiang}
\author[1,2]{Jingyao Wang}
\author[1,2]{Changwen Zheng}
\author[3]{Hui Xiong}
\author[4]{Gang Hua}

\affil[1]{University of Chinese Academy of Sciences}
\affil[2]{Institute of Software Chinese Academy of Sciences}
\affil[3]{Hong Kong University of Science and Technology}
\affil[4]{Amazon.com, Inc., Bellevue, WA}

\begin{document}

\maketitle

\textbf{Abstract:} 
In this paper, we investigate what constitutes a good representation or model in self-supervised learning (SSL). We argue that a good representation should exhibit universality, characterized by three essential properties: discriminability, generalizability, and transferability. While these capabilities are implicitly desired in most SSL frameworks, existing methods lack an explicit modeling of universality, and its theoretical foundations remain underexplored. To address these gaps, we propose General SSL (GeSSL), a novel framework that explicitly models universality from three complementary dimensions: the optimization objective, the parameter update mechanism, and the learning paradigm. GeSSL integrates a bi-level optimization structure that jointly models task-specific adaptation and cross-task consistency, thereby capturing all three aspects of universality within a unified SSL objective. Furthermore, we derive a theoretical generalization bound, ensuring that the optimization process of GeSSL consistently leads to representations that generalize well to unseen tasks. Empirical results on multiple benchmark datasets demonstrate that GeSSL consistently achieves superior performance across diverse downstream tasks, validating its effectiveness in modeling universal representations.

\section{Introduction}
\label{sec:1}

Self-supervised learning (SSL) has revolutionized machine learning by enabling models to learn meaningful representations from unlabeled data, thereby significantly reducing reliance on large labeled datasets \citep{gui2024survey}. SSL methods are generally divided into two categories: discriminative SSL (D-SSL) and generative SSL (G-SSL). D-SSL approaches, such as SimCLR \citep{simclr}, BYOL \citep{byol}, and Barlow Twins \citep{barlowtwins}, focus on distinguishing between different augmented views of the same image, learning representations by maximizing the similarity between positive pairs and minimizing it with negative ones. In contrast, G-SSL methods like MAE \citep{mae} aim to reconstruct missing or corrupted parts of the input data, learning representations by capturing inherent visual structures and patterns. Both D-SSL and G-SSL have demonstrated remarkable ability in representation learning.

Whether using D-SSL or G-SSL methods, most researches focus on determining which factors, e.g., network architectures \citep{dino}, optimization strategies \citep{ni2021close}, prior assumptions \citep{wmse}, inductive biases \citep{byol}, etc., lead to good representations or models. However, a fundamental question persists: Why these factors can lead to a ``good'' representation or model? To address this question, the common practice is to evaluate the learned representations or models on various downstream tasks, that is, if the performance is strong, the representation or model is deemed good. Yet, a key challenge remains in understanding the underlying mechanisms by which these factors yield a good representation or model. In other words, we often lack direct explanations of how specific methodological choices influence the quality of the representation or model. For instance, why does an asymmetric dual-branch network architecture in methods like BYOL enhance performance on downstream tasks? Similarly, why does enforcing a uniform distribution on feature representations serve as an inductive bias for obtaining good representations in methods like SimCLR?

In this paper, we shift focus from designing SSL methods in terms of ``which factors should be adopted'' to exploring ``what directly constitutes a good representation or model''. The advantage of this shift is that, once we have identified the key properties that define a good representation, we can directly incorporate these properties into the optimization objective of SSL. As a result, we no longer need to justify whether a particular “do” operation can implicitly lead to good representations, because the representation itself is explicitly modeled as the target of learning. Thus, we concentrate on the question: What characteristics should a good representation or model possess? Inspired by the evaluation methods of most SSL and unsupervised learning approaches \citep{simclr, byol, mae}, we answer this question by that a good representation or model should satisfy three constraints: 1) Discriminability: For a single task, the model should achieve the expected performance on the training set; 2) Generalizability: For a single task, the trained model should generalize to unseen datasets while maintaining its performance; 3) Transferability: The trained model should generalize to multiple different tasks while guaranteeing its performance. We next consolidate the three dimensions, e.g., discriminability, generalization, and transferability, into a single criterion: universality. When these capabilities are jointly satisfied within one framework, the resulting representation or model is said to possess high universality. Hence, a ``good'' representation or model can be succinctly defined as one with high universality: it separates classes effectively on the current task, generalizes robustly to unseen yet in‑distribution data, and transfers efficiently to unfamiliar scenarios, thereby furnishing a reliable, reusable knowledge foundation for diverse downstream objectives.

Given the definition of Universality, a central challenge is how to formally integrate its properties into the SSL process. To address this, we propose General SSL (GeSSL), a unified framework that explicitly models Universality by embedding its three core components, e.g., discriminability, generalizability, and transferability, into SSL training. For discriminability, GeSSL not only leverages alignment-based objectives commonly used in existing SSL methods but also introduces an additional discriminative loss to improve class separation in an unsupervised setting. For generalizability, GeSSL employs a bi-level optimization mechanism that separates training data into support and query sets, thereby simulating an update-then-evaluate process to directly model generalization behavior. Moreover, it encourages shared feature extractor across multiple tasks, which indirectly enhances generalization by promoting causal consistency. Lastly, for transferability, GeSSL adopts an episodic training paradigm in which multiple mini-batches are treated as distinct tasks, allowing the model to estimate and adapt to the underlying task distribution and generalize to unseen scenarios. In this way, GeSSL provides a principled approach to modeling Universality within the SSL framework. To further establish the soundness of this framework, we provide formal performance guarantees showing that GeSSL’s training objective leads to bounded generalization error on novel tasks. This is achieved under smoothness and boundedness assumptions, demonstrating that the jointly optimized representation can be reliably adapted to unseen tasks with good predictive performance.

\textbf{Our contributions}: \textbf{(i)} We theoretically define SSL universality, encompassing discriminability, generalizability, and transferability (Sections \ref{sec:4.1}). \textbf{(ii)} We propose GeSSL, a novel framework that models universality through a bi-level learning paradigm (Section \ref{sec:4.3}). \textbf{(iii)} Theoretical and empirical evaluations on benchmark datasets demonstrate the advantages of GeSSL (Sections \ref{sec:5}, \ref{sec:6}).

\section{Revisiting SSL from a Task Perspective}
\label{CauDC:ulf}

During the training phase, the data is organized into mini-batches, i.e., a mini-batch is denoted as ${X_{tr}} =  \{{{x_i}} \}_{i = 1}^{N}$, where $x_i$ is the $i$-th sample, and $N$ is the batch size. In D-SSL, each sample $x_i$ undergoes stochastic data augmentation to generate two augmented views, i.e., $x^1_i$ and $x^2_i$. In G-SSL, each sample $x_i$ is partitioned into multiple small blocks, some blocks are masked, and the remaining blocks are reassembled into a new sample $x^1_i$. The original sample is then referred to as $x^2_i$. We can also regard the partition-reassemble operation as augmentation. Consequently, each augmented dataset in both D-SSL and G-SSL is represented as $X_{tr}^{aug} = \{ {x_i^1,x_i^2} \}_{i=1}^{N}$. Each $\{ {x_i^1,x_i^2} \}$ constitutes the $i$-th sample pair, and the SSL objective is to learn a feature extractor $f$ from these pairs.

D-SSL methods typically have two main objectives: alignment and regularization \citep{simclr,oord2018representation,hjelm2018learning}. The alignment objective maximizes the similarity between paired samples in the embedding space, while the regularization objective constrains the learning behavior via inductive biases. For example, SimCLR \citep{simclr} enforces a uniform distribution over the feature representations. G-SSL methods \citep{mae} can also be viewed as implementing alignment within a pair using an encoding-decoding structure: sample $x^1_i$ is input into this structure to generate an output that is made as consistent as possible with sample $x^2_i$. Notably, alignment in D-SSL is often implemented using anchor points, where one sample in a pair is viewed as the anchor, and the training process gradually pulls the other sample towards this anchor. This concept of an anchor is also applicable to G-SSL, where $x^2_i$ is treated as the anchor, and the training process involves constraining $x^1_i$ to approach $x^2_i$.

Regardless of whether it is G-SSL or D-SSL, the anchor can be regarded as a learning target. Specifically, SSL can be interpreted as follows: In a data augmentation pair, one sample (the anchor) is designated as the target. By constraining the other augmented sample in the feature space to move toward this anchor, consistency in feature representations is achieved. This dynamic adjustment causes samples within the same pair to become tightly clustered, thus, the anchor plays a role similar to that of a clustering center. In other words, for a mini-batch $X_{tr}^{aug}$ in SSL, each pair within the batch can be considered as belonging to a specific class, where the class center serves as the anchor. Thus, $X_{tr}^{aug}$ can be interpreted as a multi-class classification task with $N$ classes. Given the role of the ``alignment part'' in SSL, the learning process within a single mini-batch can be viewed as performing a classification task. More details about SSL task construction are provided in Appendix \ref{sec:app_G.5}.

\section{Methodology}
\label{sec:4}

\begin{figure*}
    \centering
    \includegraphics[width=\textwidth]{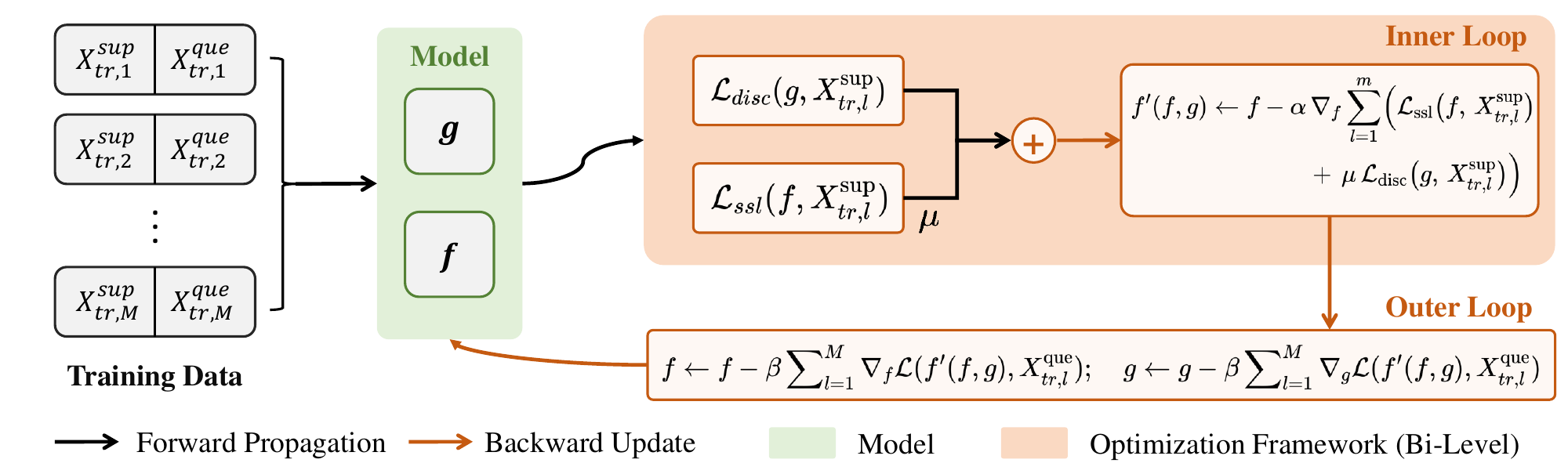}
    \vspace{-0.15in}
    \caption{Overview of GeSSL. The meaning of the different components is marked below the figure.}
    \label{fig:GeSSL}
\end{figure*}

In this section, we first present the definition and explanation of universality. Then, we propose a way to model universality in SSL, named General SSL (GeSSL). The framework of GeSSL is illustrated in Figure \ref{fig:GeSSL}. Finally, we give some high-level explanations for the proposed modeling method.

\subsection{Definition and Explanation of Universality}
\label{sec:4.1}

Typically, a learnable task consists of a training dataset and a test dataset, where the training dataset is used to train the model and the test dataset is used to evaluate its performance. Each element in the training or test dataset is usually represented as a tuple consisting of an input sample and its corresponding label. If we treat a task as a basic unit, then each task can be viewed as being sampled from a task distribution $P_{t}$. Based on this, we present the definition of universality as follows:

\begin{definition}[Universality]\label{definition:1}
For a set of training tasks and a disjoint set of test tasks, i.e., with no class-level overlap and each sample is with a label, the model $f_{\theta}$, or the representation extracted by it, is said to exhibit universality if it satisfies: \textbf{1) Discriminability}: For a training task with labeled training dataset, a model $f_{\theta}$ trained on the training dataset can predict the labels of all training samples with high accuracy. \textbf{2) Generalizability}: For a training task with training and test datasets, a model $f_{\theta}$ trained on the training dataset can predict the labels of all test samples with high accuracy. \textbf{3) Transferability}: For a training task and a test task, a model $f_{\theta}$ trained on the training dataset of the training task can predict the labels of all samples of the test task with high accuracy.
\end{definition}

Discriminability, Generalizability, and Transferability are not new concepts. At first glance, Universality may seem like a simple combination of these three and therefore lacks novelty. However, the core purpose of proposing \textbf{Universality} is to more clearly answer a fundamental question: what ``directly'' constitutes a good representation or model. When we revisit SSL from this perspective, the value of \textbf{Universality} becomes evident, because it directly quantifies the essential qualities that a good representation should possess, thus forming a sharp contrast with the motivations of existing SSL methods. We next explain this argument further.

Existing SSL methods typically follow the idea of ``do what can lead to a good representation'', such as using contrastive learning or masked prediction. However, why these operations lead to good representations often requires strong prior assumptions or extensive empirical validation, making them costly and difficultly to explain. In contrast, \textbf{Universality} explicitly incorporates the three key capabilities of a good representation into a unified learning objective, making the ``do what'' much more straightforward: find an SSL method that directly models \textbf{Universality} in its training objective. This not only makes it easier to explain the effectiveness of the ``do what'' (since the training objective itself is \textbf{Universality}), but also greatly simplifies the technical path for identifying truly effective learning strategies. Using causal paths as a metaphor, the difference between the two approaches can be described as follows: 1) Traditional SSL: ``$\text{do what} \to \text{universality} \to \text{good representation}$''; 2) Our proposed SSL framework: ``$\text{universality} \to \text{good representation}$''. Therefore, explicitly proposing and defining \textbf{Universality} not only offers a new conceptual perspective for designing SSL methods, but also opens up a more direct and interpretable path for practical implementation, e.g., how to model \textbf{Universality}. It thus carries both conceptual novelty and practical value.

\subsection{Explicit Modeling Universality in SSL}
\label{sec:4.3}

Let the mini-batches be denoted by $X^{{\text{aug}}} = \{ X_{tr,l}^{{\text{aug}}}\} _{l = 1}^{m}$, where $l$ indexes the $l$-th mini-batch. The $l$-th mini-batch is denoted as
$X_{tr,l}^{{\text{aug}}} = \{ x_{i,l}^1,x_{i,l}^2,x_{i,l}^{{\text{anchor}}}\} _{i = 1}^{n}$, with $\{x_{i,l}^{1},\,x_{i,l}^{2},\,x_{i,l}^{\text{anchor}}\}$ representing the $i$-th pair and $x_{i,l}^{\text{anchor}}$ is the anchor within that pair. For D-SSL methods, all three samples in a pair originate from the same source image and are produced by applying different data-augmentation pipelines to that image. Any of the three augmented views may serve as the anchor. Whichever sample is chosen, the triplet can always be written as $\{x_{i,l}^{1},\,x_{i,l}^{2},\,x_{i,l}^{\text{anchor}}\}$. To keep the notation concise, we do not label which element is the anchor, instead, we treat pairs with different anchor choices as distinct pairs indexed by $i$. For G-SSL methods, the anchor of each pair is the source sample.
The remaining two elements, $x_{i,l}^{1}$ and $x_{i,l}^{2}$, are generated from that anchor by applying two different masking operations.

During training, every mini-batch is divided at the pair level into two disjoint subsets: support set $X_{tr,l}^{\sup } = \{ x_{i,l}^1,x_{i,l}^{{\text{anchor}}}\} _{i = 1}^{n}$ and query set $X_{tr,l}^{{\text{que}}} = \{ x_{i,l}^2,x_{i,l}^{{\text{anchor}}}\} _{i = 1}^{n}$. At each training step, $m$ such mini-batches are processed in parallel. The GeSSL objective over these mini-batches is presented as:
\begin{equation}\label{asdgsg}
\mathop {\min }\limits_{f,g} \sum\nolimits_{l = 1}^{m} {{\mathcal{L}_{ssl}}(f',X_{tr,l}^{{\text{que}}})}, \;{\rm{s}}{\rm{.t}}{\rm{. }}\; f' = \arg \mathop {\min }\limits_f \sum\nolimits_{l = 1}^{m} {[{\mathcal{L}_{ssl}}(f,X_{tr,l}^{\text{sup}}) + \mu {\mathcal{L}_{disc}}(f,g,X_{tr,l}^{\text{sup}})]},
\end{equation}
where ${\mathcal{L}_{ssl}}(\cdot)$ represents the loss function in SSL method, e.g., the contrastive loss in D-SSL and the MSE loss in G-SSL, $\mu$ is a hyperparameter, and ${\mathcal{L}_{disc}}(f,g,X_{tr,l}^{\text{sup}})$ is a defined discriminative loss:
\begin{equation}\label{SADASD}
{\mathcal{L}_{disc}}(f,g,X_{tr,l}^{\text{sup}})= \sum\nolimits_{i=1}^{n} \sum\nolimits_{j=1}^{n} [
\mathbbm{1}_{\{d_j^i \leq a_i\}} \cdot d_j^i + \mathbbm{1}_{\{d_j^i > a_i\}} \cdot (-d_j^i)
],
\end{equation}
where $a_i \in \mathbb{R}$ is the output of the function $g$, the input to $g$ is the mean and covariance matrix of the vector set $\{f(x_{j,l}^1) - f(x_{i,l}^{\text{anchor}})\}_{j=1}^{n}$, $\mathbbm{1}_{\{\cdot\}}$ denotes the indicator function, which evaluates whether a given condition is satisfied, $d_j^i=d(f(x_{j,l}^1),f(x_{i,l}^{{\text{anchor}}}))$, and $d(\cdot)$ is the cosine distance. As we can see, minimizing ${\mathcal{L}_{\text{disc}}}(g, X_{tr,l}^{\text{sup}})$ can be interpreted as: 1) when $d_j^i \leq a_i$, minimize $d_j^i$; 2) when $d_j^i > a_i$, maximize $d_j^i$. Since the indicator function is non-differentiable, optimizing $g$ leads to a zero-gradient problem. Therefore, we replace Equation (\ref{SADASD}) with a differentiable approximation. Based on \cite{maddison2016concrete, musgrave2020metric, sohn2016improved}, this differentiable approximation can be expressed as:
\begin{equation}\label{SsgADAsdfSD}
{\mathcal{L}_{disc}}(f,g,X_{tr,l}^{\text{sup}}) =  \sum\nolimits_{i=1}^{n} \sum\nolimits_{j=1}^{n} \left[ w(a_i) \cdot d_j^i + (1 - w(a_i)) \cdot (-d_j^i) \right],
\end{equation}
where $w(a_i) = \text{Sigmoid} ( k \cdot (a_i - d_j^i) )$, and $k \in \mathbbm{R}^+$ is a hyperparameter. Finally, the specific optimization process of Equation (\ref{asdgsg}) is divided into the following two steps of iteration:

\textbf{Inner-Loop Optimization:} In this step, $g$ is fixed, and GeSSL learns a proxy model 
$f'$ by minimizing the constraint of Equation (\ref{asdgsg}). The update of $f'$ can be obtained by the follows: 
\begin{equation}
\label{equ:first-level}
    f'(f,g) \gets f - \alpha \nabla_{f}  \sum\nolimits_{l = 1}^{m} {[{\mathcal{L}_{ssl}}(f,X_{tr,l}^{\text{sup}}) + \mu {\mathcal{L}_{disc}}(f,g,X_{tr,l}^{\text{sup}})]},
\end{equation}
where $\alpha$ is the learning rate, and $f'(f,g)$ represents that $f'$ is a function of $f$ and $g$, this is because that Equation (\ref{equ:first-level}) explicitly represents the process of updating parameters based on gradient descent. Typically, $f'$ undergoes $\varsigma$ updates by executing Equation (\ref{equ:first-level}) $\varsigma$ times, and each resulting $f'$ can be expressed as a function of $f$ and $g$, i.e., $f' = f'(f, g)$. We set $\varsigma = 1$ for computational convenience. It should be noted that the terms $\mathcal{L}_{ssl}$ and $\mathcal{L}_{disc}$ are all calculated based on the support set.

\textbf{Outer-Loop Optimization:} In this step, GeSSL learns the optimal model $f$ and $g$, based on the agent model $f'$. The learning process is presented as the follows: 
\begin{equation}
\label{equ:second-level}
    f \gets f - \beta \sum\nolimits_{l =1}^{m} \nabla_{f}  \mathcal{L}_{ssl} (f'(f,g),X_{tr,l}^{\text{que}});\quad g \gets g - \beta \sum\nolimits_{l =1}^{m} \nabla_{g}  \mathcal{L}_{ssl} (f'(f,g),X_{tr,l}^{\text{que}}),
\end{equation}
where $\beta$ is the learning rate, $\mathcal{L}_{ssl} (f'(f,g),X_{tr,l}^{\text{que}})$ is calculated based on the query set and the proxy model $f'$. From Equation (\ref{equ:second-level}), $f'$ contains both the parameters of $f$ and the first-order derivatives of $f$, and since $\mathcal{L}_{ssl}(f'(f, g), X_{tr,l}^{\text{que}})$ can be viewed as a function of $f'(f, g)$, the gradient of $\mathcal{L}_{ssl}(f'(f, g), X_{tr,l}^{\text{que}})$ with respect to $f$ involves both the first and second order derivatives of $f$.

\textbf{Explanation for Bi-Level Optimization}: Equation (\ref{SADASD}) belongs to a bi-level optimization objective. The inner constraint aims to learn a proxy model $f'$, while the outer objective is to ultimately learn a better $f$ based on $f'$. The advantage of this design lies in the fact that it enables $\mathcal{L}_{ssl}$ to be minimized twice, thereby potentially yielding a better $f$. Specifically, the first minimization of $\mathcal{L}_{ssl}$ occurs during the learning of $f'$, as obtaining $f'$ involves minimizing $\mathcal{L}_{ssl}$. As discussed in Section \ref{CauDC:ulf}, a single mini-batch in SSL can be regarded as a task, and dividing a task’s dataset into a support set and a query set allows us to reasonably assume that both sets are drawn from the same underlying distribution. Therefore, if $f'$ minimizes $\mathcal{L}_{ssl}$ on the support set, it can also be expected to perform well on the query set. Moreover, in the outer objective of Equation (\ref{SADASD}), we further adjust $f$ to obtain a new $f'$ such that the value of $\mathcal{L}_{ssl}$ computed using this updated $f'$ is lower than that computed using the previous $f'$, thus, this constitutes the second minimization. Also, this process can be interpreted as modeling the behavior of selecting the best among many local minima. Hence, compared to jointly optimizing the outer and inner objectives in a single stage, the bi-level optimization framework can yield a better $f$, because only a better $f$ leads to a lower $\mathcal{L}_{ssl}$ when propagated through $f'$.

\textbf{Explanation for Discriminability}: GeSSL models discriminability from two complementary perspectives. The first dimension is captured via the loss $\mathcal{L}_{ssl}$, while the second dimension is addressed through the loss $\mathcal{L}_{disc}$. In the first dimension, both D-SSL and G-SSL formulations of $\mathcal{L}_{ssl}$ include an ``alignment term'', which enforces each augmented sample in a pair to align closely with its corresponding anchor. According to Section \ref{CauDC:ulf}, each mini-batch can be regarded as a task, where samples within the same pair are assumed to belong to the same semantic class, while samples across different pairs correspond to different classes. The anchor sample within each class acts as a proxy for the class center. Therefore, minimizing the alignment term effectively encourages intra-class compactness by pulling together samples of the same class. From this perspective, $\mathcal{L}_{ssl}$ implicitly models discriminability. However, this modeling is limited due to the coarse class assignment strategy in SSL. Specifically, current SSL methods treat augmented views derived from the same source as belonging to the same class, while failing to consider that different sources may actually belong to the same underlying class. As a result, augmented samples from different sources, but sharing semantic similarity, are not encouraged to align. Furthermore, $\mathcal{L}_{ssl}$ lacks an explicit mechanism for pushing apart samples from different classes. Hence, relying solely on $\mathcal{L}_{ssl}$ under this simplistic partitioning is insufficient for capturing a rich notion of discriminability.

To address the above limitation, GeSSL introduces $\mathcal{L}_{disc}$ as a second mechanism to enhance discriminability. Minimizing $\mathcal{L}_{disc}$ can be interpreted as follows: for a given anchor and a learnable threshold, if the distance between the anchor and an augmented sample is less than or equal to the threshold, the sample is pulled closer to the anchor, otherwise, it is pushed farther away. This operation is applied across all anchors, enabling the model to approximately group together semantically similar samples while separating dissimilar ones. In essence, $\mathcal{L}_{disc}$ compensates for the shortcomings of $\mathcal{L}_{ssl}$ by explicitly promoting both intra-class compactness and inter-class separability. However, the effectiveness of $\mathcal{L}_{disc}$ hinges on the quality of the threshold: an inaccurate threshold can undermine its discriminative power. This is where the bi-level optimization framework in GeSSL plays a crucial role. The threshold, like the model $f$, is learned by adjusting an auxiliary network $g$, such that the proxy model $f'$ computed based on $f$ and $g$ leads to a lower $\mathcal{L}_{ssl}$. In other words, GeSSL optimizes $g$ so that the learned threshold helps find a better local minimum of $\mathcal{L}_{ssl}$ via $f'$. Only when $g$ learns an accurate threshold can this two-stage minimization be effective. Thus, the bi-level framework implicitly regularizes the learning of $g$, ensuring its accuracy. Ultimately, this design guarantees that the introduction of $\mathcal{L}_{disc}$, guided by a well-learned threshold, enables GeSSL to model discriminability in a more comprehensive and principled manner.

\textbf{Explanation for Generalizability}: GeSSL models generalizability from two dimensions. The first is direct modeling. Specifically, during training, GeSSL divides each mini-batch task into a support set and a query set, with no overlap between the two. Under this mechanism, GeSSL first fine-tunes $f$ using the support set to obtain a task-specific model $f'$. Then, it evaluates the performance of $f'$ using the query set and uses the evaluation result to update the original model $f$ via backpropagation. This process crucially involves two rounds of gradient updates: the first transforms $f$ into $f'$, and the second refines $f$ based on the feedback from the query set. It is this “update-then-evaluate” mechanism that allows the query set to serve not just as training data, but as a simulated test set. From this perspective, the roles of the support and query sets in GeSSL resemble the training and test sets in conventional training: the support set enables task adaptation, while the query set assesses generalization. However, unlike conventional training where test set performance is evaluated only after training completes, GeSSL incorporates ``performing well on the test set'' as an explicit training objective, thus directly modeling generalizability. The second dimension is indirect modeling. Prior research has shown that when a model performs well across multiple diverse tasks, it often implies that the model has learned causal representations. \cite{ahuja2020invariant} argues that consistent performance across tasks suggests the model has captured stable causal features. \cite{scholkopf2021toward} and \cite{ahuja2023interventional} further contend that causal representations are a sufficient condition for achieving generalizability. Based on this theoretical foundation, the training strategy of GeSSL inherently reflects causal modeling: in each training round, multiple mini-batches (each corresponding to a distinct task ) are simultaneously fed into the model. GeSSL uses the same $f$ and its updated version $f'$ to adapt to all these tasks. This cross-task consistency constraint encourages the model to discover stable features shared across tasks, thereby capturing underlying causal structures. In essence, this enables GeSSL to indirectly model generalizability by promoting the learning of causal representations.

\textbf{Explanation for Transferability}: The training process of GeSSL can be regarded as an episodic learning process like meta-learning. Specifically, each episode of GeSSL consists of $m$ mini-batch tasks, and the entire learning process can be divided into multiple episodes. Based on Section \ref{CauDC:ulf}, we consider the learning process of GeSSL as estimating the true task distribution from discrete training tasks, which enables the GeSSL model to generalize to new, unseen tasks (i.e., test tasks). Therefore, we conclude that GeSSL achieves model transferability through its learning paradigm.

\textbf{Comparison of GeSSL and Meta-Learning}: It is important to note that the primary goal of GeSSL is to provide an effective method for modeling Universality. In other words, GeSSL seeks to answer the question: how can Universality be modeled? Its design is explicitly inspired by the concept of Universality. At the same time, it is crucial to highlight the differences between GeSSL and traditional meta-learning methods. First, meta-learning relies on explicit supervision, whereas GeSSL operates in a self-supervised manner, constructing its own pseudo-labels without the need for manual annotations. Second, although both methods follow an episodic training paradigm, meta-learning benefits from accurate supervision, which allows it to model discriminability effectively. In contrast, GeSSL relies on heuristically constructed labels that are often noisy, which can hinder its ability to model discriminability accurately. To address this, GeSSL introduces an additional loss term $\mathcal{L}_{disc}$ to mitigate the impact of noisy labels and enhance its capacity to model true class structures. Moreover, in meta-learning, a separate task-specific model is learned for each task, meaning different tasks have different adapted models. GeSSL, on the other hand, learns a single unified adapted model $f'$ for all tasks. This key difference enables GeSSL to better capture shared structures across tasks, thereby learning representations with stronger generalizability. In summary, GeSSL is not only distinct from traditional meta-learning in its methodology and learning signals but also differs fundamentally from approaches that directly transplant meta-learning paradigms into SSL. GeSSL’s design reflects a unique theoretical motivation and a novel approach to learning universal representation.

\section{Theoretical Analysis}
\label{sec:5}

In this section, we provide performance guarantees for GeSSL. Specifically, we prove that through the objective of GeSSL (Equation (\ref{equ:second-level})), the performance of the SSL model on new tasks is guaranteed.
We assume a task distribution $\mathcal{T}$, where each task $\tau\sim\mathcal{T}$ comprises a support set $S_\tau$ (used for rapid task‐specific adaptation) and a query set $Q_\tau$ (used to evaluate generalization). For any parameter vector $\theta$ and task $\tau$, we denote the supervised loss incurred on the unseen query set $Q_\tau$ as $\mathcal{L}_{\text{sup}}(\theta;\tau)$; the self-supervised and discriminative losses on the support set $S_\tau$ as $\mathcal{L}_{{ssl}}(\theta;S_\tau)$ and $\mathcal{L}_{{disc}}(\theta;S_\tau)$; letting $\theta' = A(\theta, S_\tau)$ be the adapted parameters after applying the adaptation operator $A$ to $\theta$ using $S_\tau$; $\mathcal{L}_{{query}}(\theta';Q_\tau)$ be the resulting SSL loss in query set. By jointly optimizing these losses, our goal is to learn representations that are both transferable and generalizable to new tasks while ensuring discriminative performance. 
Next, we provide the main theorem with detailed proofs in Appendix \ref{sec:app_B}.
\begin{theorem}\label{theorem:1}
    Let $\theta^*$ denote the parameter after bi-level training over $N$ tasks (mini-batches). For any new task \(\tau_{\text{test}} \sim \mathcal{T}\), let \(\theta^*_{\text{test}} = A(\theta^*, S_{\tau_{\text{test}}})\) denote the adapted parameter, under Assumption~\ref{ass:main}, with probability at least $1 - \delta$, we have:
    \begin{equation}
    \resizebox{0.93\linewidth}{!}{$
        \mathbb{E}_{\tau_{\text{test}}} \bigl[ \mathcal{L}_{\text{sup}}(\theta^*_{\text{test}}; \tau_{\text{test}}) \bigr] \le \frac{1}{N} \sum\nolimits_{i=1}^N \biggl[ \mathcal{L}_{\text{ssl}}(\theta', S_{\tau_i}) + \mathcal{L}_{\text{disc}}(\theta', S_{\tau_i}) + \mathcal{L}_{\text{query}}(\theta', Q_{\tau_i}) \biggl] + \mathcal{O}(\sqrt{\frac{1}{N} \ln \frac{1}{\delta}}),
    $}    
    \end{equation}
    where $\theta'$ is the adapted parameter for training task $\tau_i$ (the $i$-th mini-batch).
\end{theorem}
This theorem states that, under standard assumptions such as smoothness and boundedness, the bi-level training procedure of GeSSL, which jointly optimizes the SSL loss, the discriminative loss (to enhance class separability), and the query loss (to guarantee generalization to new tasks), provides an upper bound of order $\mathcal{O}(\sqrt{\ln(1/\delta)/N})$ on the supervised loss for unseen tasks. This result formally validates both the effectiveness and the broad applicability of the GeSSL strategy.

\section{Empirical Evaluation}
\label{sec:6}
In this section, we conduct extensive experiments on various settings to verify the effectiveness of GeSSL. 
For unsupervised and semi-supervised learning, we select CIFAR-10 \cite{CIFAR-10-100}, CIFAR-100 \cite{CIFAR-10-100}, STL-10 \cite{STL-10}, Tiny ImageNet \cite{TinyImagenet}, ImageNet-100 \cite{Imagenet100} and ImageNet \cite{ImageNet}; For transfer learning, we select PASCAL VOC \cite{PASCAL}, COCO \cite{COCO}, Flower102 \cite{nilsback2008automated}, Food101 \cite{bossard2014food}, etc.; For few-shot learning, we select Omniglot \cite{Omniglot}, miniImageNet \cite{miniImagenet}, CIFAR-FS \cite{CIFAR-FS}, CUB \cite{cub}, Cars \cite{cars}, etc., for evaluation. We select both D-SSL and G-SSL baselines for comparison.
All results are reported via five runs on NVIDIA 4090 GPUs. More details and additional results are provided in Appendix \ref{sec:app_C}-\ref{sec:app_G}.

\begin{table}[t]
\begin{minipage}[t]{0.49\textwidth}
	\centering
	\caption{The Top-1 and Top-5 classification accuracies of linear classifier on the ImageNet-100 and ImageNet (200 Epochs) with ResNet-50.}
	\label{tab:2} 
 \resizebox{\linewidth}{!}{
	\begin{tabular}{lcccc}
		\toprule
            \multirow{2.5}{*}{Method} & \multicolumn{2}{c}{ImageNet-100} & \multicolumn{2}{c}{ImageNet} \\
		\cmidrule(lr){2-3} \cmidrule(lr){4-5} 
		& Top-1 & Top-5 & Top-1 & Top-5\\
	    \midrule
	    SimCLR \cite{simclr} & 70.15 $\pm$ 0.16 & 89.75 $\pm$ 0.14 & 68.32 $\pm$ 0.31 & 89.76 $\pm$ 0.23 \\
		MoCo \cite{moco} & 72.80 $\pm$ 0.12 & 91.64 $\pm$ 0.11 & 67.55 $\pm$ 0.27 & 88.42 $\pm$ 0.11 \\ 
            SimSiam \cite{simsiam} & 73.01 $\pm$ 0.21 & 92.61 $\pm$ 0.27 & 70.02 $\pm$ 0.14 & 88.76 $\pm$ 0.23 \\ 
            Barlow Twins \cite{barlowtwins} & 75.97 $\pm$ 0.23 & 92.91 $\pm$ 0.19 & 69.94 $\pm$ 0.32 & 88.97 $\pm$ 0.27 \\
		MAE \cite{mae} & 76.56 $\pm$ 0.16 & 93.24 $\pm$ 0.24 & 70.73 $\pm$ 0.25 & 91.41 $\pm$ 0.27 \\
            DINO \cite{dino} & 75.43 $\pm$ 0.18 & 93.32 $\pm$ 0.19 & 70.58 $\pm$ 0.24 & 91.32 $\pm$ 0.27 \\
            W-MSE \cite{wmse} & 76.01 $\pm$ 0.27 & 93.12 $\pm$ 0.21 & 70.85 $\pm$ 0.31 & 91.57 $\pm$ 0.20 \\
            RELIC v2 \cite{RELIC-v2} & 75.88 $\pm$ 0.15 & 93.52 $\pm$ 0.13 & 70.98 $\pm$ 0.21 & 91.15 $\pm$ 0.26 \\
		LMCL \cite{LMLC} & 75.89 $\pm$ 0.19 & 92.89 $\pm$ 0.28 & 70.83 $\pm$ 0.26& 90.04 $ \pm$ 0.21 \\
        ReSSL \cite{ressl} & 75.77 $\pm$ 0.21 & 92.91 $\pm$ 0.27 & 69.92 $\pm$ 0.24 &91.25 $\pm$ 0.12\\
        CorInfoMax \cite{CorInfoMax}& 75.54 $\pm$ 0.20 & 92.23 $\pm$ 0.25 & 70.83 $\pm$ 0.15 &91.53 $\pm$ 0.22 \\
        MEC \cite{MEC}& 75.38 $\pm$ 0.17 & 92.84 $\pm$ 0.20 & 70.34 $\pm$ 0.27 & 91.25 $\pm$ 0.38 \\
        VICRegL \cite{Vicregl}& 75.96 $\pm$ 0.19 & 92.97 $\pm$ 0.26 & 70.24 $\pm$ 0.27 & 91.60 $\pm$ 0.24 \\
        \midrule
    \rowcolor{blue!10}SimCLR + GeSSL & 72.96 $\pm$ 0.24 & 92.50 $\pm$ 0.17 & 69.88 $\pm$ 0.21 & 91.32 $\pm$ 0.25 \\
    \rowcolor{blue!10}MoCo + GeSSL & 74.35 $\pm$ 0.24 & 94.10 $\pm$ 0.31 & 69.60 $\pm$ 0.30 & 91.28 $\pm$ 0.39 \\
    \rowcolor{blue!10}SimSiam + GeSSL & 75.93 $\pm$ 0.24 & 95.51 $\pm$ 0.38 & 72.04 $\pm$ 0.22 & 89.43 $\pm$ 0.40 \\
    \rowcolor{blue!10}Barlow Twins + GeSSL & 77.55 $\pm$ 0.29 & 93.48 $\pm$ 0.30 & 72.84 $\pm$ 0.26 & 89.50 $\pm$ 0.19 \\
    \rowcolor{blue!10}MAE + GeSSL & 78.45 $\pm$ 0.31 & \textbf{96.17 $\pm$ 0.26} & 71.45 $\pm$ 0.24 & 89.68 $\pm$ 0.27 \\
    \rowcolor{blue!10}DINO + GeSSL & 77.13 $\pm$ 0.29 & 95.75 $\pm$ 0.30 & 73.52 $\pm$ 0.30 & \textbf{94.05 $\pm$ 0.26} \\
    \rowcolor{blue!10}VICRegL + GeSSL & \textbf{78.48 $\pm$ 0.34} & 95.90 $\pm$ 0.17 & \textbf{73.91 $\pm$ 0.36} & 93.77 $\pm$ 0.35 \\

		\bottomrule
	\end{tabular}
 }
\end{minipage}
\hfill 
\begin{minipage}[t]{0.48\textwidth}
	\centering
	\caption{The semi-supervised learning accuracies ($\pm $ 95\% confidence interval) on ImageNet with the ResNet-50 pre-trained on Imagenet.}
\label{tab:3}
\resizebox{0.98\linewidth}{!}{
		\begin{tabular}{lccccc}
		\toprule
		\multirow{2.5}{*}{Method} &\multirow{2.5}{*} {Epochs} &\multicolumn{2}{c}{1\%} & \multicolumn{2}{c}{10\%} \\
	    \cmidrule(lr){3-4} \cmidrule(lr){5-6}
	    & & Top-1 & Top-5 & Top-1 & Top-5 \\
	     \midrule
	    
	     MoCo \cite{moco}&200&43.8 $\pm$ 0.2 & 72.3 $\pm$ 0.1 &61.9 $\pm$ 0.1 &84.6 $\pm$ 0.2\\
	     BYOL \cite{byol}&200&54.8 $\pm$ 0.2 &78.8 $\pm$ 0.1&68.0 $\pm$ 0.2&88.5 $\pm$ 0.2\\
	   \midrule
	     \rowcolor{blue!10}MoCo + GeSSL & 200 & 46.6 $\pm$ 0.3 & 74.5 $\pm$ 0.3 & 63.8 $\pm$ 0.2 & 85.9 $\pm$ 0.2 \\
          \rowcolor{blue!10}BYOL + GeSSL & 200 & \textbf{57.3 $\pm$ 0.2} & \textbf{79.8 $\pm$ 0.2} & \textbf{71.1 $\pm$ 0.2} & \textbf{90.1 $\pm$ 0.3} \\
	    \midrule
        SimCLR \cite{simclr} & 1000 & 48.3 $\pm$ 0.2 & 75.5 $\pm$ 0.1 & 65.6 $\pm$ 0.1 & 87.8 $\pm$ 0.2\\
	MoCo \cite{moco} &1000 &52.3 $\pm$ 0.1 & 77.9 $\pm$ 0.2 &68.4 $\pm$ 0.1 &88.0 $\pm$ 0.2\\
	BYOL \cite{byol} & 1000 & 56.3 $\pm$ 0.2 & 79.6 $\pm$ 0.2 & 69.7 $\pm$ 0.2& 89.3 $\pm$ 0.1\\
        SimSiam \cite{simsiam} & 1000 & 54.9 $\pm$ 0.2 & 79.5 $\pm$ 0.2 & 68.0 $\pm$ 0.1 &89.0 $\pm$ 0.3 \\
        Barlow Twins \cite{barlowtwins} & 1000 & 55.0 $\pm$ 0.1& 79.2 $\pm$ 0.1 & 67.7 $\pm$ 0.2 & 89.3 $\pm$ 0.2\\
	     RELIC v2 \cite{RELIC-v2} &1000 & 55.2 $\pm$ 0.2 & 80.0 $\pm$ 0.1& 68.0 $\pm$ 0.2 & 88.9 $\pm$ 0.2\\
	     LMCL \cite{LMLC} & 1000 & 54.8 $\pm$ 0.2 & 79.4 $\pm$ 0.2 & 70.3 $\pm$ 0.1  & 89.9 $\pm$ 0.2\\
	     ReSSL \cite{ressl} & 1000 & 55.0 $\pm$ 0.1 & 79.6 $\pm$ 0.3 & 69.9 $\pm$ 0.1 & 89.7 $\pm$ 0.1\\
	     SSL-HSIC \cite{ssl-hsic} & 1000 & 55.4 $\pm$ 0.3 & 80.1 $\pm$ 0.2 & 70.4 $\pm$ 0.1 & 90.0 $\pm$ 0.1 \\
      CorInfoMax \cite{CorInfoMax}& 1000 & 55.0 $\pm$ 0.2 & 79.6 $\pm$ 0.3 & 70.3 $\pm$ 0.2 & 89.3 $\pm$ 0.2\\
      MEC \cite{MEC}& 1000 & 54.8 $\pm$ 0.1 & 79.4 $\pm$ 0.2&  70.0 $\pm$ 0.1 & 89.1 $\pm$ 0.1\\
      VICRegL \cite{Vicregl}& 1000 & 54.9 $\pm$ 0.1 & 79.6 $\pm$ 0.2 & 67.2 $\pm$ 0.1  & 89.4 $\pm$ 0.2\\
	   \midrule
    \rowcolor{blue!10}SimCLR + GeSSL & 1000 & 51.1 $\pm$ 0.2 & 77.7 $\pm$ 0.1 & 67.8 $\pm$ 0.3 & 89.8 $\pm$ 0.3 \\
    \rowcolor{blue!10}MoCo + GeSSL   & 1000 & 54.0 $\pm$ 0.3 & 78.8 $\pm$ 0.1 & 71.6 $\pm$ 0.2 & 89.3 $\pm$ 0.2 \\
    \rowcolor{blue!10}BYOL + GeSSL   & 1000 & \textbf{59.6 $\pm$ 0.3} & \textbf{81.9 $\pm$ 0.2} & \textbf{71.8 $\pm$ 0.2} & 91.3 $\pm$ 0.2 \\
    \rowcolor{blue!10}Barlow Twins + GeSSL & 1000 & 58.1 $\pm$ 0.3 & 80.5 $\pm$ 0.2 & 68.9 $\pm$ 0.2 & \textbf{92.3 $\pm$ 0.3} \\

		\bottomrule
	\end{tabular}}
\end{minipage}
\end{table}

\begin{table*}[t]
	\centering
	\caption{The results of transfer learning on object detection and instance segmentation with C4-backbone as the feature extractor. ``AP'' is the average precision, ``$\text{AP}_{N}$'' represents the average precision when the IoU (Intersection and Union Ratio) threshold is $N\%$.}
\label{tab:4}
		\resizebox{\linewidth}{!}{
		
		\begin{tabular}{lcccccccccccc}
		\toprule
		\multirow{2.5}{*}{Method} &\multicolumn{3}{c}{VOC 07 detection} & \multicolumn{3}{c}{VOC 07+12 detection} &\multicolumn{3}{c}{COCO detection}&\multicolumn{3}{c}{COCO instance segmentation}\\
	    \cmidrule(lr){2-4} \cmidrule(lr){5-7} \cmidrule(lr){8-10} \cmidrule(lr){11-13} 
	    & \(\mathbf{AP_{50}}\)& \(\mathbf{AP}\) & \(\mathbf{AP_{75}}\)& \(\mathbf{AP_{50}}\)& \(\mathbf{AP}\) & \(\mathbf{AP_{75}}\)& \(\mathbf{AP_{50}}\)& \(\mathbf{AP}\) & \(\mathbf{AP_{75}}\)& \(\mathbf{AP^{mask}_{50}}\)& \(\mathbf{AP^{mask}}\) & \(\mathbf{AP^{mask}_{75}}\)\\
	       \midrule
	     Supervised & 74.4 & 42.4 & 42.7 & 81.3 & 53.5 & 58.8 & 58.2 & 38.2 & 41.2 & 54.7 & 33.3 & 35.2\\
	   \midrule
	     SimCLR \cite{simclr} & 75.9 & 46.8 & 50.1 & 81.8 & 55.5 & 61.4 & 57.7 & 37.9 & 40.9 & 54.6 & 33.3 & 35.3\\
	     MoCo \cite{moco} & 77.1 & 46.8 & 52.5 & 82.5 & 57.4 & 64.0 & 58.9 & 39.3 & 42.5 & 55.8 & 34.4 & 36.5\\
	     BYOL \cite{byol} & 77.1 & 47.0 & 49.9 & 81.4 & 55.3 & 61.1 & 57.8 & 37.9 & 40.9 & 54.3 & 33.2 & 35.0\\
	     SimSiam \cite{simsiam} & 77.3 & 48.5 & 52.5 & 82.4 & 57.0 & 63.7 & 59.3 & 39.2 & 42.1 & 56.0 & 34.4 & 36.7\\
	     Barlow Twins \cite{barlowtwins} & 75.7 & 47.2 & 50.3 & 82.6 & 56.8 & 63.4 & 59.0 & 39.2 & 42.5 & 56.0 & 34.3 & 36.5\\
         MAE \cite{mae} & 77.4 & 48.6 & 53.0 & 82.9 & 57.8 & 63.9 & 60.2 & 39.1 & 42.9 & 55.9 & 35.2 & 36.4 \\  
        SwAV \cite{swav} & 75.5 & 46.5 & 49.6 & 82.6 & 56.1 & 62.7 & 58.6 & 38.4 & 41.3 & 55.2 & 33.8 & 35.9\\
	     MEC \cite{MEC} & 77.4 & 48.3 & 52.3 & 82.8 & 57.5 & 64.5 & 59.8 & 39.8 & 43.2 & 56.3 & 34.7 & 36.8\\
	     RELIC v2 \cite{RELIC-v2} & 76.9 & 48.0 & 52.0 & 82.1 & 57.3 & 63.9 & 58.4 & 39.3 & 42.3 & 56.0 & 34.6 & 36.3\\
	 CorInfoMax \cite{CorInfoMax}& 76.8 & 47.6 & 52.2 & 82.4 & 57.0 & 63.4 & 58.8 & 39.6 & 42.5 & 56.2 & 34.8 & 36.5\\
      VICRegL \cite{Vicregl}& 75.9 & 47.4 & 52.3 & 82.6 & 56.4 & 62.9 & 59.2 & 39.8 & 42.1 & 56.5 & 35.1 & 36.8\\    
     \midrule
\rowcolor{blue!10}SimCLR + GeSSL   & 78.1  & 49.4  & 52.1  & 84.5  & 58.2  & 63.3  & 59.1  & 40.1  & 43.4  & 56.8  & 35.9  & 36.5  \\
\rowcolor{blue!10}MoCo + GeSSL     & 78.7  & 50.0  & \textbf{54.9}  & \textbf{85.6}  & 60.5  & 65.9  & 61.5  & \textbf{42.8}  & \textbf{44.8}  & \textbf{58.9}  & 37.0  & 39.0  \\
\rowcolor{blue!10}BYOL + GeSSL     & 78.7  & 49.8  & 53.2  & 84.9  & 59.0  & 64.8  & 60.7  & 40.9  & 44.0  & 57.5  & 36.2  & 38.1  \\
\rowcolor{blue!10}MAE + GeSSL & 79.1 & \textbf{51.0} & 54.4 & 85.4 & \textbf{61.2} & \textbf{65.9} & 62.1 & 42.1 & 44.8 & 58.2 & \textbf{38.3} & 39.1 \\ 
\rowcolor{blue!10}SimSiam + GeSSL  & \textbf{79.3}  & 50.5  & 54.1  & 85.0  & 59.4  & 65.8  & 62.0  & 41.5  & 44.3  & 58.4  & 37.5  & \textbf{39.6}  \\
\rowcolor{blue!10}SwAV + GeSSL     & 78.4  & 49.3  & 52.3  & 84.8  & 58.7  & 65.1  & 61.3  & 40.7  & 43.9  & 57.0  & 36.6  & 38.8  \\
\rowcolor{blue!10}VICRegL + GeSSL  & 78.9  & 50.5  & 54.6  & 85.4  & 59.8  & 66.0  & \textbf{62.2}  & 42.2  & 44.5  & 58.7  & 37.8  & 39.4  \\

		\bottomrule
	\end{tabular}
	}
\end{table*}

\begin{table*}[t]
  \caption{Few-shot learning accuracies ($\pm $ 95\% confidence interval) on miniImageNet, Omniglot, and CIFAR-FS with C4. See Appendix \ref{sec:app_E} for the baselines' details, and Appendix \ref{sec:app_F} for full results.}
  \label{tab:few-shot}
  \centering
\resizebox{\linewidth}{!}{\begin{tabular}{lccccccccc}
  \toprule
\multirow{2}{*}{\textbf{Method}} & \multicolumn{3}{c}{\textbf{Omniglot}} & \multicolumn{3}{c}{\textbf{\emph{mini}ImageNet}} & \multicolumn{3}{c}{\textbf{CIFAR-FS}}\\
  \cmidrule(r){2-4}
  \cmidrule(r){5-7}
  \cmidrule(r){8-10}
    & \textbf{(5,1)} & \textbf{(5,5)} & \textbf{(20,1)} & \textbf{(5,1)} & \textbf{(5,5)} & \textbf{(20,1)} & \textbf{(5,1)} & \textbf{(5,5)} & \textbf{(20,1)}\\
    \midrule
    \rowcolor{gray!40}\multicolumn{10}{c}{\emph{Unsupervised Few-shot Learning}}\\
    \midrule
    CACTUs \cite{hsu2018unsupervised} & 65.29 $\pm$ 0.21 & 86.25 $\pm$ 0.19 & 49.54 $\pm$ 0.21 & 39.32 $\pm$ 0.28 & 53.54 $\pm$ 0.27 & 31.99 $\pm$ 0.29 & 40.02 $\pm$ 0.23 & 58.16 $\pm$ 0.22 & 35.88 $\pm$ 0.25  \\
    UMTRA \cite{khodadadeh2019unsupervised}& 83.32 $\pm$ 0.37 & 94.23 $\pm$ 0.35 & 75.84 $\pm$ 0.34 & 39.23 $\pm$ 0.34 & 51.78 $\pm$ 0.32 & 30.27 $\pm$ 0.34 & 41.61 $\pm$ 0.40 & 60.55 $\pm$ 0.38 & 37.10 $\pm$ 0.39  \\
    LASIUM \cite{khodadadeh2020unsupervised}& 82.38 $\pm$ 0.36 & 95.11 $\pm$ 0.36 & 70.23 $\pm$ 0.36 & 42.12 $\pm$ 0.38 & 54.98 $\pm$ 0.37 & 34.26 $\pm$ 0.35 & 45.33 $\pm$ 0.32 & 62.65 $\pm$ 0.33 & 38.40 $\pm$ 0.33  \\
    SVEBM \cite{kong2021unsupervised}& 87.07 $\pm$ 0.28 & 94.13 $\pm$ 0.27 & 73.33 $\pm$ 0.28 & 44.74 $\pm$ 0.29 & 58.38 $\pm$ 0.28 & 39.71 $\pm$ 0.30 & 47.24 $\pm$ 0.25 & 63.10 $\pm$ 0.28 & 40.10 $\pm$ 0.28  \\
    GMVAE \cite{lee2021meta}& 90.89 $\pm$ 0.32 & 96.05 $\pm$ 0.32 & 81.51 $\pm$ 0.33 & 42.28 $\pm$ 0.36 & 56.97 $\pm$ 0.38 & 39.83 $\pm$ 0.36 & 47.45 $\pm$ 0.36 & 63.20 $\pm$ 0.35 &  41.55 $\pm$ 0.35 \\
    PsCo \cite{jang2023unsupervised}& \textbf{96.18 $\pm$ 0.21} & 98.22 $\pm$ 0.23 & 89.32 $\pm$ 0.23 & 46.35 $\pm$ 0.24 & 63.05 $\pm$ 0.23 & 40.84 $\pm$ 0.27 & 51.77 $\pm$ 0.27 & 69.66 $\pm$ 0.26 & 45.08 $\pm$ 0.27 \\
    \midrule
    \rowcolor{gray!40}\multicolumn{10}{c}{\emph{Self-supervised Learning}}\\
    SimCLR \cite{simclr} & 90.83 $\pm$ 0.21 & 97.67 $\pm$ 0.21 & 81.67 $\pm$ 0.23 & 42.32 $\pm$ 0.38 & 51.10 $\pm$ 0.37 & 36.36 $\pm$ 0.36 & 49.44 $\pm$ 0.30 & 60.02 $\pm$ 0.29 & 39.29 $\pm$ 0.30  \\
    MoCo \cite{moco}& 87.83 $\pm$ 0.20 & 95.52 $\pm$ 0.19 & 80.03 $\pm$ 0.21 & 40.56 $\pm$ 0.34 & 49.41 $\pm$ 0.37 & 36.52 $\pm$ 0.38 & 45.35 $\pm$ 0.31 & 58.11 $\pm$ 0.32 & 37.89 $\pm$ 0.32 \\
    SwAV \cite{swav}& 91.28 $\pm$ 0.19 & 97.21 $\pm$ 0.20 & 82.02 $\pm$ 0.20 & 44.39 $\pm$ 0.36 & 54.91 $\pm$ 0.36 & 37.13 $\pm$ 0.37 & 49.39 $\pm$ 0.29 & 62.20 $\pm$ 0.30 & 40.19 $\pm$ 0.32  \\
    \rowcolor{blue!10}SimCLR + GeSSL & 94.35 $\pm$ 0.31 & \textbf{98.41 $\pm$ 0.19} & 90.23 $\pm$ 0.24 & 46.51 $\pm$ 0.29 & 62.56 $\pm$ 0.29 & 39.56 $\pm$ 0.12 & \textbf{52.72 $\pm$ 0.15} & 67.52 $\pm$ 0.11 & 46.81 $\pm$ 0.14 \\
    \rowcolor{blue!10}MoCo + GeSSL & 93.15 $\pm$ 0.15 & 97.84 $\pm$ 0.14 & 88.84 $\pm$ 0.13 & 47.23 $\pm$ 0.14 & 61.05 $\pm$ 0.12 & 40.75 $\pm$ 0.11 & 51.58 $\pm$ 0.09 & 66.25 $\pm$ 0.08 & 44.48 $\pm$ 0.11 \\
    \rowcolor{blue!10}SwAV + GeSSL & 96.18 $\pm$ 0.14 & 98.25 $\pm$ 0.18 & \textbf{91.62 $\pm$ 0.22} & \textbf{48.60 $\pm$ 0.15} & \textbf{63.56 $\pm$ 0.08} & \textbf{41.54 $\pm$ 0.23} & 52.33 $\pm$ 0.28 & \textbf{69.58 $\pm$ 0.25} & \textbf{47.56 $\pm$ 0.15} \\
    \bottomrule
  \end{tabular}}
\end{table*}

\subsection{Performance Comparison}
\label{sec:6.2}

\textbf{Unsupervised Learning.} We adopt the most commonly used protocol \cite{simclr}, freezing the feature extractor and training a linear classifier on top of it. We use Adam \cite{kingma2014adam} with Momentum and weight decay set at $0.8$ and $10^{-4}$. The linear classifier runs for 500 epochs with a batch size of 128 and a learning rate that starts at $5 \times 10^{-2}$ and decays to $5 \times 10^{-6}$. We use ResNet-18 for small-scale datasets (CIFAR-10, CIFAR-100, STL-10, and Tiny ImageNet) while using ResNet-50 for the medium-scale (ImageNet-100) and large-scale (ImageNet) datasets.
Table \ref{tab:2} shows that applying GeSSL significantly outperforms the state-of-the-art (SOTA) methods on all datasets and SSL baselines. 
The results demonstrate its ability to enhance SSL performance. See Appendix \ref{sec:app_F.1} for more details.

\textbf{Semi-supervised Learning.} We adopt the commonly used protocol \cite{barlowtwins} and create two balanced subsets by sampling 1\% and 10\% of the training dataset. We fine-tune the models for 50 epochs with learning rates of 0.05 and 1.0 for the classifier, 0.0001 and 0.01 for the backbone on the 1\% and 10\% subsets. 
Table \ref{tab:3} shows that the performance with GeSSL is superior to the SOTA methods. For example, when only 1\% labels are available, the improvement of GeSSL reaches more than 3\%.

\textbf{Transfer Learning.}  
We use Faster R-CNN \cite{ren2015faster} for VOC detection and Mask R-CNN \cite{he2017mask} for COCO detection and segmentation with the same C4-backbone \cite{wu2019detectron2}. We train the Faster R-CNN on the VOC 07+12 set (16K images) and reduce the initial learning rate by 10 at 18K and 22K iterations, while training on the VOC 07 set (5K images) with fewer iterations. For the Mask R-CNN, we train it on the COCO 2017 train split and report on the val split. See Appendix \ref{sec:app_F.3} for details.
Table \ref{tab:4} shows the great performance improvements achieved by GeSSL. After introducing GeSSL, the models achieve SOTA performance, surpassing the original baselines by about 3.4\%.

\begin{figure*}[t]
  \begin{minipage}{0.32\textwidth}
    \centering
    \includegraphics[width=\linewidth]{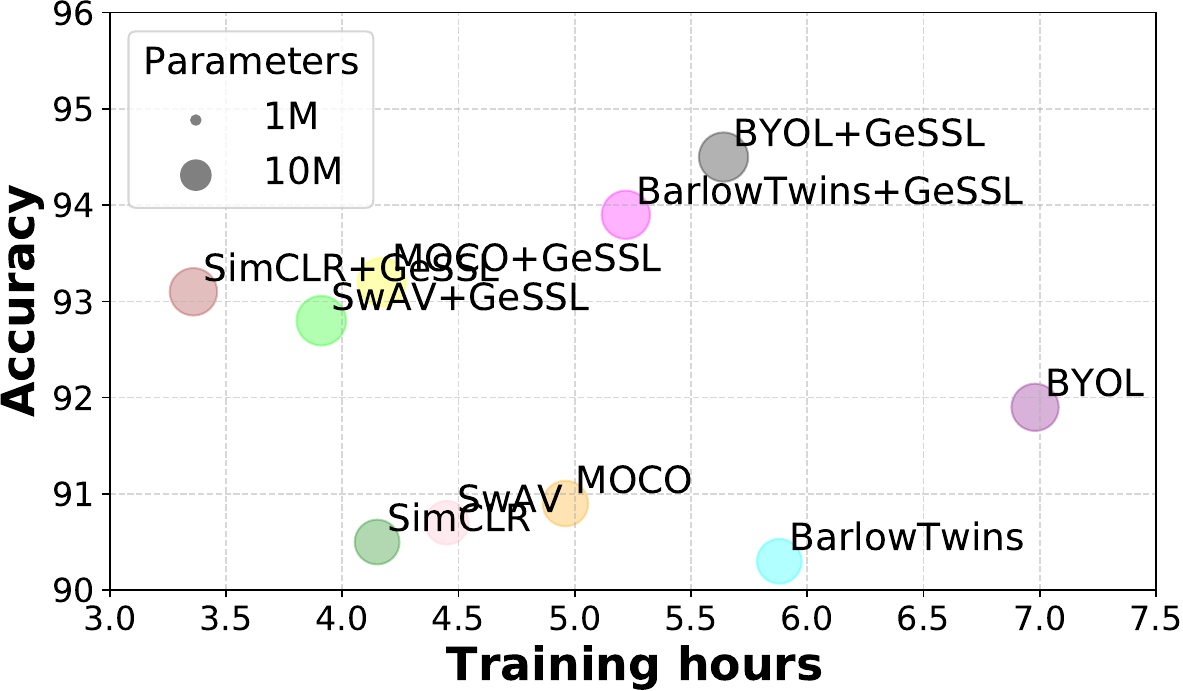}
    \caption{Model efficiency.}
    \label{fig:ab_2}
  \end{minipage}
  \hfill
  \begin{minipage}{0.33\textwidth}
    \centering
    \includegraphics[width=\linewidth]{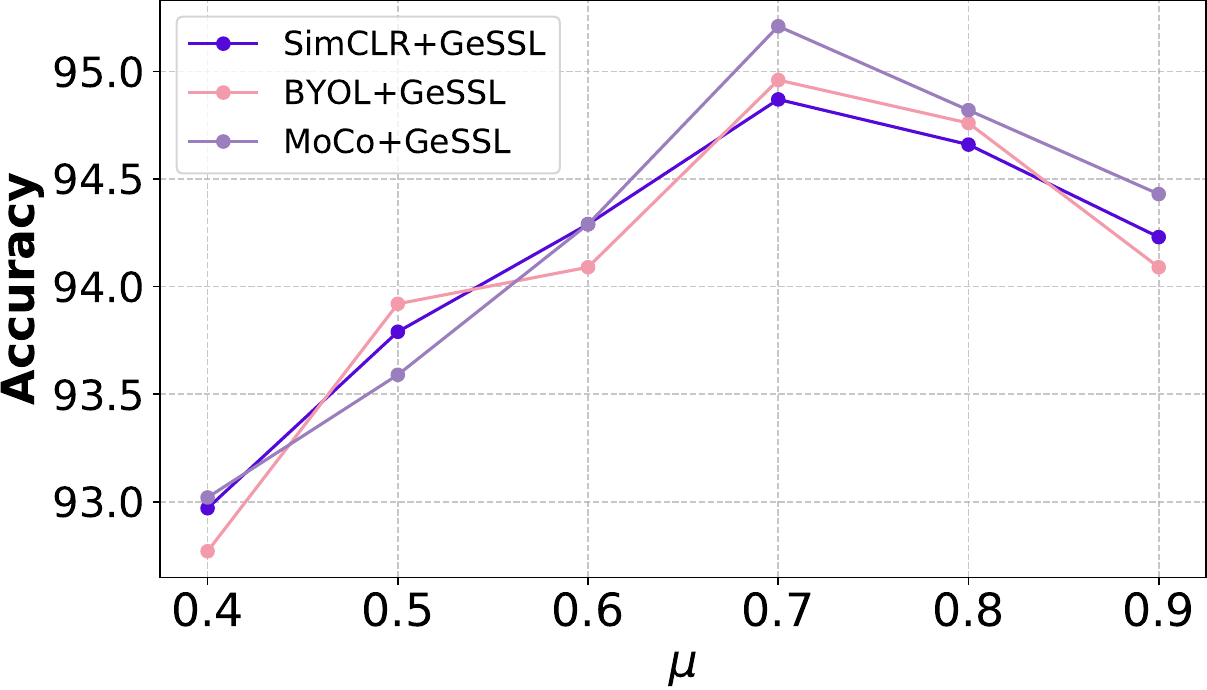}
    \caption{Ablation study of $\mu$.}
    \label{fig:abla_para_mu}
  \end{minipage}
  \hfill
  \begin{minipage}{0.33\textwidth}
    \centering
    \includegraphics[width=\linewidth]{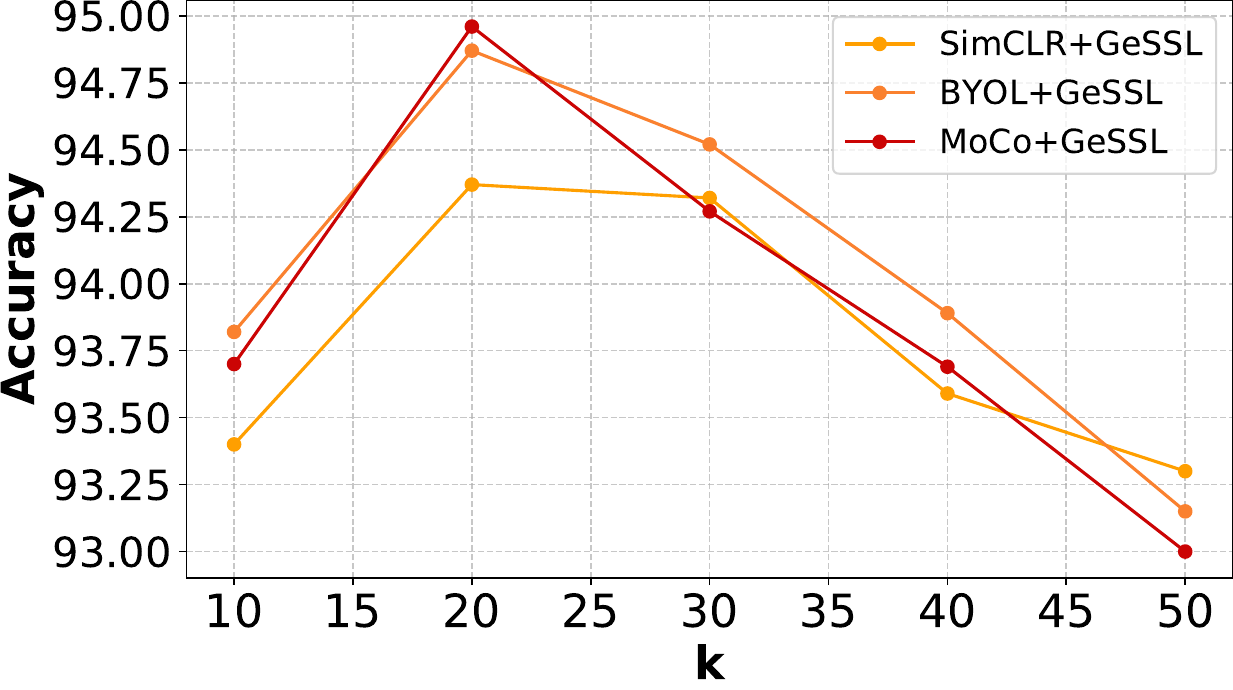}
    \caption{Ablation study of $k$.}
    \label{fig:abla_para_k}
  \end{minipage}
\end{figure*}

\textbf{Few-shot Learning.} 
We adopt the commonly used protocol \cite{jang2023unsupervised} on miniImageNet, Omniglot, and CIFAR-FS. 
For the few-shot SSL task, we randomly select $N$ samples without class-level overlap for each task, and then apply 2-times data augmentation, obtaining a $N$-way 2-shot task with $N$ classes and $2N$ samples. We use the SGD optimizer, setting the momentum and weight decay values to $0.9$ and $10^{-4}$ respectively. We evaluate the trained model's performance in some unseen samples sampled from a new class. 
Table \ref{tab:few-shot} shows the standard few-shot learning results of GeSSL compared with the baselines. From the results, we can see that our framework still achieves remarkable performance improvement, demonstrating the superiority of GeSSL. 
See Appendix \ref{sec:app_F.4} for more details.

\subsection{Ablation Study and Analysis}
\label{sec:6.6}
We conduct various ablation studies to evaluate how GeSSL works well, with details in Appendix \ref{sec:app_G}.

\noindent\textbf{Model efficiency.} We evaluate the trade-off performance of multiple baselines using GeSSL on STL-10 \cite{STL-10}. Figure \ref{fig:ab_2} shows that GeSSL achieves great performance and efficiency improvements with acceptable parameter size. Combining Appendix \ref{sec:app_G.4}, although GeSSL brings a larger memory footprint and parameter size costs, it is relatively negligible compared to the improvements.

\noindent\textbf{Effects of $\mathcal{L}_{disc}$.} 
To assess the impact of $\mathcal{L}_{disc}$, we visualize t-SNE feature clusters results before and after adding $\mathcal{L}_{disc}$ (see Appendix \ref{sec:app_G_ablation_disc} for details). Figure \ref{fig:app_G_ablation_disc} shows that incorporating $\mathcal{L}_{disc}$ yields sharper class boundaries on multiple SSL baselines, demonstrating its effect on discriminability.

\noindent\textbf{Parameter Sensitivity}
We evaluate the impact of hyperparameters, $\mu$, $k$, and batch size $M$ (see Appendix \ref{sec:app_G.5} for details). We search $\mu$ over $[0.3,1.0]$ with a step of 0.05 and $k$ over $[10,50]$ with 10, then refine in the best subranges. Figures \ref{fig:abla_para_mu}-\ref{fig:abla_para_k} show that the optimal settings are $\mu=0.7$ and $k=30$.

\noindent\textbf{Evaluation of the bi-level optimization.} 
To evaluate the benefit of our bi-level optimization, we compare it against two alternatives: (i) optimizing inner and outer objectives jointly in a single stage; and (ii) training a separate $f'$ for each mini-batch. As shown in Figure \ref{fig:abla_bi-level}, our bi-level optimization achieves SOTA performance. See Appendix \ref{sec:app_G.4} for details and more experiments.

\section{Related Work}
\label{sec:2}

Self-supervised learning (SSL) learns representations by transferring knowledge from pretext tasks without requiring labeled data. As outlined by \cite{jaiswal2020survey} and \cite{kang2023benchmarking}, SSL methods can be categorized into two main approaches: discriminative and generative SSL. Discriminative SSL methods, such as SimCLR \cite{simclr} and BYOL \cite{byol}, leverage stochastic data augmentation to create two augmented views from the same input sample. The goal is to maximize the similarity between these views in the embedding space, thereby learning meaningful representations. In contrast, generative SSL methods, like MAE \cite{mae} and VideoMAE \cite{tong2022videomae}, employ an encoder-decoder structure. These methods divide the input into multiple blocks, mask a subset of them, and reassemble the remaining blocks in their original positions.
Although SSL methods have demonstrated empirical success, several challenges remain \cite{jaiswal2020survey}. SSL models often struggle to generalize (i) when data is scarce \cite{krishnan2022self}, and (ii) in noisy real-world environments \cite{goyal2021self}. Moreover, the performance of SSL models is highly sensitive to the alignment between pretext and downstream tasks, which can impede effective transfer (demonstrated in Section \ref{sec:6}).
Thus, the universality of SSL is hard to get. Previous studies \cite{oord2018representation, hjelm2018learning, mizrahi20244m, tian2020makes, oquab2023dinov2} primarily focus on the empirical success of SSL methods, without addressing the critical question of what defines a ``good representation''. In this work, we bridge this gap by explicitly defining ``a good representation'' through a formalized framework, characterizing it as discriminability, generalizability, and transferability. More analyses and comparisons are provided in Appendices \ref{sec:app_F}-\ref{sec:app_H}.
\section{Conclusion}
\label{sec:7}

In this study, we explore the universality of SSL. We first unify SSL paradigms, i.e., discriminative and generative SSL, from the task perspective and propose the definition of SSL universality. It is a fundamental concept that involves discriminability, generalizability, and transferability. Then, we propose GeSSL to explicitly model universality into SSL through bi-level optimization, which introduces an auxiliary network to guide the model learn in the best direction. Extensive theoretical and empirical analyses demonstrate the effectiveness of GeSSL. 

\bibliography{main}

\newpage
\appendix
\onecolumn

\section*{Appendix}
The appendix is organized into several sections: 

\begin{itemize}
    \item Appendix \ref{sec:app_B} contains the analyses and proofs of the presented definitions and theorems.
    \item Appendix \ref{sec:app_C} presents the implementation and architecture of our GeSSL.
    \item Appendix \ref{sec:app_D} provides details for all datasets used in the experiments.
    \item Appendix \ref{sec:app_E} provides details for the baselines mentioned in the main text.
    \item Appendix \ref{sec:app_F} showcases additional experiments, full results, and experimental details of the comparison experiments that were omitted in the main text due to space limitations.
    \item Appendix \ref{sec:app_G} provides the additional experiments and full details of the ablation studies that were omitted in the main text due to page limitations.
    \item Appendix \ref{sec:app_H} provides the discussion about the proposed methodology.
\end{itemize}
Note that before we illustrate the details and analysis, we provide a brief summary about all the experiments conducted in this paper, as shown in Table \ref{tab:app}.

\begin{table}[htpb]
    \centering
    \caption{Illustration of the experiments conducted in this work. Note that all experimental results are obtained after five rounds of experiments.}
    \begin{tabular}{p{0.4\textwidth}|p{0.3\textwidth}|p{0.2\textwidth}}
    \toprule
        \textbf{Experiments} & \textbf{Location} & \textbf{Results}\\
    \midrule    
        Experiments of unsupervised learning on six benchmark dataset & Section \ref{sec:6.2} and Appendix \ref{sec:app_F.1} & Table \ref{tab:2}, Table \ref{tab:1}, Table \ref{tab:appendix_imagenet}, and Table \ref{tab:q2-1}\\
    \midrule    
        Experiments of semi-supervised learning on on ImageNet with two settings & Section \ref{sec:6.2} & Table \ref{tab:3} and Table \ref{tab:q2-2}\\
    \midrule    
        Experiment of transfer learning & Section \ref{sec:6.2} and Appendix \ref{sec:app_F.3} & Table \ref{tab:4}, Table \ref{tab:toy}, and Table \ref{tab:trans_video}\\
    \midrule    
        Experiment of few-shot learning on standard and cross-domain scenarios & Section \ref{sec:6.2} and Appendix \ref{sec:app_F.4} & Table \ref{tab:few-shot} and Table \ref{tab:cross-domain} \\
    \midrule
        Ablation study-Model efficiency & Section \ref{sec:6.6} and Appendix \ref{sec:app_G.2} & Figure \ref{fig:ab_2} and Table \ref{tab:app_model} \\
    \midrule
        Ablation study-Effect of $\mathcal{L}_{disc}$ & Section \ref{sec:6.6} and Appendix \ref{sec:app_G_ablation_disc} & Figure \ref{fig:app_G_ablation_disc} \\ 
    \midrule
        Ablation study-Parameter sensitivity & Section \ref{sec:6.6} and Appendix \ref{sec:app_G.5} & Figure \ref{fig:abla_para_k}, Figure \ref{fig:abla_para_mu}, Figure \ref{fig:app_batchsize}, and Figure \ref{fig:app_n} \\
    \midrule
        Ablation study-Evaluation of the bi-level optimization & Section \ref{sec:6.6} and Appendix \ref{sec:app_G.4} & Figure \ref{fig:ab_4} \\ 
    \midrule
        Universality of existing SSL methods & Appendix \ref{sec:app_F.5} & Figure \ref{fig:app_sigma} and Table \ref{tab:learning} \\
    \midrule
        Evaluation of generative SSL on three scenarios & Appendix \ref{sec:app_F.6} & Figure \ref{fig:app_gssl}, Table \ref{tab:ViT-Large}, Table \ref{tab:gssl_1}, and Table \ref{tab:gssl_3}\\
    \midrule
        Evaluation on more modalities & Appendix \ref{sec:app_F.7} & Table \ref{tab:app_modal} \\
    \bottomrule
    \end{tabular}
    \label{tab:app}
\end{table}

\section{Proofs}
\label{sec:app_B}
Before giving the main theorem, we first provide the assumptions.
\begin{assumption}\label{ass:main}
The following conditions are assumed to hold simultaneously:
\begin{enumerate}[label=(A\arabic*)]
\item IID Tasks: The training tasks $\tau_{1},\dots,\tau_{N}$ are sampled i.i.d. from the task distribution $\mathcal{T}$.
\item Bounded Losses: For any parameter $\theta$ and task $\tau$, all loss components satisfy $ 0 \le \mathcal{L}_{\sup}(\theta;\tau) \le \mathcal{L}_{\max}$, $0 \le \mathcal{L}_{ssl}(\theta;S_{\tau})\le \mathcal{L}_{\max}$, $0 \le \mathcal{L}_{disc}(\theta;S_{\tau})\le \mathcal{L}_{\max}$, and $0 \le \mathcal{L}_{query}(\theta';Q_{\tau})\le \mathcal{L}_{\max}$.
\item Gradient Lipschitz Continuity: There exists a constant $G>0$ such that for all $\theta,\theta'$ and any $\tau$, with $ \bigl\|\nabla_\theta\bigl(\mathcal{L}_{ssl}(\theta;S_\tau)+\mathcal{L}_{disc}(\theta;S_\tau)\bigr)
  -\nabla_\theta\bigl(\mathcal{L}_{ssl}(\theta';S_\tau)+\mathcal{L}_{disc}(\theta';S_\tau)\bigr)\bigr\|
  \le G\|\theta-\theta'\|$.
\item Inner-Loop Reduction: There exists $\Delta>0$ such that for all $\theta$ and $\tau$, have $ \mathcal{L}_{ssl}\bigl(A(\theta,S_\tau);S_\tau\bigr) + \mathcal{L}_{disc}\bigl(A(\theta,S_\tau);S_\tau\bigr)
  \le
  \mathcal{L}_{ssl}(\theta;S_\tau) + \mathcal{L}_{disc}(\theta;S_\tau) - \Delta$.
\item Fast Adaptation: There exists $C>0$ and a small sample size $m\ll|Q_\tau|$ such that for all $\theta$ and $\tau$, have $\mathcal{L}_{\sup}\bigl(A(\theta,S_\tau);\tau\bigr)
  \le
  \frac{C}{\sqrt{m}}\Bigl(\mathcal{L}_{ssl}(\theta;S_\tau)+\mathcal{L}_{disc}(\theta;S_\tau)\Bigr)$.
\end{enumerate}
\end{assumption}
These assumptions are recognized as mild conditions in both theory and practice. Specifically, (A1) assumes that training and test tasks are drawn i.i.d. to ensure that strategies learned on the training set generalize to new tasks—an assumption ubiquitous in generalization analyses \cite{Baxter_2000,maml}. In our setting, tasks (mini-batches) are constructed by randomly sampling from the data distribution and applying augmentations, which amounts to independently drawing from an underlying class distribution and sample distribution joint space and thus naturally satisfies the i.i.d. condition. (A2) imposes bounded losses, a requirement extensively validated in practice \cite{10.1093/acprof:oso/9780199535255.001.0001}. When deriving generalization bounds, we typically invoke concentration inequalities such as Hoeffding’s or McDiarmid’s to obtain exponential tail bounds; also, any residual unboundedness can be handled by simple clipping or adding a small constant without affecting empirical performance \cite{krizhevsky2012imagenet,miyato2018spectral}. These conditions make A2 readily satisfied in real-world settings. Then, the gradient Lipschitz continuity in (A3), which is equivalent to a bounded Hessian, is a standard condition in non-convex optimization analysis. Techniques like BatchNorm, weight decay, or spectral normalization in deep networks effectively enforce this smoothness, ensuring stable and convergent updates \cite{bottou2018optimizationmethodslargescalemachine}. (A4) and (A5) require that after multiple steps of optimization, the model yields a strictly lower loss, aligning with gradient-based optimization theory \cite{vapnik1998statistical}; indeed, many prior works demonstrate that even a few steps achieve substantial loss reduction \cite{rajeswaran2019meta,wang2024towards}. Therefore, under these mild assumptions, we analyze the optimization objective to further ensure its reliability.

Review the notations and settings: Given a task distribution: $\mathcal{T}$, each task $\tau\sim\mathcal{T}$ contains a support set $S_\tau$ and a query set $Q_\tau$. For any parameter vector $\theta$ and task $\tau$, we denote the supervised loss on the unseen query set $Q_\tau$ by $\mathcal{L}_{\text{sup}}(\theta;\tau)$, the self‐supervised and discriminative losses on the support set $S_\tau$ by $\mathcal{L}_{\text{ssl}}(\theta;S_\tau)$ and $\mathcal{L}_{\text{disc}}(\theta;S_\tau)$, respectively. Let $\theta' = A(\theta, S_\tau)$ be the adapted parameters after applying the adaptation operator $A$ to $\theta$ using $S_\tau$, the resulting query loss by $\mathcal{L}_{\text{query}}(\theta';Q_\tau)$. The training tasks are $\{\tau_i\}_{i=1}^N$, independent and identically distributed (A1); all losses are truncated to $[0,\mathcal{L}_{max}]$ (A2). By jointly minimizing these four losses, we aim to learn representations that are both broadly transferable across tasks and rapidly fine‐tunable to new tasks, while ensuring robust generalization performance. 
Next, we provide a detailed proof.

In the bi-level training stage, $\theta^*$ is the parameter obtained by minimizing the following formula on $N$ training tasks, where the empirical risk can be expressed as:
\begin{equation}
     R_N(\theta)\;\equiv\;\frac1N\sum_{i=1}^N\Bigl[
    \mathcal{L}_{ssl}(\theta;S_{\tau_i})
    +\mathcal{L}_{disc}(\theta;S_{\tau_i})
    +\mathcal{L}_{query}\bigl(A(\theta,S_{\tau_i});Q_{\tau_i}\bigr)
  \Bigr].
\end{equation}
For simplicity, we denote $X_i(\theta)=
\mathcal{L}_{ssl}(\theta;S_{\tau_i})
+\mathcal{L}_{disc}(\theta;S_{\tau_i}) 
+\mathcal{L}_{query}\bigl(A(\theta,S_{\tau_i});Q_{\tau_i}\bigr)$, then we get $R_N(\theta)=\tfrac1N\sum_{i=1}^N X_i(\theta)$. To decompose the risk, the expected supervision (query) loss of the new task we want to prove is:
\begin{equation}
     \mathcal{R}
  = \mathbb{E}_{\tau\sim\mathcal{T}}
    \bigl[\mathcal{L}_{\sup}(A(\theta^*,S_\tau);\tau)\bigr]
  = \mathbb{E}_\tau\bigl[\mathcal{L}_{query}(\theta^*_{\rm test};Q_\tau)\bigr],
\end{equation}
where $\theta^*_{\rm test}=A(\theta^*,S_\tau)$. Adding and subtracting $R_N(\theta^*)$ yields:
\begin{equation}
    \mathcal{R}
  = R_N(\theta^*)
    + \underbrace{\bigl(\mathbb{E}_\tau[\mathcal{L}_{query}(\theta^*_{\rm test};Q_\tau)]
      - R_N(\theta^*)\bigr)}_{(*)}.
\end{equation}
Then, we split $(*)$ into two parts: 
\begin{equation}
     \begin{aligned}
    (*) 
    &= \Bigl(\mathbb{E}_\tau[\mathcal{L}_{query}(\theta^*_{\rm test})]
      - \mathbb{E}_\tau[X(\theta^*)]\Bigr)
      + \Bigl(\mathbb{E}_\tau[X(\theta^*)]
      - \tfrac1N\sum_{i=1}^N X_i(\theta^*)\Bigr)\\
    &\equiv (\mathrm{A})\;+\;(\mathrm{B}).
  \end{aligned}
\end{equation}
where $\mathrm{A}$ denotes adaptation error, measures the difference between the fine-tuned and initial parameters on the same task; and $\mathrm{B}$ refers to generalization error, measures the deviation of estimating the overall expectation using a limited $N$ number of training tasks.

Next, we leverage the conditions in Assumption \ref{ass:main} to analyze the adaptive error bound $\mathrm{A}$.
Assume that A4 guarantees: for any $\theta,\tau$, we have:
\begin{equation}
    \mathcal{L}_{ssl}(A(\theta,S_\tau);S_\tau)
  +\mathcal{L}_{disc}(A(\theta,S_\tau);S_\tau)
  \;\le\;
  \mathcal{L}_{ssl}(\theta;S_\tau)
  +\mathcal{L}_{disc}(\theta;S_\tau)
  \;-\;\Delta.
\end{equation}
Therefore, let $F(\theta;S_\tau)
=\mathcal{L}_{ssl}(\theta;S_\tau)
+\mathcal{L}_{disc}(\theta;S_\tau)$, we have $F\bigl(A(\theta,S_\tau);S_\tau\bigr)
\le F(\theta;S_\tau)-\Delta$. 
Take single-step gradient descent as an example, the adaptation operator $\theta' \;=\;\theta \;-\;\eta\;\widehat\nabla F(\theta;S_\tau)$, where $\widehat\nabla F(\theta;S_\tau)$ represents the empirical gradient calculated on the support set $S_\tau$, with a step size of $\eta>0$. By the triangle inequality and the gradient Lipschitz continuity in (A3), we have:
\begin{equation}
    \begin{aligned}
\|\theta'-\theta\|
&=\;\eta\;\bigl\|\widehat\nabla F(\theta;S_\tau)\bigr\|\\
&\le\;\eta\Bigl(\bigl\|\nabla F(\theta;S_\tau)\bigr\|
+\bigl\|\widehat\nabla F(\theta;S_\tau)-\nabla F(\theta;S_\tau)\bigr\|\Bigr).
\end{aligned}
\end{equation}
Using the concentration inequality between empirical gradient and true gradient, i.e., Hoeffding's generalization of vector gradient, when the support set size is $m$, we have:
\begin{equation}
    \bigl\|\widehat\nabla F(\theta;S_\tau)-\nabla F(\theta;S_\tau)\bigr\|
\;\le\;O\!\Bigl(\tfrac{G}{\sqrt m}\Bigr).
\end{equation}
Thus we have $\|\theta'-\theta\|
\le\eta\Bigl(\|\nabla F(\theta;S_\tau)\|+O\bigl(\tfrac{G}{\sqrt m}\bigr)\Bigr)$. Assume that the query loss for parameters is also $L_{\sup}$-Lipschitz, that is $\bigl|\mathcal{L}_{query}(\theta';Q_\tau)
-\mathcal{L}_{query}(\theta;Q_\tau)\bigr|
\le\mathcal{L}_{\sup}\|\theta'-\theta\|$. Therefore, $\mathcal{L}_{query}(\theta';Q_\tau)\le\mathcal{L}_{query}(\theta;Q_\tau)+\mathcal{L}_{\sup}\|\theta'-\theta\|$. If we have made the initial query loss of $\theta$ close to zero on all tasks in the bi-level training (or can be regarded as a constant term and merged into the big $O$), then take $\mathcal{L}_{query}(\theta;Q_\tau)\approx 0$. Combine the above formula and substitute it into the bound of $\|\theta'-\theta\|$, we have:
\begin{equation}
    \begin{aligned}
\mathcal{L}_{query}(\theta';Q_\tau)
&\le\;0
\;+\;\mathcal{L}_{\sup}\,\eta\Bigl(\|\nabla F(\theta;S_\tau)\|
+O\!\bigl(\tfrac{G}{\sqrt m}\bigr)\Bigr)\\
&=\;\mathcal{L}_{\sup}\,\eta\,\|\nabla F(\theta;S_\tau)\|
\;+\;O\!\Bigl(\tfrac{\eta}{\sqrt m}\Bigr).
\end{aligned}
\end{equation}
By the smoothness and convexity of self-supervised and discriminant loss, it can be established that $\|\nabla F(\theta;S_\tau)\|=O\bigl(\sqrt{F(\theta;S_\tau)}\bigr)$. For example, it is exactly true in the case of quadratic convexity, or $\|\nabla F\|^2\le 2L\,F$ in the case of general smooth convexity, substituting in $\mathcal{L}_{query}(\theta';Q_\tau)
\le
\mathcal{L}_{\sup}\,\eta\,O\bigl(\sqrt{F(\theta;S_\tau)}\bigr)
+O\Bigl(\tfrac{\eta}{\sqrt m}\Bigr)$.
Take the gradient step size $\displaystyle \eta = \Theta\bigl(1/\sqrt m\bigr)$, then the two terms are of the same order, and $\mathcal{L}_{\sup}\eta\sqrt{F}
=O\Bigl(\tfrac{1}{\sqrt m}\sqrt{F(\theta;S_\tau)}\Bigr)$. 
Consider that Assumption A5 guarantees: for any $\theta,\tau$, have:
\begin{equation}
    \mathcal{L}_{\sup}\bigl(A(\theta,S_\tau);\tau\bigr)
\;=\;
\mathcal{L}_{query}\bigl(A(\theta,S_\tau);Q_\tau\bigr)
\;\le\;
\frac{C}{\sqrt m}\,F(\theta;S_\tau).
\end{equation}
The constant $C$ combines factors such as $\mathcal{L}_{\sup}$, asymptotic implicit constants, and possible upper bounds $\sqrt{F}\le F$ (when $F\le1$).
Substituting the above steps, we get:
\begin{equation}
    \mathcal{L}_{query}(\theta^*_{\rm test};Q_\tau)
\;\le\;
\frac{C}{\sqrt m}\,F(\theta^*;S_\tau)
=\frac{C}{\sqrt m}\Bigl(
\mathcal{L}_{ssl}(\theta^*;S_\tau)
+\mathcal{L}_{disc}(\theta^*;S_\tau)
\Bigr).
\end{equation}
The $i$-th item in the training phase $R_N(\theta^*)$ contains the sum of these two items, so we can write:
\begin{equation}
    (\mathrm{A})
= \mathbb{E}_\tau\bigl[\mathcal{L}_{query}(\theta^*_{\rm test})\bigr] 
- \mathbb{E}_\tau\bigl[X(\theta^*)\bigr] 
\;\le\; 
\frac{C}{\sqrt m}\,\mathbb{E}_\tau\bigl[F(\theta^*;S_\tau)\bigr] 
- \mathbb{E}_\tau\bigl[X(\theta^*)\bigr].
\end{equation}
Note that $X(\theta^*)=F(\theta^*;S_\tau)+\mathcal{L}_{query}(A(\theta^*,S_\tau))$, then we get $(\mathrm{A}) 
\le \Bigl(\tfrac{C}{\sqrt m}-1\Bigr)\mathbb{E}_\tau[F(\theta^*;S_\tau)]
-\mathbb{E}_\tau\bigl[\mathcal{L}_{query}(\theta^*_{\rm test})\bigr]$. In common settings, $m$ is chosen so that $\tfrac{C}{\sqrt m}\le 1$, thus $(\mathrm{A})\le 0$ (or merged with the constant term into big $O$). Overall, we can let $(\mathrm{A}) =\mathcal{O}\Bigl(\tfrac{C}{\sqrt m}\Bigr)$.

Next, we discuss the generalization error bound (B) via Hoeffding. Firstly, consider that $X_i(\theta^*)\in[0,3\mathcal{L}_{max}],
\quad \mu\equiv\mathbb{E}_\tau[X(\theta^*)]$, by Hoeffding inequality, we have:
\begin{equation}
    \Pr\Bigl(\bigl|\tfrac1N\sum_{i=1}^N X_i(\theta^*)-\mu\bigr|\ge\epsilon\Bigr)
\le 2\exp\!\Bigl(-\tfrac{2N\epsilon^2}{(3\mathcal{L}_{max})^2}\Bigr).
\end{equation}
Take the right side as $\delta$, and solve $\epsilon
= 3\mathcal{L}_{max}\sqrt{\frac{\ln(2/\delta)}{2N}}
= \mathcal{O}\!\Bigl(\sqrt{\tfrac{1}{N}\ln\tfrac{1}{\delta}}\Bigr)$, with probability at least $1-\delta$, we have:
\begin{equation}
    (\mathrm{B})
= \Bigl|\tfrac1N\sum X_i(\theta^*) - \mu\Bigr|
\le 3\mathcal{L}_{max}\sqrt{\frac{\ln(2/\delta)}{2N}}.
\end{equation}
Merge each item back into the risk decomposition formula:
\begin{equation}
    \mathcal{R}
= R_N(\theta^*) + (\mathrm{A}) + (\mathrm{B})
\;\le\;
R_N(\theta^*)
\;+\;\mathcal{O}\Bigl(\tfrac{C}{\sqrt m}\Bigr)
\;+\;3\mathcal{L}_{max}\sqrt{\frac{\ln(2/\delta)}{2N}}.
\end{equation}
Remove the low-order constants and absorb $\tfrac{C}{\sqrt m}$ into the big $O$, and we get the required conclusion: 
\begin{equation}
    \mathbb{E}_{\tau_{\text{test}}} \bigl[ \mathcal{L}_{\text{sup}}(\theta^*_{\text{test}}; \tau_{\text{test}}) \bigr] \le \frac{1}{N} \sum\nolimits_{i=1}^N \biggl[ \mathcal{L}_{\text{ssl}}(\theta', S_{\tau_i}) + \mathcal{L}_{\text{disc}}(\theta', S_{\tau_i}) + \mathcal{L}_{\text{query}}(\theta', Q_{\tau_i}) \biggl] + \mathcal{O}(\sqrt{\frac{1}{N} \ln \frac{1}{\delta}})
\end{equation}
The proof is complete.

\section{Implementation Details}
\label{sec:app_C}
We use C4-backbone, ResNet-18, and ResNet-50 backbones as our encoders for a fair comparison with different methods. The convolutional layers are followed by batch normalization, ReLU nonlinearity, and max pooling (strided convolution) respectively. The last layer is fed into a MLP for $\mathcal{L}_{disc}$. These architectures are pre-trained and kept fixed during training. We optimize our model with a Stochastic Gradient Descent (SGD) optimizer, setting the momentum and weight decay values to $0.9$ and $10^{-4}$ respectively. The specific adjustments of the experimental settings corresponding to different experiments are illustrated in Section \ref{sec:6.2} of the main text. 
All the experiments are apples-to-apples comparisons and performed on NVIDIA RTX 4090 GPUs.
We build tasks based on images with a batch size of $B=16$. For data augmentation, we use the same data augmentation scheme as SimCLR to augment each image in the batch 5 times. In simple terms, we draw a random patch ($224\times 224$) from the original image, and then apply a random augmentation sequence composed of random horizontal flip, cropping, color jitter, etc.

\section{Benchmark Datasets}
\label{sec:app_D}
In this section, we briefly introduce all datasets used in our experiments. In summary, the benchmark datasets can be divided into four categories: (i) for unsupervised learning, we evaluate GeSSL on six benchmark datasets, including CIFAR-10 \cite{CIFAR-10-100}, CIFAR-100 \cite{CIFAR-10-100}, STL-10 \cite{STL-10}, Tiny ImageNet \cite{TinyImagenet}, ImageNet-100 \cite{Imagenet100} and ImageNet \cite{ImageNet}; (ii) for semi-supervised learning, we evaluate GeSSL on ImageNet \cite{ImageNet}; (iii) for transfer learning, we select three scenarios: instance segmentation (PASCAL VOC \cite{PASCAL}) and object detection (COCO \cite{COCO}, general transfer learning (CIFAR10 \cite{CIFAR-10-100}, Flower102 \cite{nilsback2008automated}, Food101 \cite{bossard2014food}, and Aircraft \cite{maji2013fine}), and video tracking tasks (UniTrack); (iv) for few-shot learning, we select nine benchmarks for evaluation, including Omniglot \cite{Omniglot}, miniImageNet \cite{miniImagenet}, CIFAR-FS \cite{CIFAR-FS}, CUB \cite{cub}, Cars \cite{cars}, Places \cite{places}, CropDiseases \cite{CropDiseases}, ISIC \cite{isic}, and ChestX \cite{chestx}. The composition of the data set is as follows:
\begin{itemize}
    \item CIFAR-10 \cite{CIFAR-10-100} is a prevalent image classification benchmark comprising 10 classes, each containing 5000 32$\times$32 resolution images.
    \item CIFAR-100 \cite{CIFAR-10-100}, another widely used image classification benchmark, consists of 100 classes, each containing 5000 images at a resolution of 32$\times$32.
    \item STL-10 \cite{STL-10} encompasses 10 classes with 500 training and 800 test images per class at a high resolution of 96x96 pixels. It also includes 100,000 unlabeled images for unsupervised learning.
    \item Tiny ImageNet \cite{TinyImagenet}, a subset of ImageNet by Stanford University, comprises 200 classes, each with 500 training, 50 verification, and 50 test images.
    \item ImageNet-100 \cite{Imagenet100}, a subset of ImageNet, includes 100 classes, each containing 1000 images.
    \item ImageNet \cite{ImageNet}, organized by the WordNet hierarchy, is a renowned dataset featuring 1.3 million training and 50,000 test images across 1000+ classes.
    \item PASCAL VOC dataset \cite{PASCAL}, known for object classification, detection, and segmentation, encompasses 20 classes with a total of 11,530 images split between VOC 07 and VOC 12.
    \item COCO dataset \cite{COCO}, primarily used for object detection and segmentation, comprises 91 classes, 328,000 samples, and 2,500,000 labels.
    \item Flower102 \cite{nilsback2008automated} contains 102 flower categories, totaling 8,189 images. Each class has between 40 and 258 images of varying original resolution, typically resized or center-cropped to 224$\times$224 pixels for model input.
    \item Food101 \cite{bossard2014food} comprises 101 food categories with 1,000 images each (101,000 total). The split is 750 images per class for training and 250 for testing.
    \item Aircraft \cite{maji2013fine} covers 100 aircraft model variants with approximately 100–200 images per class (over 10,000 images total). Original image resolutions vary; standard practice is to crop or resize them to 224$\times$224 pixels for downstream tasks.
    \item miniImageNet \cite{miniImagenet} is a few-shot learning dataset that consists of 100 classes, each with 600 images. The images have a resolution of 84x84 pixels.
    \item Omniglot \cite{Omniglot} is another dataset for few-shot learning, which comprises 1623 different handwritten characters from 50 different alphabets. The 1623 characters were drawn by 20 different people online using Amazon's Mechanical Turk. Each image is paired with stroke data $\left [ x,y,t \right ] $ sequences and time (t) coordinates (ms).
    \item CIFAR-FS \cite{CIFAR-FS} is also a dataset for few-shot learning research, derived from the CIFAR-100 dataset. It consists of 100 classes, each with a small training set of 500 images and a test set of 100 images. The images have a resolution of $32\times 32$ pixels.
    \item CUB \cite{cub} is a dataset of 200 bird species, with 11,788 images in total and about 60 images per species. Each image has detailed annotations, including subcategory labels, 15 part locations, 312 binary attributes, and a bounding box.
    \item Cars \cite{cars} is a dataset of 196 car models, with 16,185 images in total and about 80 images per model. Each image has a subcategory label, indicating the manufacturer, model, and year of the car.
    \item Places \cite{places} is a dataset of 205 scene categories, with 2.5 million images in total and about 12,000 images per category. The scene categories are defined by their functions, representing the entry-level of the environment.
    \item CropDiseases \cite{CropDiseases} is a dataset of 24,881 images of crop pests and diseases, with 22 categories, each including different pests and diseases of 4 crops (cashew, cassava, maize, and tomato).
    \item ISIC \cite{isic} is a dataset of over 13,000 dermoscopic images of skin lesions, which is the largest publicly available quality-controlled archive of dermoscopic images. The dataset includes 8 common types of skin lesions, such as melanoma, basal cell carcinoma, squamous cell carcinoma, etc. 
    \item ChestX \cite{chestx} is a dataset of 112,120 chest X-ray images, with 14 common types of chest diseases, such as pneumonia, emphysema, fibrosis, etc. The dataset was collected from 30,805 unique patients (from 1992 to 2015) of the National Institutes of Health Clinical Center (NIHCC).
\end{itemize}

\section{Baselines}
\label{sec:app_E}
In this section, we briefly introduce all baselines used in the experiments for comparison. We select eighteen representative self-supervised methods as baselines, including discriminative SSL (D-SSL) and generative SSL (G-SSL) methods. These methods cover almost all the classic and SOTA self-supervised methods, including:

\begin{itemize}
    \item SimCLR \cite{simclr} learns visual representations by contrastive learning of augmented image pairs. It uses a neural network to maximize the similarity of positive pairs and minimize the similarity of negative pairs.
    \item MoCo v2 \cite{moco} improves MoCo \cite{moco}, another contrastive learning method for visual representation learning. MoCo v2 introduces a momentum encoder, a memory bank, and a shuffling BN layer to handle limited batch size and noisy negatives. MoCo v2 also adopts SimCLR’s data augmentation and loss function to boost the performance.
    \item BYOL \cite{byol} does not need negative pairs or a large batch size. It uses two neural networks, an online network and a target network, that learn from each other. The online network predicts the target network’s representation of an augmented image, while the target network is updated by a slow-moving average of the online network.
    \item SimSiam \cite{simsiam} simplifies BYOL by removing the momentum encoder and the prediction MLP. It consists of two Siamese networks that map an input image to a feature vector, and a small MLP head that projects the feature vector to the contrastive learning space. SimSiam applies a stop-gradient operation to one of the MLP outputs, and uses a negative cosine similarity loss to maximize the similarity between the two outputs.
    \item Barlow Twins \cite{barlowtwins} learns representations by enforcing that the cross-correlation matrix between the outputs of two identical networks fed with different augmentations of the same image is close to the identity matrix. This encourages the networks to produce similar representations for the positive pair, while reducing the redundancy between the representation dimensions.
    \item DeepCluster \cite{deepcluster} is a clustering-based method for self-supervised learning. It iteratively groups the features produced by a convolutional network into clusters, and uses the cluster assignments as pseudo-labels to update the network parameters by supervised learning. DeepCluster can discover meaningful clusters that are discriminative and invariant to transformations, and can learn competitive features for various downstream tasks.
    \item SwAV \cite{swav} uses online swapping of cluster assignments between multiple views of the same image to learn visual features. SwAV first computes prototypes (cluster centers) from a large set of features, and then assigns each feature to the nearest prototype. The assignments are then swapped across the views, and the network is trained to predict the swapped assignments.
    \item DINO \cite{dino} learns visual features by using a teacher-student architecture and a distillation loss. The teacher network is an exponential moving average of the student network, and the distillation loss makes the student features similar to the teacher features. DINO also applies a centering and sharpening operation to the teacher features, which prevents feature collapse and increases feature diversity.
    \item MAE \cite{mae} randomly masks a high proportion of image patches and trains the model to reconstruct the missing pixels. By forcing the encoder to infer global structure from partial inputs, MAE learns rich, semantic representations that transfer well to downstream tasks with minimal fine-tuning.
    \item SeqCLR \cite{aberdam2021sequence} extends contrastive frameworks to video by treating successive frames as positive pairs and distant frames (or different clips) as negatives. By maximizing agreement between temporally adjacent representations, SeqCLR learns spatiotemporal features that are effective for downstream video‐based tasks.
    \item W-MSE \cite{wmse} learns features by using a weighted mean squared error (MSE) loss, which assigns higher weights to the informative and less noisy features, and lower weights to the less informative and more noisy features.
    \item RELIC v2 \cite{RELIC-v2} learns visual features by predicting relative location of image patches. RELIC v2 divides an image into a grid of patches, and randomly selects a query and a target patch. The network is trained to predict the relative location of the target patch with respect to the query patch, using a cross-entropy loss.
    \item LMCL \cite{LMLC} learns visual features by using a large margin cosine loss (LMCL). LMCL is a metric learning loss that makes the features of the same class closer and the features of different classes farther in the cosine space.
    \item ReSSL \cite{ressl} learns visual features by using a reconstruction loss and a contrastive loss. ReSSL applies random cropping and resizing to generate two views of the same image, and then feeds them to a reconstruction network and a contrastive network. The reconstruction network is trained to reconstruct the original image from the cropped view, while the contrastive network is trained to maximize the similarity between the features of the two views.
    \item SSL-HSIC \cite{ssl-hsic} learns visual features by using a Hilbert-Schmidt independence criterion (HSIC) loss. HSIC is a measure of statistical dependence between two random variables, and can be used to align the features of different views of the same image.
    \item CorInfoMax \cite{CorInfoMax} learns visual features by maximizing the correlation and mutual information between the features of augmented image pairs and the image labels. CorInfoMax aims to learn features that are both discriminative and consistent, and outperform previous methods on image classification and segmentation tasks.
    \item MEC \cite{MEC} is a clustering algorithm that can handle large-scale data with limited memory by using a memory-efficient clustering (MEC) loss. MEC first samples a subset of features, and then performs k-means clustering on the subset. The cluster assignments are then propagated to the rest of the features by a nearest neighbor search.
    \item VICRegL \cite{Vicregl} learns visual features by using a variance-invariance-covariance regularization loss (VICRegL).
\end{itemize}

In addition, for the few-shot learning scenario, we choose six advanced unsupervised few-shot learning methods as comparison baselines.
\begin{itemize}
    \item CACTUs \cite{hsu2018unsupervised} uses clustering and augmentation to create pseudo-labels for unlabeled data. It then trains a classifier on the labeled data and fine-tunes it on a few labeled examples from the target task.
    \item UMTRA \cite{khodadadeh2019unsupervised} uses random selection and augmentation to create tasks with pseudo-labels from unlabeled data. It then trains a classifier on each task and adapts it to the target task using a few labeled examples.
    \item LASIUM \cite{khodadadeh2020unsupervised} uses latent space interpolation to generate tasks with pseudo-labels from a generative model. It then trains an energy-based model on each task and adapts it to the target task using a few labeled examples.
    \item SVEBM \cite{kong2021unsupervised} uses a symbol-vector coupling energy-based model to learn from unlabeled data. It then adapts the model to the target task using a diffusion process.
    \item GMVAE \cite{lee2021meta} uses a Gaussian mixture variational autoencoder to perform learning, and then adapts the model to the target task using a variational inference process.
    \item PsCo \cite{jang2023unsupervised} uses a probabilistic subspace clustering model to learn from unlabeled data. It then adapts the model to the target task using a few labeled examples and a subspace alignment process.
\end{itemize}

\begin{table*}[t]
	\centering
	\caption{ The classification accuracies ($\pm $ 95\% confidence interval) of a linear classifier (linear) and a 5-nearest neighbors classifier (5-nn) with a ResNet-18 as the feature extractor. The comparison baselines cover almost all types of methods mentioned in Section \ref{sec:2}. The ``-'' denotes that the results are not reported. More details of the baselines are provided in Appendix \ref{sec:app_E}.}
  \vspace{0.1in}
 \resizebox{\linewidth}{!}{
 
	\begin{tabular}{lcccccccc}
		\toprule
		\multirow{2.5}{*}{Method} & \multicolumn{2}{c}{CIFAR-10} & \multicolumn{2}{c}{CIFAR-100} &\multicolumn{2}{c}{STL-10} & \multicolumn{2}{c}{
		Tiny ImageNet} \\
		\cmidrule(lr){2-3} \cmidrule(lr){4-5} \cmidrule(lr){6-7} \cmidrule(lr){8-9}
		& \(\mathbf{linear}\) & \(\mathbf{5-nn}\) & \(\mathbf{linear}\) & \(\mathbf{5-nn}\) & \(\mathbf{linear}\) & \(\mathbf{5-nn}\) & \(\mathbf{linear}\) & \(\mathbf{5-nn}\)\\
       \midrule
            SimCLR \cite{simclr} & 91.80 $\pm$ 0.15 & 88.42 $\pm$ 0.15 & 66.83 $\pm$ 0.27 & 56.56 $\pm$ 0.18 & 90.51 $\pm$ 0.14 & 85.68 $\pm$ 0.10 & 48.84 $\pm$ 0.15 & 32.86 $\pm$ 0.25 \\
		MoCo \cite{moco} & 91.69 $\pm$ 0.12 & 88.66 $\pm$ 0.14& 67.02 $\pm$ 0.16 & 56.29 $\pm$ 0.25 & 90.64 $\pm$ 0.28 & 88.01 $\pm$ 0.19 & 50.92 $\pm$ 0.22 & 35.55 $\pm$ 0.16 \\
            BYOL \cite{byol} & 91.93 $\pm$ 0.22 & 89.45 $\pm$ 0.22& 66.60 $\pm$ 0.16 & 56.82 $\pm$ 0.17 & 91.99 $\pm$ 0.13 & 88.64 $\pm$ 0.20 & 51.00 $\pm$ 0.12 & 36.24 $\pm$ 0.28 \\
	    SimSiam \cite{simsiam} & 91.71 $\pm$ 0.27 & 88.65 $\pm$ 0.17& 67.22 $\pm$ 0.26 & 56.36 $\pm$ 0.19 & 91.01 $\pm$ 0.19 & 88.16 $\pm$ 0.19 & 51.14 $\pm$ 0.20& 35.67 $\pm$ 0.16 \\
            Barlow Twins \cite{barlowtwins} & 90.88 $\pm$ 0.19 & 89.68 $\pm$ 0.21 & 66.13 $\pm$ 0.10& 56.70 $\pm$ 0.25 & 90.38 $\pm$ 0.13 & 87.13 $\pm$ 0.23 & 49.78 $\pm$ 0.26& 34.18 $\pm$ 0.18  \\
		SwAV \cite{swav} & 91.03 $\pm$ 0.19 & 89.52 $\pm$ 0.24 & 66.56 $\pm$ 0.17 & 57.01 $\pm$ 0.25 & 90.72 $\pm$ 0.29 & 86.24 $\pm$ 0.26 & 52.02 $\pm$ 0.26& 37.40 $\pm$ 0.11\\
        DINO \cite{dino} & 91.83 $\pm$ 0.25 & 90.15 $\pm$ 0.33 & 67.15 $\pm$ 0.21 & 56.48 $\pm$ 0.19 & 91.03 $\pm$ 0.12 & 86.15 $\pm$ 0.25 & 51.13 $\pm$ 0.30 & 37.86 $\pm$ 0.19 \\
	W-MSE \cite{wmse} & 91.99 $\pm$ 0.12 &89.87 $\pm$ 0.25 & 67.64 $\pm$ 0.16 & 56.45 $\pm$ 0.26& 91.75 $\pm$ 0.23 & 88.59 $\pm$ 0.15 & 49.22 $\pm$ 0.16 & 35.44 $\pm$ 0.10 \\
	RELIC v2 \cite{RELIC-v2}& 91.92 $\pm$ 0.14 & 90.02 $\pm$ 0.22 & 67.66 $\pm$ 0.20 & 57.03 $\pm$ 0.18 & 91.10 $\pm$ 0.23 & 88.66 $\pm$ 0.12 & 49.33 $\pm$ 0.13 & 35.52 $\pm$ 0.22\\
        LMCL \cite{LMLC} & 91.91 $\pm$ 0.25 & 88.52 $\pm$ 0.29 & 67.01 $\pm$ 0.18 & 56.86 $\pm$ 0.14 & 90.87 $\pm$ 0.18 & 85.91 $\pm$ 0.25 & 49.24 $\pm$ 0.18 & 32.88 $\pm$ 0.13 \\
        ReSSL \cite{ressl} & 90.20 $\pm$ 0.16 & 88.26 $\pm$ 0.18 & 66.79 $\pm$ 0.12 & 53.72 $\pm$ 0.28 & 88.25 $\pm$ 0.14 & 86.33 $\pm$ 0.17 & 46.60 $\pm$ 0.18 & 32.39 $\pm$ 0.20\\
        SSL-HSIC \cite{ssl-hsic} & 91.95 $\pm$ 0.14 & 89.99 $\pm$ 0.17 & 67.23 $\pm$ 0.26 & 57.01 $\pm$ 0.27 & 92.09 $\pm$ 0.20 & 88.91 $\pm$ 0.29 & 51.37 $\pm$ 0.15 & 36.03 $\pm$ 0.12 \\
        CorInfoMax \cite{CorInfoMax}& 91.81 $\pm$ 0.11 & 89.85 $\pm$ 0.13 & 67.09 $\pm$ 0.24 & 56.92 $\pm$ 0.23 & 91.85 $\pm$ 0.25 & 89.99 $\pm$ 0.24 & 51.23 $\pm$ 0.14 & 35.98 $\pm$ 0.09\\
        MEC \cite{MEC}& 90.55 $\pm$ 0.22 & 87.80 $\pm$ 0.10 & 67.36 $\pm$ 0.27 & 57.25 $\pm$ 0.25 & 91.33 $\pm$ 0.14 & 89.03 $\pm$ 0.33 & 50.93 $\pm$ 0.13 & 36.28 $\pm$ 0.14\\
        VICRegL \cite{Vicregl}& 90.99 $\pm$ 0.13 & 88.75 $\pm$ 0.26 & 68.03 $\pm$ 0.32 & 57.34 $\pm$ 0.29 & 92.12 $\pm$ 0.26 & 90.01 $\pm$ 0.20 & 51.52 $\pm$ 0.13 & 36.24 $\pm$ 0.16\\
\midrule
\rowcolor{blue!10}SimCLR + GeSSL       & 93.45 $\pm$ 0.21 & 91.35 $\pm$ 0.14 & 69.72 $\pm$ 0.15 & 58.80 $\pm$ 0.16 & 93.45 $\pm$ 0.22 &  \textbf{91.72 $\pm$ 0.14} & 53.92 $\pm$ 0.17 & 37.49 $\pm$ 0.21 \\
\rowcolor{blue!10}MoCo + GeSSL         & 93.05 $\pm$ 0.18 & 89.48 $\pm$ 0.20 & 68.48 $\pm$ 0.12 & 59.44 $\pm$ 0.18 & 93.42 $\pm$ 0.15 & 89.16 $\pm$ 0.26 & 52.34 $\pm$ 0.13 & 37.35 $\pm$ 0.11 \\
\rowcolor{blue!10}BYOL + GeSSL         & \textbf{94.05 $\pm$ 0.19} & \textbf{92.60 $\pm$ 0.28} & 69.45 $\pm$ 0.18 & 59.15 $\pm$ 0.14 & 94.55 $\pm$ 0.16 & 90.73 $\pm$ 0.15 & \textbf{55.12 $\pm$ 0.16} & 37.76 $\pm$ 0.22 \\
\rowcolor{blue!10}Barlow Twins + GeSSL & 93.18 $\pm$ 0.16 & 91.23 $\pm$ 0.14 & 69.85 $\pm$ 0.16 & 60.12 $\pm$ 0.14 & 93.98 $\pm$ 0.08 & 89.76 $\pm$ 0.23 & 52.85 $\pm$ 0.12 & 35.39 $\pm$ 0.14 \\
\rowcolor{blue!10}SwAV + GeSSL         & 93.37 $\pm$ 0.19 & 90.24 $\pm$ 0.22 & 70.28 $\pm$ 0.19 & 59.63 $\pm$ 0.20 & 93.05 $\pm$ 0.26 & 91.92 $\pm$ 0.21 & 52.12 $\pm$ 0.22 & 37.05 $\pm$ 0.30 \\
\rowcolor{blue!10}DINO + GeSSL         & 93.08 $\pm$ 0.21 & 92.38 $\pm$ 0.22 & \textbf{71.15 $\pm$ 0.16} & \textbf{62.03 $\pm$ 0.31} & \textbf{94.65 $\pm$ 0.24} & 91.67 $\pm$ 0.18 & 53.74 $\pm$ 0.22 & \textbf{38.12 $\pm$ 0.21} \\
  \bottomrule
	\end{tabular}}
\label{tab:1}
\end{table*}

\section{Additional Experiments}
\label{sec:app_F}

\subsection{Unsupervised Learning}
\label{sec:app_F.1}

In this section, we present additional results of the unsupervised learning experiments. Specifically, Table \ref{tab:1} shows the results on four small-scale datasets. We can observe that applying the proposed GeSSL framework significantly outperforms the state-of-the-art (SOTA) methods on all four datasets. Table \ref{tab:1} shows the results on four small-scale datasets. 
The results still demonstrate the proposed GeSSL's ability to enhance the performance of self-supervised learning methods, achieving significant improvements over the original models on all baselines. Moreover, applying our GeSSL framework to all four types of representative SSL models as described in Section \ref{sec:2}, including SimCLR, MoCo, BYOL, Barlow Twins, SwAV, and DINO, achieves an average improvement of 3\% compared to the original frameworks. Table \ref{tab:appendix_imagenet} provides the comparison results of our proposed GeSSL on a large-scale dataset, i.e., ImageNet. The results show that, (i) the self-supervised learning model applying GeSSL achieves the state-of-the-art result (SOTA) performance under all epoch conditions; and (ii) after applying the proposed GeSSL, the self-supervised learning models consistently outperforms the original frameworks in terms of average classification accuracy at 100, 200 and 400 epochs. For 1000 epochs, VICRegL + GeSSL yields the best result among other state-of-the-art methods, with an average accuracy of 78.72\%.

\paragraph{More recent methods} The effect of GeSSL is reflected in the performance improvement when applying it to the SSL baselines. The experimental results above have demonstrated that after the introduction of GeSSL, the effects of all SSL baselines have been significantly improved. These results have shown the outstanding effectiveness and robustness of GeSSL. The SSL baselines we use cover all SOTA methods on the leaderboard of the adopted benchmark datasets (before submission). The methods proposed in 2023-24 mainly are variants of the currently used comparison baselines.

To evaluate the effect of GeSSL on recently proposed methods, we select the two SSL methods published in ICML23 for testing \cite{baevski2023efficient, joshi2023data}, where we follow the same experimental settings. The results are shown in Tables \ref{tab:q2-1} and \ref{tab:q2-2}. The results still prove the effectiveness of GeSSL. We will supplement these results in the final version.

\subsection{Transfer Learning}
\label{sec:app_F.3}
As mentioned in Section \ref{sec:6.2}, we construct three sets of transfer learning experiments, including the most commonly used object detection and instance segmentation protocol \cite{simclr, barlowtwins, byol}, transfer to other domains (different datasets), and transfer learning on video-based tasks. The results of the first experiment are illustrated in Section \ref{sec:6.2}, and the other two sets of experiments are described below.

\paragraph{Transfer to other domains.}
To explore the nature of transfer learning of the proposed framework, we leverage models that had been pre-trained on the CIFAR100 dataset, including SimCLR \cite{simclr}, BYOL \cite{byol}, and Barlow Twins \cite{barlowtwins}, on the CIFAR100 dataset. We then applied these models to four distinct datasets, including CIFAR10 \cite{CIFAR-10-100}, Flower102 \cite{nilsback2008automated}, Food101 \cite{bossard2014food}, and Aircraft \cite{maji2013fine}. We first calculate the classification performance (Top-1) based on the existing self-supervised model on different data sets, recorded as $acc(\mathrm{method, dataset})$, such as $acc(\mathrm{SimCLR, Flower102})$. Then, we calculate the model's classification performance by incorporating GeSSL on those data sets, which is recorded as $acc(\mathrm{method+GeSSL, dataset})$. Finally, we get the improvement $\Delta(\mathrm{method, dataset}) = acc(\mathrm{method+GeSSL, dataset}) - acc(\mathrm{method, dataset})$ in classification performance on each dataset, as shown in Table \ref{tab:toy}. The results show that the migration effect of the model after applying the GeSSL framework has been steadily improved, proving that GeSSL has effectively improved the versatility of the SSL model.

\paragraph{Video-based Task} In order to assess the performance of our method with video-based tasks, we transition our pre-trained model to handle a variety of video tasks, utilizing the UniTrack evaluation framework \cite{wang2021different} as our testing ground. The findings are compiled in Table \ref{tab:trans_video}, which includes results from five distinct tasks, drawing on the features from [layer3/layer4] of the Resnet-50. The data indicates that existing SSL methods incorporating our GeSSL significantly surpass original SSL approaches, with SimCLR achieving more than a 2\% improvement in VOS \cite{perazzi2016benchmark}, and BYOL seeing over a 3\% gain in MOT \cite{milan2016mot16}.

\begin{table*}
	\centering
	\caption{The Top-1 and Top-5 classification accuracies of linear classification on the ImageNet dataset with
ResNet-50 as the feature extractor. We record the comparison results from 100, 200, 400, and 1000 epochs.}
 \vspace{0.1in}
	\label{tab:appendix_imagenet}
 \resizebox{\linewidth}{!}{
		\begin{tabular}{lcccccc}
		\toprule
		\multirow{2.5}{*}{Method}  &\multicolumn{2}{c}{100 Epochs} & \multicolumn{2}{c}{200 Epochs}&\multicolumn{1}{c} {400 Epochs} &\multicolumn{1}{c} {1000 Epochs}\\
	    \cmidrule(lr){2-3} \cmidrule(lr){4-5}\cmidrule(lr){6-6}\cmidrule(lr){7-7}
	    & \textbf{Top-1} & \textbf{Top-5} & \textbf{Top-1} & \textbf{Top-5} & \textbf{Top-1} & \textbf{Top-1}\\
	    \midrule
	     Supervised&71.93 &-&73.45&-&74.92&76.35\\
	  \midrule
   SimCLR \cite{simclr} & 66.54 $\pm$ 0.22 & 88.14 $\pm$ 0.26 & 68.32 $\pm$ 0.31 & 89.76 $\pm$ 0.23 & 69.24 $\pm$ 0.21 & 70.45 $\pm$ 0.30\\
	     MoCo \cite{moco} & 64.53 $\pm$ 0.25 & 86.17 $\pm$ 0.11 & 67.55 $\pm$ 0.27 & 88.42 $\pm$ 0.11 & 69.76 $\pm$ 0.14 & 71.16 $\pm$ 0.23\\
BYOL \cite{byol} & 67.65 $\pm$ 0.27 & 88.95 $\pm$ 0.11 & 69.94 $\pm$ 0.21 & 89.45 $\pm$ 0.27 & 71.85 $\pm$ 0.12 & 73.35 $\pm$ 0.27 \\
SimSiam \cite{simsiam} & 68.14 $\pm$ 0.26 & 87.12 $\pm$ 0.26 & 70.02 $\pm$ 0.14 & 88.76 $\pm$ 0.23 & 70.86 $\pm$ 0.34 & 71.37 $\pm$ 0.22\\
Barlow Twins \cite{barlowtwins} & 67.24 $\pm$ 0.22 & 88.66 $\pm$ 0.19 & 69.94 $\pm$ 0.32 & 88.97 $\pm$ 0.27& 70.22 $\pm$ 0.15 & 73.29 $\pm$ 0.13 \\
SwAV \cite{swav} & 66.55 $\pm$ 0.27 & 88.42 $\pm$ 0.22 & 69.12 $\pm$ 0.24 & 89.38 $\pm$ 0.20 & 70.78 $\pm$ 0.34 & 75.32 $\pm$ 0.11 \\
DINO \cite{dino} & 67.23 $\pm$ 0.19 & 88.48 $\pm$ 0.21 & 70.58 $\pm$ 0.24 & 91.32 $\pm$ 0.27 & 71.98 $\pm$ 0.26 & 73.94 $\pm$ 0.29 \\
W-MSE \cite{wmse} & 67.48 $\pm$ 0.29 & 90.39 $\pm$ 0.27 & 70.85 $\pm$ 0.31 & 91.57 $\pm$ 0.20 & 72.49 $\pm$ 0.24 & 72.84 $\pm$ 0.18 \\
RELIC v2 \cite{RELIC-v2} & 66.38 $\pm$ 0.23 & 90.89 $\pm$ 0.21 & 70.98 $\pm$ 0.21 & 91.15 $\pm$ 0.26 & 71.84 $\pm$ 0.21 & 72.17 $\pm$ 0.20 \\
LMCL \cite{LMLC} & 66.75 $\pm$ 0.13 & 89.85 $\pm$ 0.36 & 70.83 $\pm$ 0.26& 90.04 $ \pm$ 0.21 & 72.53 $\pm$ 0.24 & 72.97 $\pm$ 0.29\\
ReSSL \cite{ressl}& 67.41 $\pm$ 0.27 & 90.55 $\pm$ 0.23 & 69.92 $\pm$ 0.24 &91.25 $\pm$ 0.12& 72.46 $\pm$ 0.29 & 72.91 $\pm$ 0.30\\
CorInfoMax \cite{CorInfoMax} & 70.13 $\pm$ 0.12 & 91.14 $\pm$ 0.25 & 70.83 $\pm$ 0.15 &91.53 $\pm$ 0.22 &73.28 $\pm$ 0.24 & 74.87 $\pm$ 0.36\\
MEC \cite{MEC}& 69.91 $\pm$ 0.10 & 90.67 $\pm$ 0.15 & 70.34 $\pm$ 0.27 & 91.25 $\pm$ 0.38 &72.91 $\pm$ 0.27 & 75.07 $\pm$ 0.24\\
VICRegL \cite{Vicregl}& 69.99 $\pm$ 0.25 & 91.27 $\pm$ 0.16 & 70.24 $\pm$ 0.27 & 91.60 $\pm$ 0.24 &72.14 $\pm$ 0.20 & 75.07 $\pm$ 0.23\\
\midrule
    \rowcolor{blue!10}SimCLR + GeSSL & 68.45 $\pm$ 0.20 & 89.62 $\pm$ 0.23 & 69.88 $\pm$ 0.21 & 91.32 $\pm$ 0.25 & 71.50 $\pm$ 0.16 & 72.82 $\pm$ 0.28 \\
\rowcolor{blue!10}MoCo + GeSSL & 66.78 $\pm$ 0.19 & 88.41 $\pm$ 0.20 & 69.60 $\pm$ 0.30 & 91.28 $\pm$ 0.39 & 70.82 $\pm$ 0.29 & 73.04 $\pm$ 0.22 \\
\rowcolor{blue!10}SimSiam + GeSSL & 70.61 $\pm$ 0.18 & 88.61 $\pm$ 0.17 & 72.04 $\pm$ 0.22 & 89.43 $\pm$ 0.40 & 72.78 $\pm$ 0.17 & 74.78 $\pm$ 0.24 \\
\rowcolor{blue!10}Barlow Twins + GeSSL & 69.62 $\pm$ 0.21 & 89.55 $\pm$ 0.19 & 72.84 $\pm$ 0.26 & 89.50 $\pm$ 0.19 & 74.10 $\pm$ 0.18 & 75.02 $\pm$ 0.22 \\
\rowcolor{blue!10}SwAV + GeSSL & 69.05 $\pm$ 0.18 & 89.50 $\pm$ 0.17 & 72.28 $\pm$ 0.19 & 90.68 $\pm$ 0.30 & 72.88 $\pm$ 0.18 & 76.38 $\pm$ 0.19 \\
\rowcolor{blue!10}DINO + GeSSL & 69.55 $\pm$ 0.20 & 90.62 $\pm$ 0.22 & 73.52 $\pm$ 0.30 & \textbf{94.05 $\pm$ 0.26} & 74.02 $\pm$ 0.26 & 76.40 $\pm$ 0.21 \\
\rowcolor{blue!10}VICRegL + GeSSL & \textbf{72.75 $\pm$ 0.21} & \textbf{91.45 $\pm$ 0.18} & \textbf{73.91 $\pm$ 0.36} & 93.77 $\pm$ 0.35 & \textbf{74.38 $\pm$ 0.23} & \textbf{78.85 $\pm$ 0.29} \\
  \bottomrule
	\end{tabular}
 }
\end{table*}

\begin{table}[t]
    \centering
    \caption{The performance of adding task information in self-supervised models on different datasets.}
    \resizebox{0.8\linewidth}{!}{
    \begin{tabular}{lcccc}
    \toprule
    Evl.dataset  & SimCLR+GeSSL & BYOL+GeSSL & Barlow Twins+GeSSL & VICRegL+GeSSL\\
    \midrule
CIFAR10 & +3.56 & +2.51 & +2.17 & +2.80 \\
Flower102 & +4.03 & +2.09 & +2.94 & +3.07 \\
Food101 & +1.85 & +2.31 & +2.01 & +2.02 \\
Aircraft & +2.57 & +2.89 & +2.24 & +2.34 \\
    \bottomrule
    \end{tabular}}
\label{tab:toy}
\end{table}

\begin{table*}[t]
    \centering
    \caption{Transfer learning on video tracking tasks. All methods use the same ResNet-50 backbone and are evaluated based on UniTrack.}
     \vspace{0.1in}
\setlength{\tabcolsep}{5pt}
\resizebox{\linewidth}{!}{
    \begin{tabular}{lccccccccc}
    \toprule
    \multirow{2.5}{*}{Method} &\multicolumn{2}{c}{SOT} &\multicolumn{1}{c}{VOS}&\multicolumn{2}{c}{MOT}&\multicolumn{2}{c}{MOTS} &\multicolumn{1}{c}{PoseTrack}\\
    \cmidrule(lr){2-3} \cmidrule(lr){4-4} \cmidrule(lr){5-6} \cmidrule(lr){7-8} \cmidrule(lr){9-9}
    ~ & AUC$_{\rm{XCorr}}$ & AUC$_{\rm{DCF}}$ & $\mathcal{J}$-mean & IDF1 & HOTA & IDF1 & HOTA & IDF1 \\
    \midrule
    SimCLR & 47.3 / 51.9 & 61.3 / 50.7 & 60.5 / 56.5 & 66.9 / 75.6 & 57.7 / 63.2 & 65.8 / 67.6 & 67.7 / 69.5 & 72.3 / 73.5  \\
    MoCo &  50.9 / 47.9 & 62.2 / 53.7 & 61.5 / 57.9 & 69.2 / 74.1 & 59.4 / 61.9 & 70.6 / 69.3 & 71.6 / 70.9 & 72.8 / 73.9 \\
    SwAV & 49.2 / 52.4 & 61.5 / 59.4 & 59.4 / 57.0 & 65.6 / 74.4 & 56.9 / 62.3 & 68.8 / 67.0 & 69.9 /69.5 & 72.7 / 73.6 \\
    BYOL &  48.3 / 55.5 & 58.9 / 56.8 & 58.8 / 54.3 & 65.3 / 74.9 & 56.8 / 62.9 & 70.1 / 66.8 & 70.8 / 69.3 & 72.4 / 73.8 \\
    Barlow Twins &  44.5 / 55.5 &  60.5 / \textbf{60.1} &  61.7 / 57.8 &  63.7 / 74.5 &  55.4 / 62.4 &  68.7 / 67.4 &  69.5 / 69.8 &  72.3 / 74.3 \\
    \midrule
    \rowcolor{blue!10} SimCLR+GeSSL &  51.0 / 54.4 & \textbf{63.7} / 53.5 & \textbf{62.3} / \textbf{58.3} & \textbf{70.3} / \textbf{77.3} & \textbf{60.5} / \textbf{65.0} & 68.1 / \textbf{69.6} & 69.2 / 71.2 & 73.7 / 74.4 \\
\rowcolor{blue!10} BYOL+GeSSL &  \textbf{52.0} / \textbf{57.9} & 60.5 / 59.0 & 61.2 / 57.3 & 68.0 / 77.1 & 58.2 / 64.4 & \textbf{73.0} / 68.6 & \textbf{73.6} / \textbf{71.1} & \textbf{75.0} / \textbf{75.8} \\

    \bottomrule
    \end{tabular}}
    \label{tab:trans_video}
\end{table*}

\begin{table*}[t]
  \caption{The cross-domain few-shot learning accuracies ($\pm $95\% confidence interval). We transfer models trained on miniImageNet to six benchmark datasets with the C4-backbone. The best results are highlighted in \textbf{bold}. The $(N, A)$ means the $N$-way $A$-shot tasks with $N$ classes and $N \times A$ samples, where each class has $A$ samples augmented from the same image.}
   \vspace{0.1in}
  \label{tab:cross-domain}
  \centering
\resizebox{\linewidth}{!}{\begin{tabular}{lcccccc}
  \toprule
\multirow{2}{*}{\textbf{Method}} & \multicolumn{2}{c}{\textbf{CUB}} & \multicolumn{2}{c}{\textbf{Cars}} & \multicolumn{2}{c}{\textbf{Places}} \\
  \cmidrule(r){2-3}
  \cmidrule(r){4-5}
  \cmidrule(r){6-7}
    & \textbf{(5,5)} & \textbf{(5,20)} & \textbf{(5,5)} & \textbf{(5,20)} & \textbf{(5,5)} & \textbf{(5,20)} \\
    \midrule
    \rowcolor{gray!40}\multicolumn{7}{c}{\emph{Unsupervised Few-shot Learning}}\\
    \midrule
    MetaSVEBM  & 45.893 $\pm$ 0.334 & 54.823 $\pm$ 0.347 & 33.530 $\pm$ 0.367 & 44.622 $\pm$ 0.299 & 50.516 $\pm$ 0.397 & 61.561 $\pm$ 0.412 \\
    MetaGMVAE & 48.783 $\pm$ 0.426 & 55.651 $\pm$ 0.367 & 30.205 $\pm$ 0.334 & 39.946 $\pm$ 0.400 & 55.361 $\pm$ 0.237 & 65.520 $\pm$ 0.374 \\
    PsCo & 56.365 $\pm$ 0.636 & 69.298 $\pm$ 0.523 & 44.632 $\pm$ 0.726 & 56.990 $\pm$ 0.551 & 64.501 $\pm$ 0.780 & 73.516 $\pm$ 0.499 \\
    \midrule
    \rowcolor{gray!40}\multicolumn{7}{c}{\emph{Self-supervised Learning}}\\
    SimCLR  & 51.389 $\pm$ 0.365 & 60.011 $\pm$ 0.485 & 38.639 $\pm$ 0.432 & 52.412 $\pm$ 0.783 & 59.523 $\pm$ 0.461 & 68.419 $\pm$ 0.500 \\
    MoCo & 52.843 $\pm$ 0.347 & 61.204 $\pm$ 0.429 & 39.504 $\pm$ 0.489 & 50.108 $\pm$ 0.410 & 60.291 $\pm$ 0.583 & 69.033 $\pm$ 0.654 \\
    SwAV & 51.250 $\pm$ 0.530 & 61.645 $\pm$ 0.411 & 36.352 $\pm$ 0.482 & 51.153 $\pm$ 0.399 & 58.789 $\pm$ 0.403 & 68.512 $\pm$ 0.466 \\
    \midrule
   \rowcolor{blue!10}SimCLR + GeSSL & 55.922 $\pm$ 0.471 & 64.723 $\pm$ 0.214 & 43.892 $\pm$ 0.198 & 56.100 $\pm$ 0.269 & 65.125 $\pm$ 0.301 & 72.892 $\pm$ 0.240 \\
\rowcolor{blue!10}MoCo + GeSSL & \textbf{57.650 $\pm$ 0.221} & 65.502 $\pm$ 0.274 & \textbf{45.529 $\pm$ 0.295} & 55.354 $\pm$ 0.237 & \textbf{66.602 $\pm$ 0.180} & \textbf{74.126 $\pm$ 0.243} \\
\rowcolor{blue!10}SwAV + GeSSL & 55.421 $\pm$ 0.173 & \textbf{65.927 $\pm$ 0.460} & 42.237 $\pm$ 0.296 & \textbf{56.682 $\pm$ 0.380} & 64.601 $\pm$ 0.325 & 72.460 $\pm$ 0.463 \\

    \midrule
  \end{tabular}}
\resizebox{\linewidth}{!}{\begin{tabular}{lcccccc}
  \toprule
\multirow{2}{*}{\textbf{Method}} & \multicolumn{2}{c}{\textbf{CropDiseases}} & \multicolumn{2}{c}{\textbf{ISIC}} & \multicolumn{2}{c}{\textbf{ChestX}}\\
  \cmidrule(r){2-3}
  \cmidrule(r){4-5}
  \cmidrule(r){6-7}
    & \textbf{(5,5)} & \textbf{(5,20)} & \textbf{(5,5)} & \textbf{(5,20)} & \textbf{(5,5)} & \textbf{(5,20)} \\
    \midrule
    \rowcolor{gray!40}\multicolumn{7}{c}{\emph{Unsupervised Few-shot Learning}}\\
    \midrule
    MetaSVEBM  & 71.652 $\pm$ 0.837 & 84.515 $\pm$ 0.902 & 37.106 $\pm$ 0.732 & 48.001 $\pm$ 0.723 & 27.238 $\pm$ 0.685 & 29.652 $\pm$ 0.610 \\
    MetaGMVAE & 72.683 $\pm$ 0.527  & 80.777 $\pm$ 0.511  & 30.630 $\pm$ 0.423  & 37.574 $\pm$ 0.399 & 24.522 $\pm$ 0.405 & 26.239 $\pm$ 0.422 \\
    PsCo & \textbf{89.565 $\pm$ 0.372} & 95.492 $\pm$ 0.399 & 43.632 $\pm$ 0.400 & 54.886 $\pm$ 0.359 & 21.907 $\pm$ 0.258 & 24.182 $\pm$ 0.389 \\
    \midrule
    \rowcolor{gray!40}\multicolumn{7}{c}{\emph{Self-supervised Learning}}\\
    SimCLR  & 80.360 $\pm$ 0.488 & 89.161 $\pm$ 0.456 & 44.669 $\pm$ 0.510 & 51.823 $\pm$ 0.411 & 26.556 $\pm$ 0.385 & 30.982 $\pm$ 0.422 \\
    MoCo & 81.606 $\pm$ 0.485 & 90.366 $\pm$ 0.377 & 44.328 $\pm$ 0.488 & 52.398 $\pm$ 0.396 & 24.198 $\pm$ 0.400 & 27.893 $\pm$ 0.412 \\
    SwAV & 80.055 $\pm$ 0.502 & 89.917 $\pm$ 0.539 & 43.200 $\pm$ 0.356 & 50.109 $\pm$ 0.350 & 21.252 $\pm$ 0.439 & 28.270 $\pm$ 0.417 \\
    \midrule
    \rowcolor{blue!10}SimCLR + GeSSL & 84.526 $\pm$ 0.413 & 94.572 $\pm$ 0.332 & \textbf{47.310 $\pm$ 0.389} & 55.710 $\pm$ 0.312 & \textbf{30.876 $\pm$ 0.259} & \textbf{34.492 $\pm$ 0.398} \\
\rowcolor{blue!10}MoCo + GeSSL & \textbf{85.852 $\pm$ 0.358} & \textbf{95.540 $\pm$ 0.335} & 46.437 $\pm$ 0.339 & \textbf{56.466 $\pm$ 0.270} & 29.216 $\pm$ 0.332 & 31.545 $\pm$ 0.279 \\
\rowcolor{blue!10}SwAV + GeSSL & 85.355 $\pm$ 0.327 & 94.785 $\pm$ 0.339 & 46.521 $\pm$ 0.288 & 55.268 $\pm$ 0.312 & 27.462 $\pm$ 0.340 & 32.237 $\pm$ 0.199 \\
    \bottomrule
  \end{tabular}}
\end{table*}

\begin{table}[t]
\centering
\begin{minipage}[t]{0.48\textwidth} 
  \centering
  \caption{Top-1 validation accuracy on ImageNet-1K dataset for ViT-B and ViT-L.}
  \resizebox{0.8\linewidth}{!}{
    \begin{tabular}{lccc}
    \toprule
    Method & Epoch & ViT-B & ViT-L \\
    \midrule
    data2vec 2.0 & 200/150 & 80.5 & 81.8 \\
    data2vec 2.0 + GeSSL & 200/150 & 85.9 & 88.2 \\
    \bottomrule
    \end{tabular}
  }
  \label{tab:q2-1}
\end{minipage}
\hfill 
\begin{minipage}[t]{0.48\textwidth} 
  \centering
  \caption{Downstream classification accuracy of SimCLR-SAS on CIFAR-10.}
  \resizebox{1\linewidth}{!}{
    \begin{tabular}{lcc}
    \toprule
    Method & Subset Size & Top-1 Accuracy (\%) \\
    \midrule
    SimCLR-SAS & 10\% & 79.7 \\
    SimCLR-SAS + GeSSL & 10\% & 84.1 \\
    \bottomrule
    \end{tabular}
  }
  \label{tab:q2-2}
\end{minipage}
\end{table}

\subsection{Few-shot Learning}
\label{sec:app_F.4}

The outstanding performance of GeSSL in the few-shot learning scenario has been confirmed in Section \ref{sec:6.2}, where it can produce good results with limited data. However, the situation becomes complicated in scenarios where data collection is infeasible in real life, such as medical diagnosis and satellite imagery \citep{zheng2015methodologies, tang2012cross}. Therefore, the performance of the model on cross-domain few-shot learning tasks is crucial, as it determines the applicability of the learning model \citep{guo2020broader}. To ensure that GeSSL can achieve robust performance in real-world applications, we further conduct comparative experiments on cross-domain few-shot learning.

\textbf{Experimental setup.} We compare our proposed GeSSL with the few-shot learning baselines as described in Table \ref{tab:few-shot} on cross-domain few-shot learning. The details of the baselines are illustrated in Appendix \ref{sec:app_E}. We adopt six cross-domain few-shot learning benchmark datasets, and divided these datasets into two categories according to their similarity with ImageNet: i) high similarity: CUB \cite{cub}, Cars \cite{cars}, and Places \cite{places}; ii) low similarity: CropDiseases \cite{CropDiseases}, ISIC \cite{isic}, and ChestX \cite{chestx}. The $(N, A)$ in the tables means the $N$-way $A$-shot tasks with $N$ classes and $N \times A$ samples, where each class has $A$ samples augmented from the same image.

\textbf{Results}. Table \ref{tab:cross-domain} presents the performance of the model trained on miniImageNet and transfer to the six cross-domain few-shot learning benchmark datasets mentioned above. By observation, we further validate the performance of our proposed GeSSL: i) Effectiveness: achieves better results than the state-of-the-art baselines on almost all benchmark datasets; ii) Generalization: achieves nearly a 3\% improvement compared to unsupervised few-shot Learning and self-supervised learning on the datasets with significant differences from the training phase; iii) Robustness: achieves better results than the PsCo \cite{jang2023unsupervised} which introduces out-of-distribution samples, even though we do not explicitly consider out-of-distribution samples on datasets with significant differences.

\begin{figure*}[t]
    \centering
    \subfigure[SimCLR]{
        \includegraphics[width=0.15\linewidth]{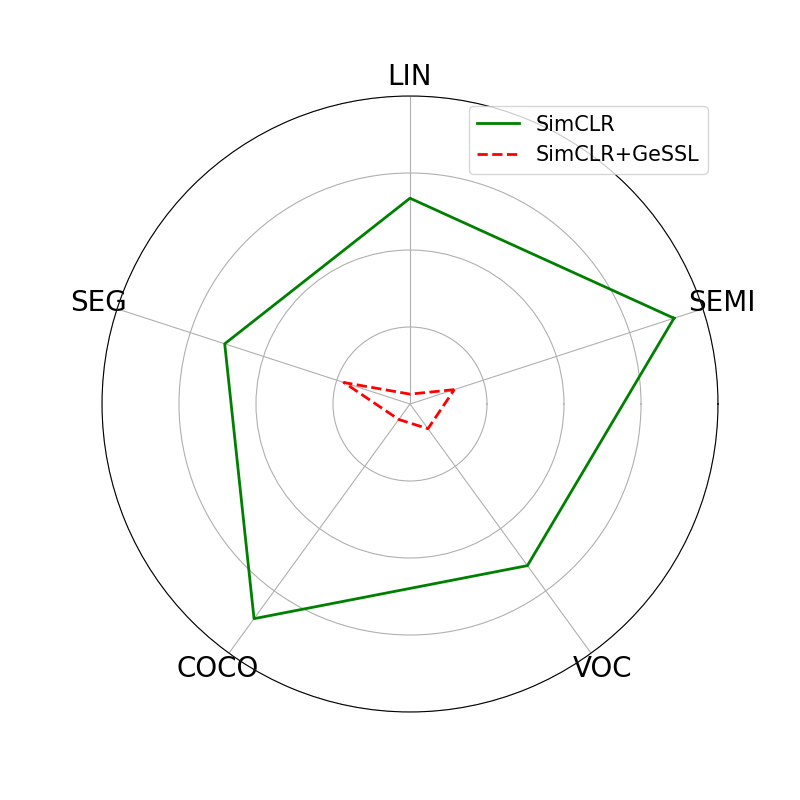}}
        \hfill
    \subfigure[BYOL]{
        \includegraphics[width=0.15\linewidth]{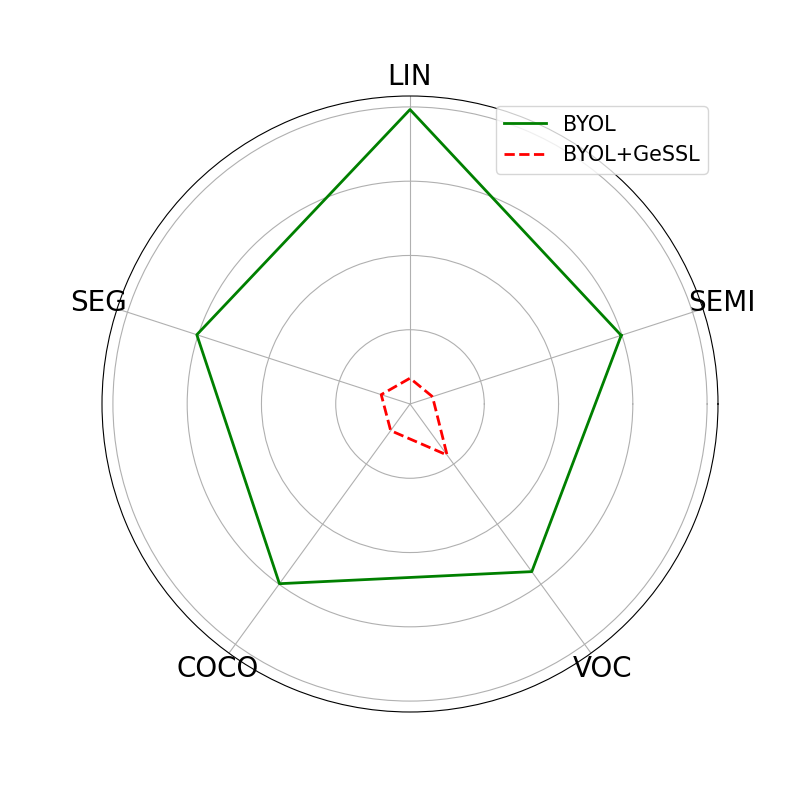}}
        \hfill
    \subfigure[MoCo]{
        \includegraphics[width=0.15\linewidth]{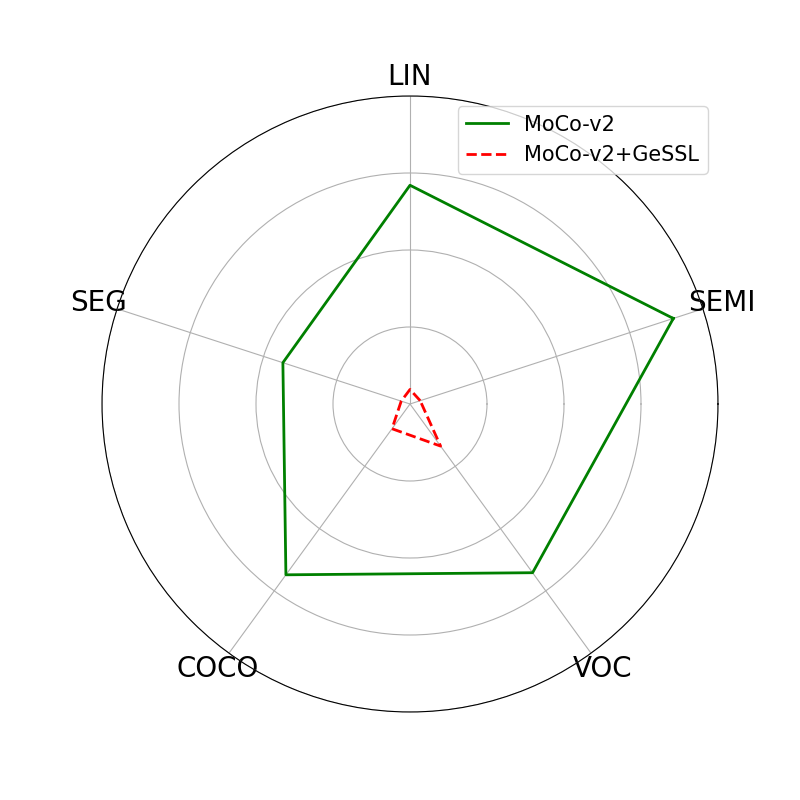}}
        \hfill
    \subfigure[SwAV]{
        \includegraphics[width=0.15\linewidth]{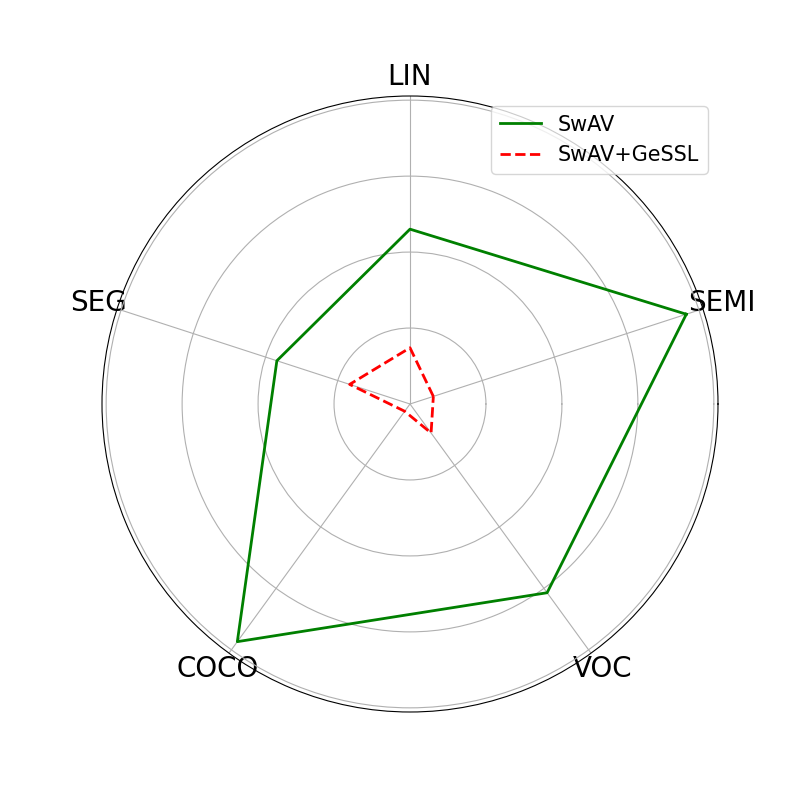}}
        \hfill
    \subfigure[SimSiam]{
        \includegraphics[width=0.15\linewidth]{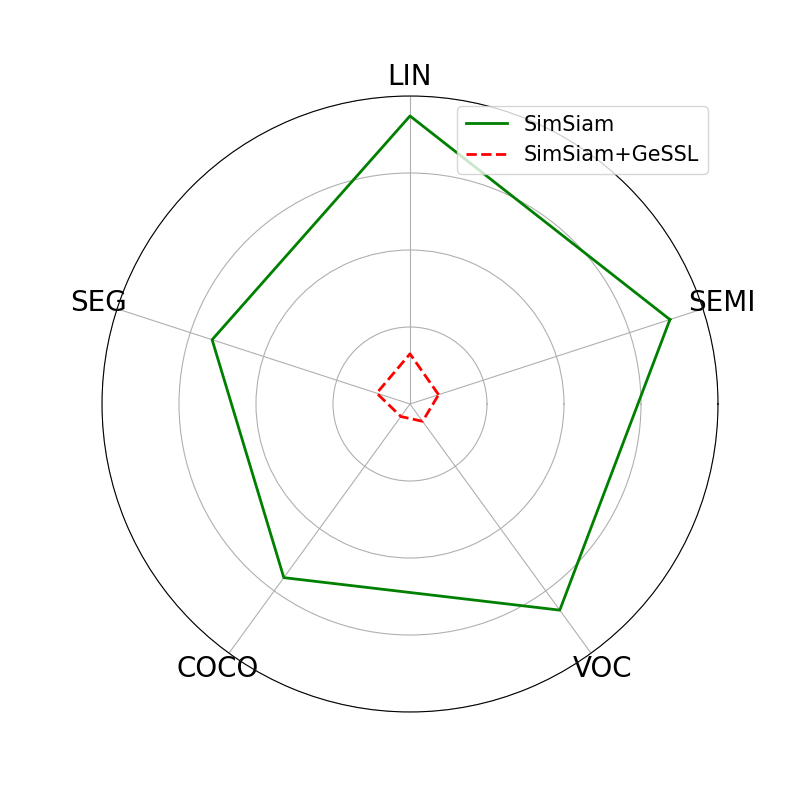}}
        \hfill
    \subfigure[BarlowTwins]{
        \includegraphics[width=0.15\linewidth]{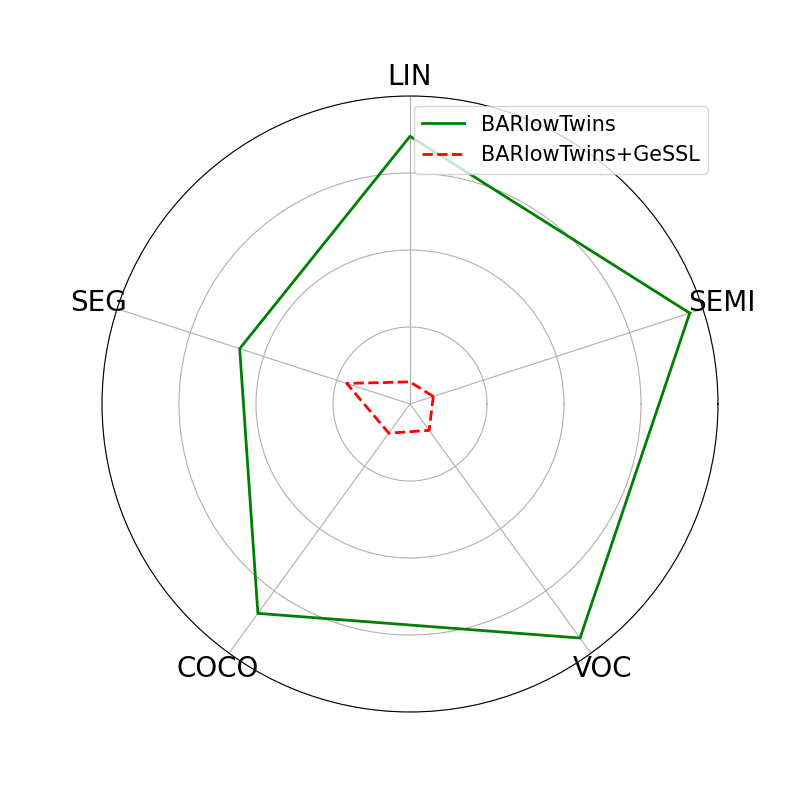}}
        \hfill
        \subfigure[SimCLR]{
        \includegraphics[width=0.15\linewidth]{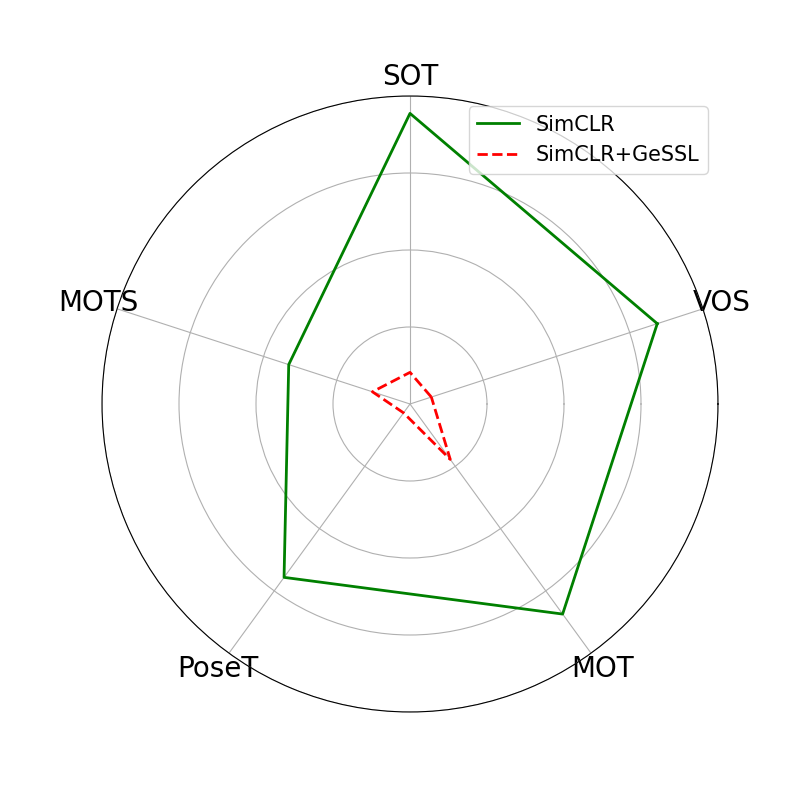}}
        \hfill
    \subfigure[BYOL]{
        \includegraphics[width=0.15\linewidth]{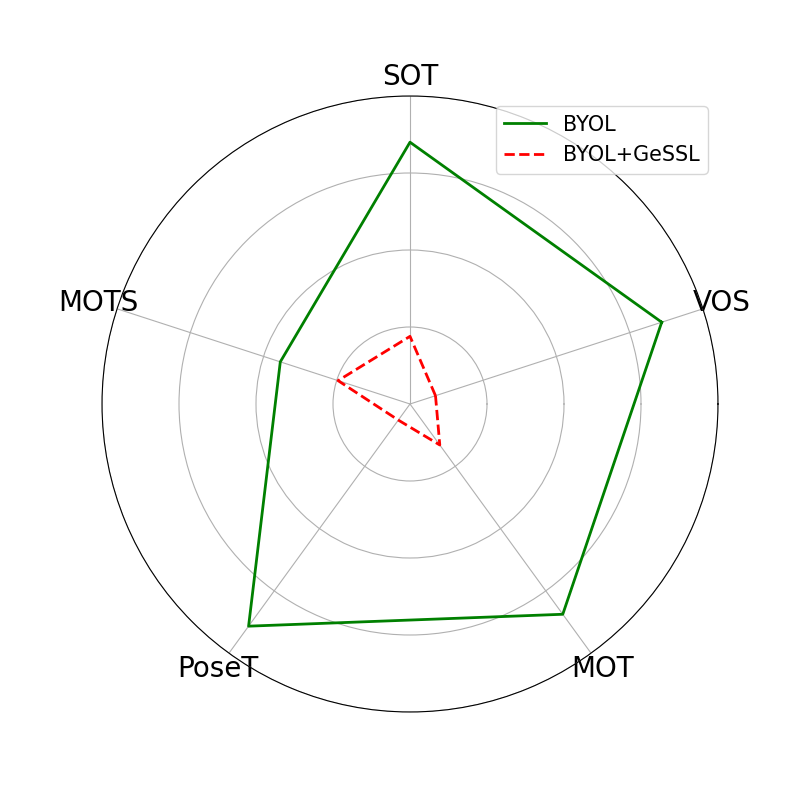}}
        \hfill
    \subfigure[MoCo]{
        \includegraphics[width=0.15\linewidth]{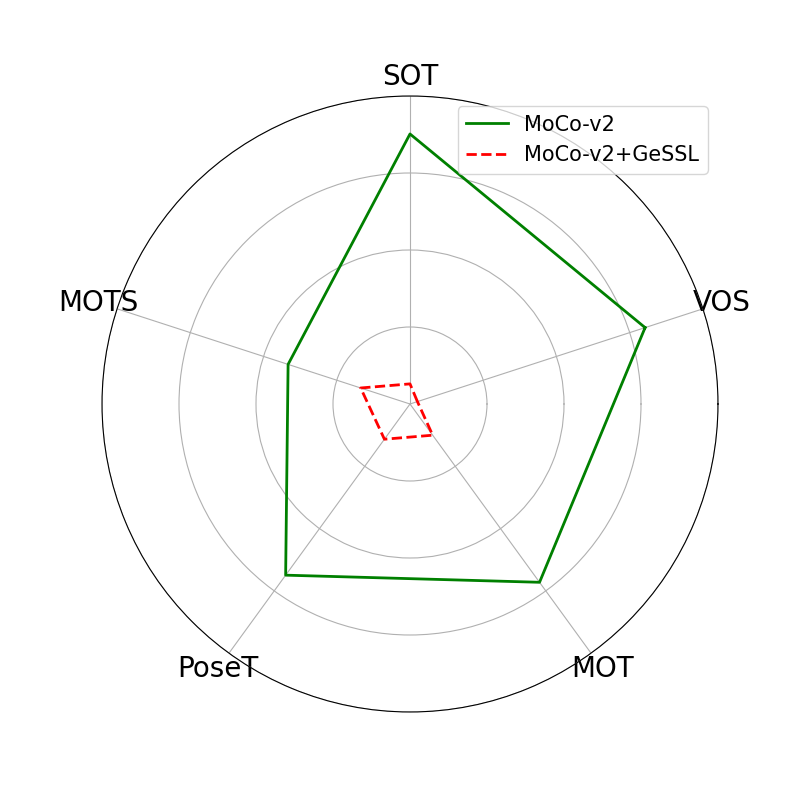}}
        \hfill
    \subfigure[SwAV]{
        \includegraphics[width=0.15\linewidth]{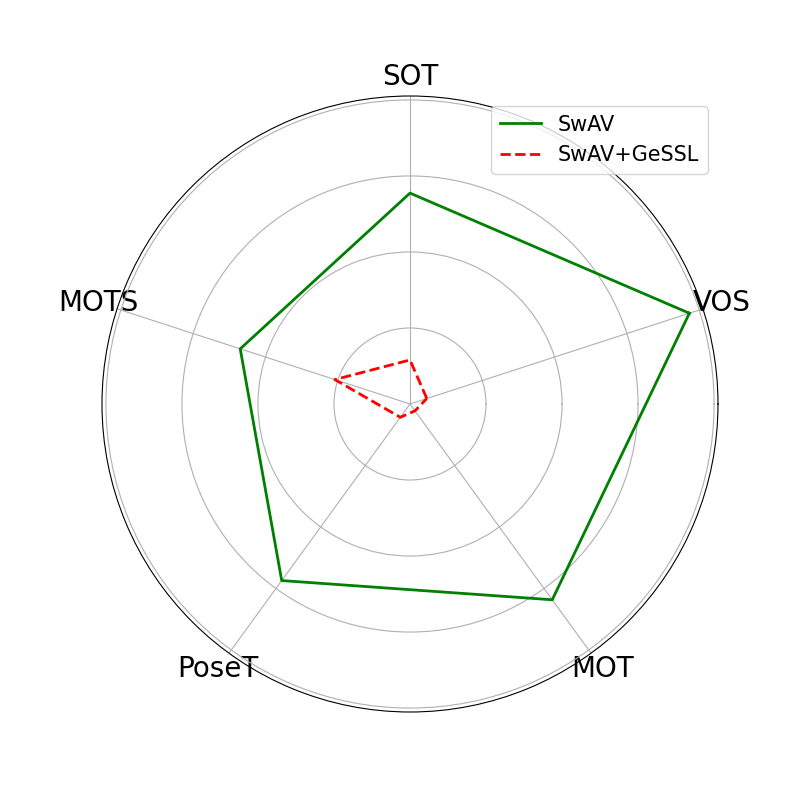}}
        \hfill
    \subfigure[InfoMin]{
        \includegraphics[width=0.15\linewidth]{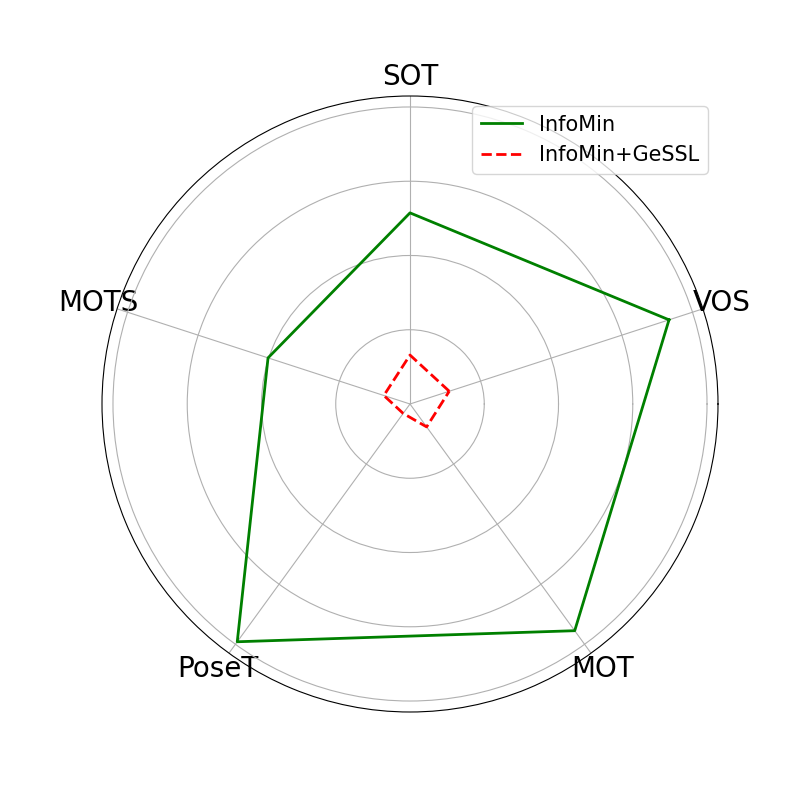}}
        \hfill
    \subfigure[InsDis]{
        \includegraphics[width=0.15\linewidth]{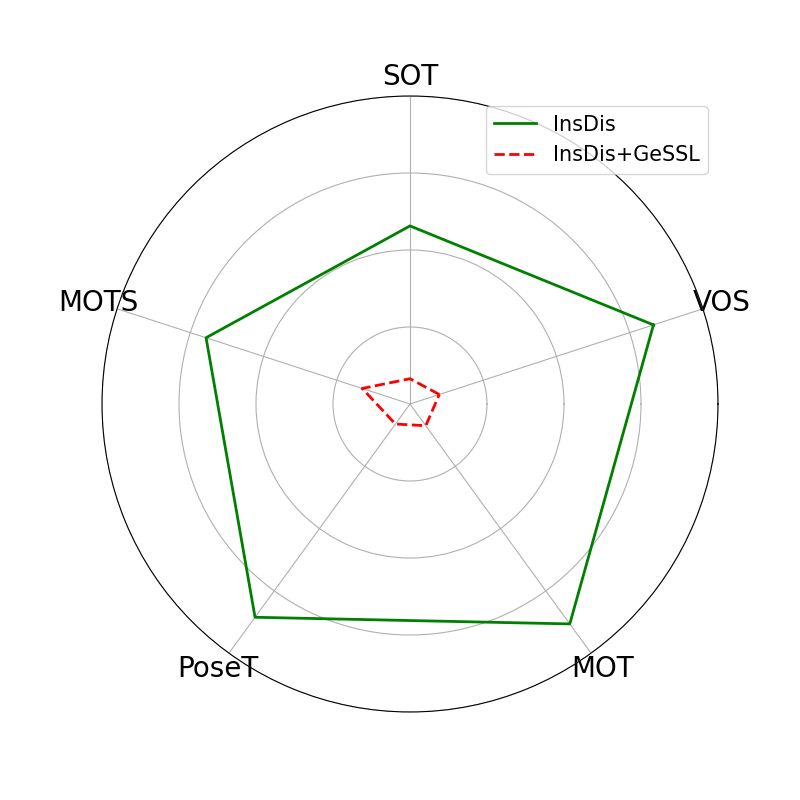}}
    \caption{Universality performance of different models on five image-based tasks (top row) and five video-based tasks (bottom row). We choose $\sigma-$measure as the measurement. It is worth noting that the smaller the $\sigma-$measurefen score, the better the effect. Meanwhile, we normalize the results of $\sigma-$measurefen scores on different datasets and compare the performance between baselines by comparing the corresponding branch of the fan chart.}
    \label{fig:app_sigma}
\end{figure*}

\begin{figure}
    \centering
    \includegraphics[width=0.6\linewidth]{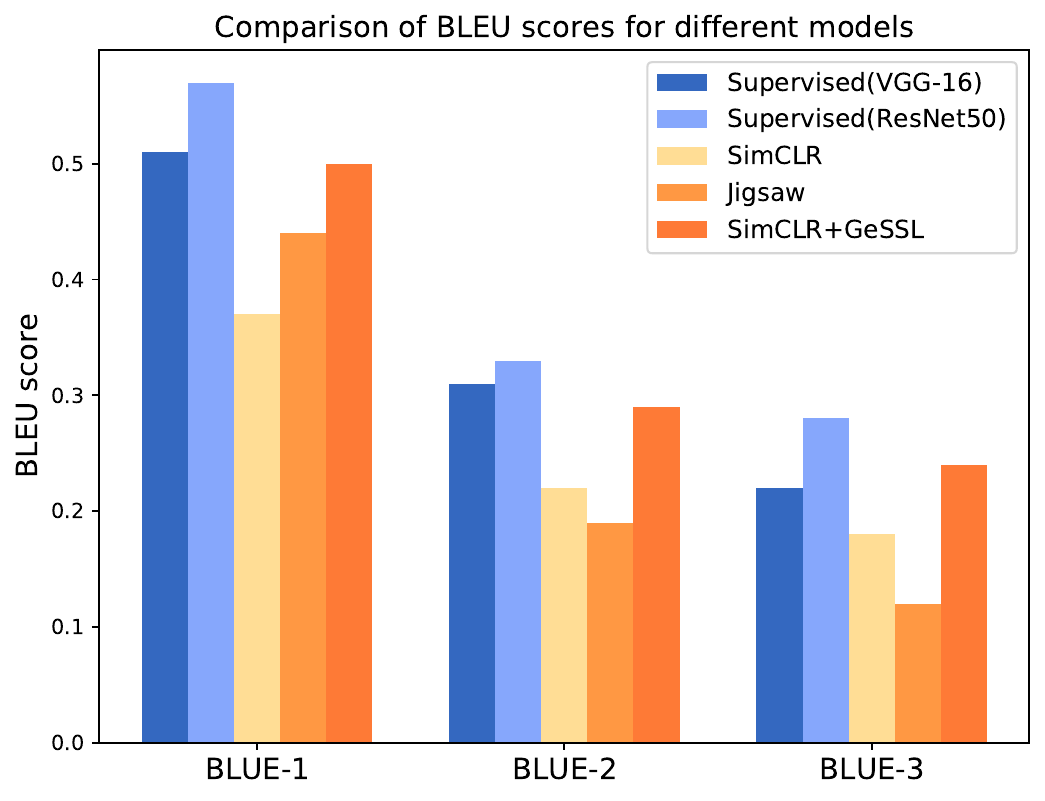}
    \caption{Comparison of BLEU scores for different models, comparing 2 fully supervised and 3 self-supervised pre-text tasks, trained on the Flickr8k.}
    \label{fig:app_gssl}
\end{figure}

\subsection{Universality of Existing SSL Methods}
\label{sec:app_F.5}
Current self-supervised learning (SSL) models overlook the explicit incorporation of universality within their objectives, and the corresponding theoretical comprehension remains inadequate, posing challenges for SSL models to attain universality in practical, real-world applications \cite{huang2021towards, sun2020adashare, ericsson2022self}. 
Therefore, we propose a provable $\sigma-$measure to help evaluate the model universality, and further build GeSSL based on it to explicitly model universality into the SSL's learning objective. In this Section, we specifically quantify the universality scores of existing SSL methods based on $\sigma-$measure, and verify that our proposed GeSSL actually improves the model universality.

Specifically, the $\sigma$-measurement score assesses the difference in performance between the learned model and the optimal model for each task. The optimal model is assumed to output the ground truth, and the performance difference is quantified using the KL divergence between the predicted and true class probability distributions. It compares the predicted class probabilities produced by classifier $\pi$ to the true labels across SSL tasks, such as comparing the predicted values $[0.81, 0.09, 0.03, 0.07]$ to the true labels $[1, 0, 0, 0]$.
Take LIN task with SimCLR as an example, we train SimCLR and SimCLR+GeSSL on the COCO dataset for 200 epochs, then add a MLP after the feature extractor. A new mini-batch is input into both SimCLR and SimCLR+GeSSL to generate class probability distributions for each sample, and the KL divergence between these predicted and true distributions is calculated. After normalization, the scores for the LIN task are obtained, with similar evaluations conducted for other baselines and tasks.

In the experiments, we chose two scenarios based on images and videos to evaluate the model versatility following \cite{liu2022self}. The image-based tasks include linear probing (top-1 accuracy) with 800-epoch pre-trained models (LIN), semi-supervised classification (top-1 accuracy) using 1\% subset of training data (SEMI), object detection (AP) on VOC dataset (VOC) and COCO dataset (COCO), instance segmentation ($\text{AP}^{\text{mask}}$) on COCO dataset (SEG). For video-based tasks, we compute rankings in terms of AUC for SOT, $\mathcal{J}$-mean for VOS, IDF-1 for MOT, IDF-1 for PoseTracking, and IDF-1 for MOTS, respectively. Next, we evaluate the $\sigma$-measurement scores of different baselines before and after the introduction of GeSSL and after training for 200 epochs. Among them, the better model is set to the result of ground truth, and the calculation of $\sigma$-measurement score is performed on a series of randomly sampled tasks.

Figure \ref{fig:app_sigma} shows the comparison results. Note that the lower $\sigma-$measure denotes the better performance. From the results, we can observe that: (i) the $\sigma$-measurement score of the existing SSL model is low and it is difficult to achieve good results in multiple domains and tasks; (ii) after the introduction of GeSSL, the $\sigma$-measurement score of the SSL models are significantly decreased. The results demonstrate that the existing SSL model has limited universality (proves the description in Section \ref{sec:1}), and the performance improvement brought by GeSSL is achieved by improving the universality.

Considering that the above experiments evaluate the evaluation universality of SSL models, here, we construct the following numerical experiments to evaluate learning universality: In the first 20-200 epochs of training (each epoch contains multiple tasks), we evaluate the average performance of multiple $f'$ in each epoch. Each $f_{\theta}^l$ is obtained by updating $f_{\theta}$ on the corresponding support set. We calculate the accuracy of SimCLR before and after the introduction of GeSSL and the ratio $r$ of their effects on the CIFAR-10 data set. If $r<1$, it means that the representation effect learned by the model in each epoch of training is better when introducing GeSSL. The results for every 20 epochs are shown in Table \ref{tab:learning}. The results show that: (i) $r$ is always less than 1, which proves that the representation effect learned after the introduction of GeSSL is significantly improved; (ii) after the introduction of GeSSL, the accuracy of the model is significantly improved, and it becomes stable after 80 epochs, i.e., great results can be achieved for even based on just one iteration and few data. These results show that ``the model $f_{\theta}$ achieves comparable performance on each task quickly with few data during training'' after introducing GeSSL.

\begin{table*}[t]
\centering
\caption{The performance of introducing GeSSL during training. All results are recorded during training using the $\sigma$-measurement.}
\label{tab:learning}
\resizebox{1\linewidth}{!}{
\begin{tabular}{lcccccccccc}
\toprule
\multirow{2}{*}{Metric}& \multicolumn{9}{c}{Training Epochs} \\
& 20 & 40 & 60 & 80 & 100 & 120 & 140 & 160 & 180 & 200 \\
\midrule
Accuracy of SimCLR & 20.1 & 43.6 & 51.2 & 60.2 & 70.3 & 77.2 & 82.3 & 86.1 & 88.7 & 88.6 \\
Accuracy of SimCLR + GeSSL & 42.4 & 67.1 & 83.0 & 92.9 & 93.0 & 94.4 & 94.1 & 93.2 & 94.1 & 94.2 \\
Performance Ratio $r$ & 0.474 & 0.650 & 0.617 & 0.648 & 0.756 & 0.818 & 0.875 & 0.924 & 0.943 & 0.941 \\
\bottomrule
\end{tabular}}
\end{table*}

\subsection{Evaluation on Generative Self-supervised Learning}
\label{sec:app_F.6}

In this Section, we evaluate the effectiveness of the proposed GeSSL on the generative self-supervised learning paradigm. We conduct experiments on three scenarios, including image generation, image captioning, and object detection and segmentation.
\paragraph{Evaluation on Image Generation} 
To explore the effect of GeSSL on generative SSL, we conduct a set of experiments on ImageNet-1K dataset~\cite{deng2009imagenet}.
Specifically, we begin by conducting self-supervised pre-training on the ImageNet-1K (IN1K) training set. Following this, we carry out supervised training to assess the representations using either (i) end-to-end fine-tuning or (ii) linear probing. The results are reported as the top-1 validation accuracy for a single 224×224 crop. For this process, we utilize ViT-Large (ViT-L/16)~\cite{dosovitskiy2020image} as the backbone. Note that ViT-L is very big (an order of magnitude bigger than ResNet-50~\cite{he2016deep}) and tends to overfit, as shown in Table \ref{tab:ViT-Large}. The comparison results are shown in Table \ref{tab:gssl_1}. We can observe that GeSSL achieves stable performance improvements

\begin{table}[t]
    \centering
    \caption{Comparison between models.}
    \begin{tabular}{c|ccccc}
    \toprule
    Method  & scratch, original & scratch, our impl. & baseline MAE & MAE + Our\\
    \midrule
    Top 1 & 76.5 & 82.7 & 85.3 & 88.1\\
    \bottomrule
    \end{tabular}
\label{tab:ViT-Large}
\end{table}

\begin{table}[t]
    \centering
    \caption{Comparisons with previous results on ImageNet-1K. The ViT models are B/16, L/16, H/14~\cite{dosovitskiy2020image}. The pre-training data is the ImageNet-1K training set (except the tokenizer in BEiT was pre-trained on 250M DALLE data~\cite{ramesh2021zero}). All results are on an image size of 224, except for ViT-H with an extra result of 448.}
    \begin{tabular}{c|cccccc}
    \toprule
    Method  & pre-train data & ViT-B & ViT-L & ViT-H & ViT-H$_{448}$\\
    \midrule
    DINO & IN1K & 82.8 & - & - & -\\
    MoCo & IN1K & 83.2 & 84.1 & - & - \\
    BEiT & IN1K+DALLE & 83.2 & 85.2 & - & - \\
    MAE & IN1K & 83.6 & 85.9 & 86.9 & 87.8\\
    \midrule
    MAE+Ours & IN1K & 87.6 & 88.5 & 89.2 & 89.7\\
    \bottomrule
    \end{tabular}
\label{tab:gssl_1}
\end{table}

\paragraph{Evaluation on Image Captioning} We use the commonly used protocol following \cite{mohamed2022image}. The dataset we use to train the pretext task is the unlabeled part of MSCOCO dataset \cite{vinyals2016show}, which contains 123K images with an average resolution of $640 \times 480$ pixels. This dataset contains color and grayscale images. For downstream tasks, we use the Flicker8K dataset \cite{hodosh2013framing}. Next, we train it using pre-trained pre-text tasks supervised by VGG-16 and ResNet-50, as well as self-supervised pre-text tasks from SimCLR and Jigsaw Puzzle solutions. In the next step, to evaluate the results, we use the BLEU (Bilingual Evaluation Research) score as the evaluation metric, which evaluates the generated sentences against the reference sentences, where a perfect match is 1 and a perfect mismatch is 0, calculating scores for 1, 2, 3 and 4 cumulative n-grams. The results are shown in Figure \ref{fig:app_gssl}. From the results, we can observe that after introducing the GeSSL framework we proposed, the model effect has been further improved, stably exceeding the SOTA of the SSL method, and even approaching the supervised learning results. The results show that our proposed GeSSL can still achieve good results in generative self-supervised learning.

\paragraph{Evaluation on Object Detection and Segmentation}
For object detection and segmentation, we fine-tune Mask R-CNN ~\cite{he2017mask} end-to-end on COCO~\cite{lin2014microsoft}. The ViT backbone is adapted for use with FPN~\cite{lin2017feature}. We report box AP for object detection and mask AP for instance segmentation. The results are shown in Table \ref{tab:gssl_3}. Compared to supervised pre-training, our MAE performs better under all configurations. Our method still achieves optimal results, demonstrating its effectiveness.

\begin{table}
    \centering
    \caption{COCO object detection and segmentation using a ViT Mask R-CNN baseline. All self-supervised entries use IN1K data without labels, and Mask AP follows a similar trend as box AP.}
    \label{tab:gssl_3}
    \resizebox{0.6\linewidth}{!}{
	\begin{tabular}{lccccc}
		\toprule
            \multirow{2.5}{*}{Method} &\multirow{2.5}{*} {pre-train data} &\multicolumn{2}{c}{AP$^{\text{box}}$} & \multicolumn{2}{c}{AP$^\text{mask}$} \\
	    \cmidrule(lr){3-4} \cmidrule(lr){5-6}
	    & & ViT-B & ViT-L & ViT-B & ViT-L \\
	    \midrule
	    supervised & IN1K w/ labels & 47.9 & 49.3 & 42.9 & 43.9 \\
		MoCo v3 & IN1K & 47.9 & 49.3 & 42.7 & 44.0 \\ 
            BEiT & IN1K+DALLE & 49.8 & 53.3 & 44.4 & 47.1\\ 
            MAE & IN1K & 50.3 & 53.3 & 44.9 & 47.2\\
            \midrule
            MAE + Our & IN1K & 54.9 & 57.3 & 47.9 & 53.0\\ 
	\bottomrule
	\end{tabular}
 }
\end{table}

\begin{table}
\caption{
Performance on for text recognition.}
\centering
\resizebox{0.4\linewidth}{!}{
\begin{tabular}{lcc}
\toprule
\textbf{Methods} & \textbf{IIIT5K} & \textbf{IC03}\\
\toprule
SimCLR \cite{simclr} & 1.7 & 3.8\\
SeqCLR \cite{aberdam2021sequence} & 35.7 & 43.6\\
\midrule
\rowcolor{blue!10}SimCLR + GeSSL & 21.4 & 20.8\\
\rowcolor{blue!10}SeqCLR + GeSSL & 41.3 & 50.6\\
\bottomrule
\end{tabular}}
\label{tab:app_modal}
\end{table}

\subsection{Evaluation on More Modalities}
\label{sec:app_F.7}

GeSSL proposed in this work can be applied in various fields and domains, e.g., instance segmentation, video tracking, sample generation, etc., as mentioned before. Here, we provide the experiments of GeSSL on text modality-based datasets, i.e., IC03 and IIIT5K \cite{yasmeen2020text}, which we have conducted before. We follow the same experimental settings as mentioned in \cite{aberdam2021sequence}. The results shown in Table \ref{tab:app_modal} demonstrate that GeSSL achieves stable effectiveness and robustness in various modalities combined with the above experiments.

\section{Details of Ablation Study}
\label{sec:app_G}
In this section, we introduce the experimental details and more comprehensive analysis of the ablation studies (Subsection \ref{sec:6.6}).

\subsection{Model Efficiency}
\label{sec:app_G.2}

This ablation study explores the efficiency of self-supervised models before and after applying GeSSL. Specifically, we choose five baselines, including SimCLR \cite{simclr}, MOCO \cite{moco}, BYOL \cite{byol}, Barlow Twins \cite{barlowtwins}, and SwAV \cite{swav}. Then, we evaluate the accuracy, training hours, and parameter size of these models on STL-10 before and after applying our proposed GeSSL. We use the same linear evaluation setting as in Section \ref{sec:6.2} of the main text. Finally, we plot the trade-off scatter plot by recording the average values of five runs. The results are shown in Figure 2 of the main text, where the horizontal axis represents the training hours and the vertical axis represents the accuracy. The center of each circle represents the result of the training time and accuracy of each model, and the area of the circle represents the parameter size. The numerical results of this experiment are shown in Table \ref{tab:app_model}. From the results, we can see that: (i) GeSSL can significantly improve the performance and computational efficiency of self-supervised learning models; (ii) our designed self-motivated target achieves the goal of guiding the model update toward universality with few samples and fast adaptation; (iii) although GeSSL optimizes based on bi-level optimization, the impact of the increased parameter size of GeSSL is negligible.

\begin{table}
\caption{
Training cost per epoch of SSL models.}
\centering
\resizebox{0.45\linewidth}{!}{
\begin{tabular}{lc}
\toprule
\textbf{Methods} & \textbf{Training Cost per Epoch (s)}\\
\toprule
SimCLR \cite{simclr} & 12.8\\
MOCO \cite{moco} & 16.9 \\ 
\rowcolor{blue!10}SimCLR + GeSSL & 9.6\\
\rowcolor{blue!10}MOCO + GeSSL & 12.0\\
\bottomrule
\end{tabular}}
\label{tab:epoch_time}
\end{table}

Note that although the optimization method used by GeSSL is more complex, one of its core goals is to accelerate model convergence, i.e., achieve greater performance improvement per unit of time. This does not imply that GeSSL always requires fewer epochs to reach the optimal result. In fact, GeSSL uses approximate implicit differentiation with finite difference (AID-FD) for updates instead of conventional explicit second-order differentiation (as mentioned in Appendix \ref{sec:app_G.4}). Moreover, GeSSL constructs a self-motivated target that guides the model to optimize more effectively in a specific task. Therefore, the efficiency improvement is reflected in the computational efficiency and effectiveness of updates per epoch, rather than simply reducing the total number of epochs.
Furthermore, to verify whether the efficiency improvement is attributable to a single epoch, we separately measured the computational overhead of SSL baseline algorithms after integrating GeSSL for a single epoch. The results, presented in Table \ref{tab:epoch_time}, demonstrate that with a consistent batch size, GeSSL enhances the computational efficiency and the effectiveness of updates per epoch for the SSL baseline algorithms.

\begin{table*}[t]
\centering
\caption{Model analysis including parameter size, training time, and performance.}
 \vspace{0.1in}
\label{tab:app_model}
\resizebox{\linewidth}{!}{
\begin{tabular}{lcccc}
\toprule
\textbf{Methods} & \textbf{Memory Footprint (MiB)} & \textbf{Parameter Size (M)}&\textbf{Training Time (h)} & \textbf{Accuracy (\%)} \\
\midrule
SimCLR & 2415 & 23.15&4.15 & 90.5 \\
MOCO & 2519 & 24.01&4.96 & 90.9 \\
BYOL & 2691 & 25.84&6.98 & 91.9 \\
BarlowTwins & 2477 & 23.15&5.88 & 90.3 \\
SwAV & 2309 & 22.07&4.45 & 90.7 \\
\rowcolor{blue!10}SimCLR+GeSSL & 2784 & 26.21 & 3.36 & 93.4 \\
\rowcolor{blue!10}MOCO+GeSSL & 2912 & 27.20 & 4.23 & 94.6 \\
\rowcolor{blue!10}BYOL+GeSSL & 2875 & 28.01 & 5.70 & 94.8 \\
\rowcolor{blue!10}BarlowTwins+GeSSL & 2856 & 27.11 &5.39 & 94.2 \\
\rowcolor{blue!10}SwAV+GeSSL & 3012 & 28.61 & 3.96 & 93.2 \\
\bottomrule
\end{tabular}}
\end{table*}

\begin{figure}
    \centering
    \includegraphics[width=\linewidth]{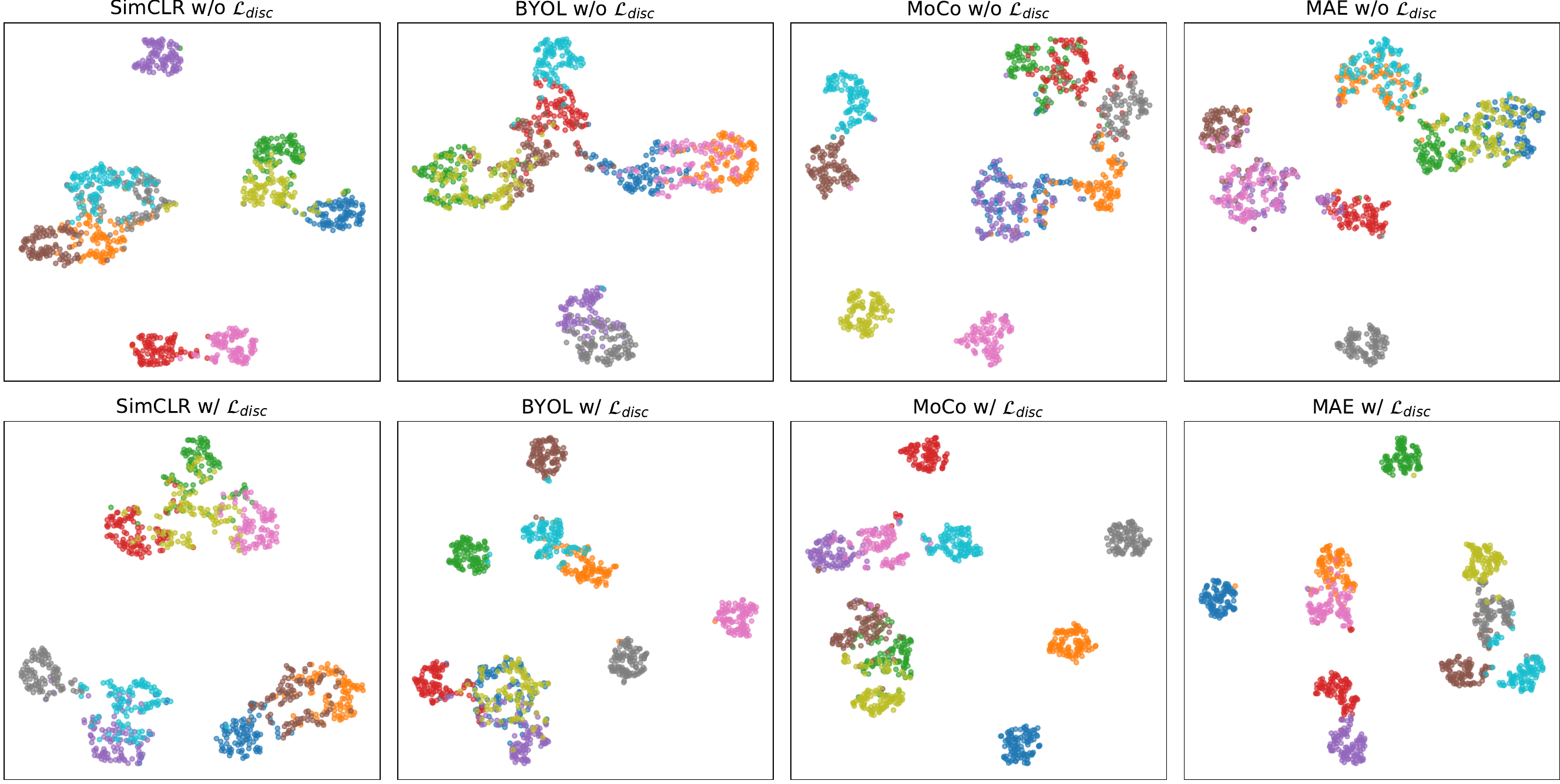}
    \caption{Ablation study of $\mathcal{L}_{disc}$. We perform t-SNE visualization to evaluate the classification performance of the SSL model before and after introducing $\mathcal{L}_{disc}$.}
    \label{fig:app_G_ablation_disc}
\end{figure}

\begin{figure*}[t]
  \begin{minipage}{0.48\textwidth}
    \centering
    \includegraphics[width=\linewidth]{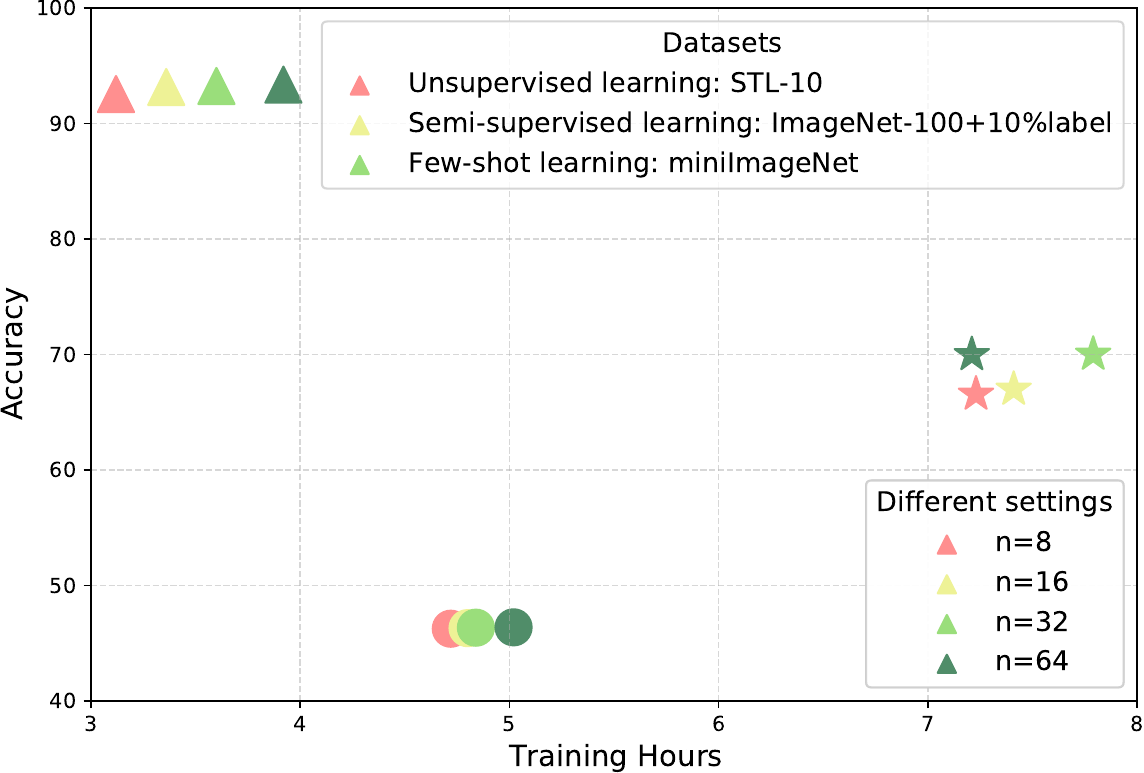}
    \caption{Ablation study of the number of pairs.}
    \label{fig:app_batchsize}
  \end{minipage}
  \hfill
  \begin{minipage}{0.48\textwidth}
    \centering
    \includegraphics[width=\linewidth]{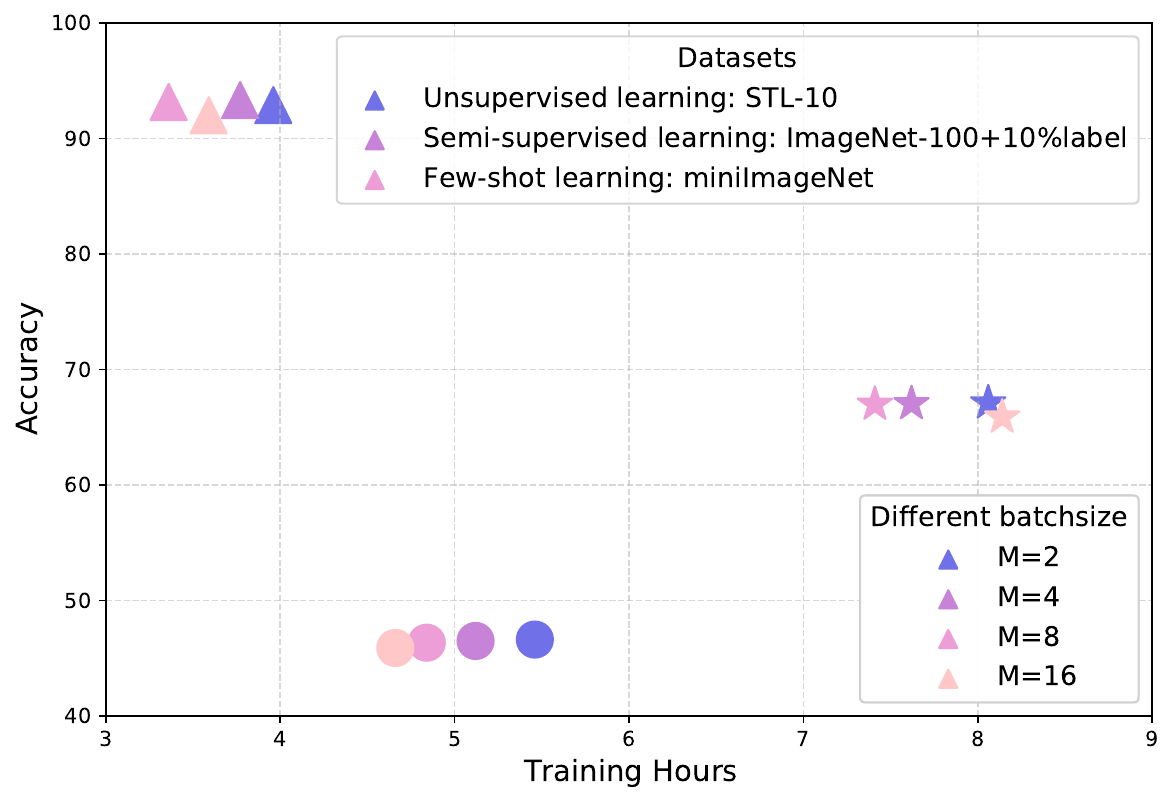}
    \caption{Ablation study of the batchsize.}
    \label{fig:app_n}
  \end{minipage}
\end{figure*}

\begin{figure*}[t]
  \begin{minipage}{0.48\textwidth}
    \centering
    \includegraphics[width=\linewidth]{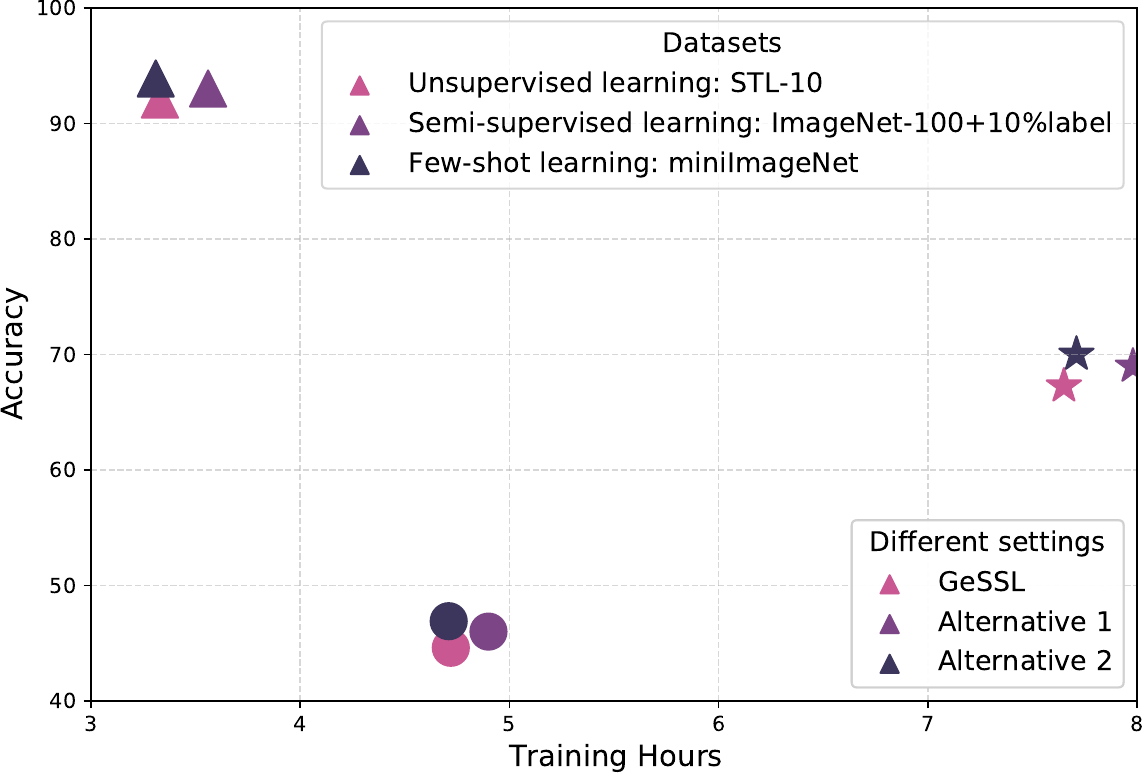}
    \caption{Evaluation of bi-level optimization.}
    \label{fig:abla_bi-level}
  \end{minipage}
  \hfill
  \begin{minipage}{0.48\textwidth}
    \centering
    \includegraphics[width=\linewidth]{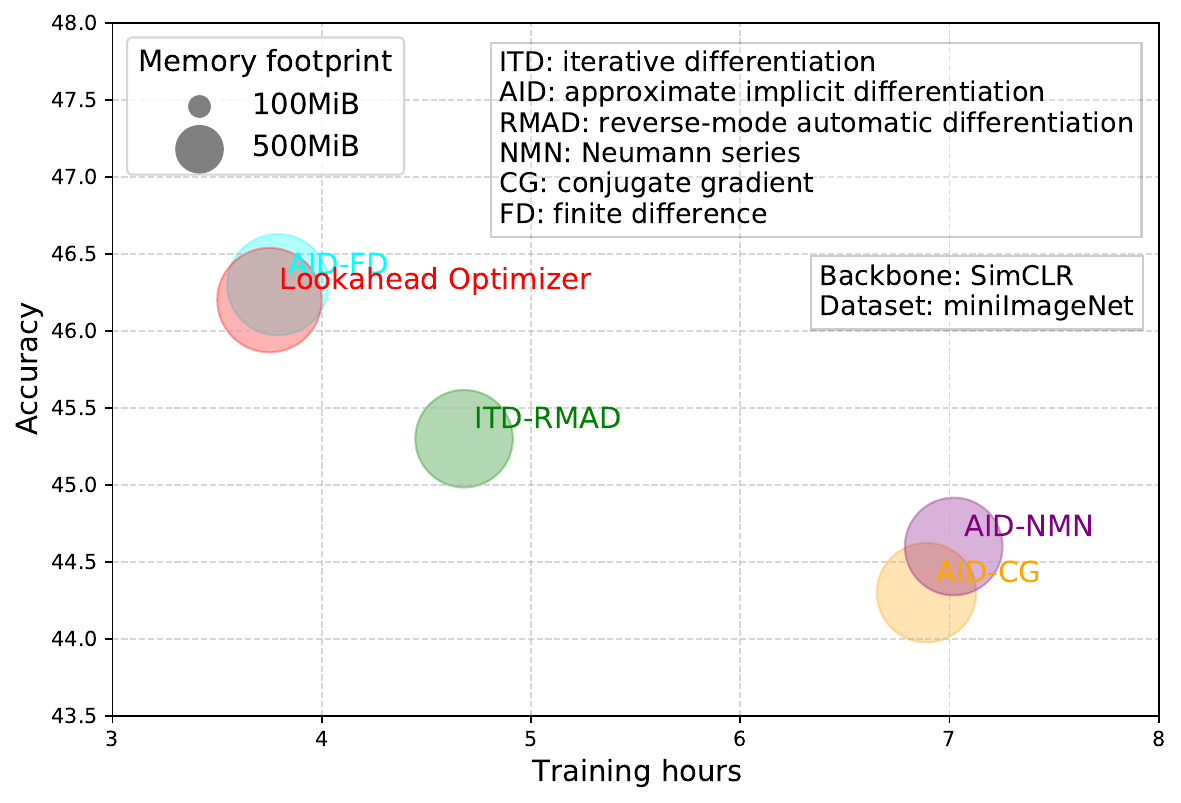}
    \caption{Implementation of optimization.}
    \label{fig:ab_4}
  \end{minipage}
\end{figure*}

\subsection{Ablation Study of $\mathcal{L}_{disc}$}
\label{sec:app_G_ablation_disc}

To evaluate the impact of $\mathcal{L}_{disc}$, we design a series of experiments. $\mathcal{L}_{disc}$ is intended to enhance discriminative power by enforcing constraints that sharpen the SSL model’s decision boundaries.  We therefore visualize classification performance before and after adding $\mathcal{L}_{disc}$.  Using SimCLR, BYOL, MoCo, and MAE as baselines, we randomly select 10 classes from ImageNet-100 (100 samples per class) and compare each model with and without $\mathcal{L}_{disc}$.  As shown in Figure \ref{fig:app_G_ablation_disc}, introducing $\mathcal{L}_{disc}$ produces noticeably sharper class boundaries, demonstrating its effectiveness in improving model discriminability.

\subsection{More Experiments of Parameter Sensitivity}
\label{sec:app_G.5}

Considering that our framework updates the self-supervised model $f_{\theta}$ in GeSSL based on $M$ tasks simultaneously, the number of sampled samples per batch of self-supervised learning directly determines the class diversity of the data in the task. In this section, we further conduct ablation experiments on the number of pairs within each batch and the batch size (the number of tasks) that are learned simultaneously.

Specifically, we choose the commonly used STL-10 for unsupervised learning, ImageNet with 10\% label for semi-supervised learning, and miniImageNet for few-shot learning, and evaluate the performance of SimCLR + GeSSL under different batch sizes and different $n$ values. Figure \ref{fig:app_batchsize} shows the impact of different number of pairs for SSL. The results show that SimCLR + GeSSL always outperforms SimCLR under any batch size. A larger batch size leads to a slightly larger performance improvement for SimCLR + GeSSL, but also increases the computational resource consumption. Therefore, in this study, we build tasks based on images with a batch size of $n=16$ or $n=32$. Figure \ref{fig:app_n} shows the impact of the batchsize for the outer-loop optimization. The results indicate that $m=8$ is a better trade-off between model accuracy and time consumption. In the setting of our GeSSL, we also choose $m=8$ as the hyperparameter setting.

\begin{table}
\caption{Performance on for a large batchsize.}
\centering
\resizebox{0.5\linewidth}{!}{
\begin{tabular}{lcc}
\toprule
\textbf{Methods} & \textbf{Accuracy} & \textbf{Training Cost}\\
\toprule
SimCLR \cite{simclr} & 90.8 & 5.2\\
\rowcolor{blue!10}SimCLR + GeSSL & 93.8 & 3.9\\
\bottomrule
\end{tabular}}
\label{tab:nbatchsize}
\end{table}

In addition, considering that GeSSL updates every $m$ mini-batches, we evaluate the baseline performance under $m \times$ the original batch size. Specifically, we adopt the same experimental setup as in Figure \ref{fig:ab_2}, with the only difference being that we increase the batch size of the SimCLR baseline by a factor of $m$ and record the results. The results are shown in Table \ref{tab:nbatchsize}, which indicates that the performance of SimCLR, after converging with the larger training data, remains largely unchanged and still inferior to GeSSL.

\subsection{Evaluation and Implementation of the Bi-level Optimization}
\label{sec:app_G.4}

As mentioned in Subsection \ref{sec:6.2}, to assess the advantages of our bi-level optimization, we compare its performance against two alternatives: (i) jointly optimizing the inner and outer objectives in a single stage; and (ii) training a distinct $f'$ for each mini‐batch. The results in Figure \ref{fig:abla_bi-level} demonstrate that our bi‐level optimization (Subsection \ref{sec:4.3}) achieves state-of-the-art performance.

The model of GeSSL is updated based on bi-level optimization, and the model gradients for each level are obtained by combining the optimal response Jacobian matrices through the chain rule. In practical applications, multi-level gradient computation requires a lot of memory and computation \cite{choe2022betty}, so we hope to introduce a more concise gradient backpropagation and update method to reduce the computational complexity. Specifically, we consider two types of gradient update methods, including iterative differentiation (ITD) \cite{finn2017model} and approximate implicit differentiation (AID) \cite{grazzi2020iteration}. We provide implementations of four popular ITD/AID algorithms, including ITD with reverse-mode automatic differentiation (ITD-RMAD) \cite{finn2017model}, AID with Neumann series (AID-NMN) \cite{lorraine2020optimizing}, AID with conjugate gradient (AID-CG) \cite{rajeswaran2019meta}, and AID with finite difference (AID-FD) \cite{liu2018darts}. We also choose the recently proposed optimizer, i.e., Lookahead \cite{zhang2019lookahead} for comparison. We denote the the upper-level parameters and the lower-level parameters as $\theta$ and $\phi$, respectively. All the way of gradient update of the bi-level optimization are as follows:

\noindent\textbf{ITD-RMAD} \cite{finn2017model}, ITD with reverse-mode automatic differentiation applies the implicit function theorem to the lower-level optimization problem and computes the gradients of the upper-level objective with respect to the upper-level parameters using reverse-mode automatic differentiation. The update process is as follows:

\begin{itemize}
    \item Solve the lower-level optimization problem $\phi^* = \arg\min_\phi L(\phi, \theta)$ using gradient descent.
    \item Compute the gradient of the upper-level objective $g(\theta) = F(\phi^*, \theta)$ with respect to $\theta$ using reverse-mode automatic differentiation: 
\end{itemize}
    \begin{equation}
    \begin{array}{l}
         \nabla_\theta g(\theta) = \nabla_\theta F(\phi^*, \theta) 
         -\nabla_\phi F(\phi^*, \theta)^T (\nabla_\phi L(\phi^*, \theta))^{-1} \nabla_\theta L(\phi^*, \theta)
    \end{array}
    \end{equation}
\begin{itemize}
    \item Update the upper-level parameters using gradient descent or other methods: $\theta \leftarrow \theta - \alpha \nabla_\theta g(\theta)$.
\end{itemize}

\noindent\textbf{AID-NMN} \cite{lorraine2020optimizing}, AID with Neumann series, approximates the inverse of the Hessian matrix of the lower-level objective using a truncated Neumann series expansion and computes the gradients of the upper-level objective with respect to the upper-level parameters using forward-mode automatic differentiation. The update process is as follows:

\begin{itemize}
    \item Solve the lower-level optimization problem $\phi^* = \arg\min_\phi L(\phi, \theta)$ using gradient descent.
    \item Compute the gradient of the upper-level objective $g(\theta) = F(\phi^*, \theta)$ with respect to $\theta$ using forward-mode automatic differentiation: 
\end{itemize}
    \begin{equation}
    \begin{array}{l}
        \nabla_\theta g(\theta) = \nabla_\theta F(\phi^*, \theta) 
        - \nabla_\phi F(\phi^*, \theta)^T (\nabla_\phi L(\phi^*, \theta))^{-1} \nabla_\theta L(\phi^*, \theta) \\[8pt]
        \approx \nabla_\theta F(\phi^*, \theta)
        - \nabla_\phi F(\phi^*, \theta)^T \sum_{k=0}^K (-1)^k (\nabla_\phi^2 L(\phi^*, \theta))^k \nabla_\theta L(\phi^*, \theta)
    \end{array}
    \end{equation}
    where $K$ is the truncation order of the Neumann series.
\begin{itemize}
    \item Update the upper-level parameters using gradient descent or other methods: $\theta \leftarrow \theta - \alpha \nabla_\theta g(\theta)$.
\end{itemize}

\noindent\noindent\textbf{AID-CG} \cite{rajeswaran2019meta}, AID with conjugate gradient, solves a linear system involving the Hessian matrix of the lower-level objective using the conjugate gradient algorithm and computes the gradients of the upper-level objective with respect to the upper-level parameters using forward-mode automatic differentiation. The update process is as follows:

\begin{itemize}
    \item Solve the lower-level optimization problem $\phi^* = \arg\min_\phi L(\phi, \theta)$ using gradient descent or other methods.
    \item Compute the gradient of the upper-level objective $g(\theta) = F(\phi^*, \theta)$ with respect to $\theta$ using forward-mode automatic differentiation: 
\end{itemize}
    \begin{equation}
    \begin{array}{l}    
        \nabla_\theta g(\theta) = \nabla_\theta F(\phi^*, \theta) \\[8pt]- \nabla_\phi F(\phi^*, \theta)^T (\nabla_\phi L(\phi^*, \theta))^{-1} \nabla_\theta L(\phi^*, \theta) \approx \nabla_\theta F(\phi^*, \theta) \\[8pt]- \nabla_\phi F(\phi^*, \theta)^T v
    \end{array}
    \end{equation}
    where $v$ is the solution of the linear system $(\nabla_\phi^2 L(\phi^*, \theta)) v = \nabla_\theta L(\phi^*, \theta)$ obtained by the conjugate gradient algorithm.
\begin{itemize}
    \item Update the upper-level parameters using gradient descent or other methods: $\theta \leftarrow \theta - \alpha \nabla_\theta g(\theta)$.
\end{itemize}

\noindent\textbf{AID-FD} \cite{liu2018darts}, AID with finite difference, approximates the inverse of the Hessian matrix of the lower-level objective using a finite difference approximation and computes the gradients of the upper-level objective with respect to the upper-level parameters using forward-mode automatic differentiation. The update process is as follows:

\begin{itemize}
    \item Solve the lower-level optimization problem $\phi^* = \arg\min_\phi L(\phi, \theta)$ using gradient descent or other methods.
    \item Compute the gradient of the upper-level objective $g(\theta) = F(\phi^*, \theta)$ with respect to $\theta$ using forward-mode automatic differentiation: 
\end{itemize}
    \begin{equation}
    \begin{array}{l}      
        \nabla_\theta g(\theta) = \nabla_\theta F(\phi^*, \theta) \\[8pt]- \nabla_\phi F(\phi^*, \theta)^T (\nabla_\phi L(\phi^*, \theta))^{-1} \nabla_\theta L(\phi^*, \theta) \\[8pt]\approx \nabla_\theta F(\phi^*, \theta) \\[8pt]- \nabla_\phi F(\phi^*, \theta)^T \frac{\nabla_\theta L(\phi^* + \epsilon \nabla_\theta L(\phi^*, \theta), \theta) - \nabla_\theta L(\phi^*, \theta)}{\epsilon}
    \end{array}
    \end{equation}
    where $\epsilon$ is a small positive constant for the finite difference approximation.
\begin{itemize}    
    \item Update the upper-level parameters using gradient descent or other methods: $\theta \leftarrow \theta - \alpha \nabla_\theta g(\theta)$.
\end{itemize}

\noindent\textbf{Lookahead} \cite{zhang2019lookahead} introduces a novel approach to optimization by maintaining two sets of weights: the fast and the slow weights. The fast weights, $\theta_{\text{fast}}$, are updated frequently through standard optimization techniques, while the slow weights, $\theta_{\text{slow}}$, are updated at a lesser frequency. The key formula that updates the slow weights is given by:
\begin{equation}
\theta_{\text{slow}} \leftarrow \theta_{\text{slow}} + \alpha (\theta_{\text{fast}} - \theta_{\text{slow}})
\end{equation}
where $\alpha$ is a hyperparameter controlling the step size. This method aims to stabilize training and ensure consistent convergence.

The results shown in Figure 4 of the main text demonstrate that approximate implicit differentiation with finite difference also achieves optimal results on the SSL model. Our optimization process is also based on this setting.

\section{More Discussion}
\label{sec:app_H}

\subsection{Differences between GeSSL and Meta-Learning}
In the main text, we have illustrated the differences between GeSSL and meta-learning and the advantages of GeSSL. In this section, we further elaborate on this and list different meta-learning methods for comparison.

Meta-learning \cite{maml,wang2024towards,protonet}, often referred to as "learning to learn", has emerged as a prominent approach to improve the efficiency and adaptability of machine learning models, especially in scenarios with limited data. The fundamental idea behind meta-learning is to train models that can rapidly adapt to new tasks with minimal data by leveraging prior experiences gained from a range of related tasks.

Few-shot Learning \cite{khodadadeh2019unsupervised,jang2023unsupervised}: One of the primary areas where meta-learning has demonstrated substantial impact is in few-shot learning. Methods like Model-Agnostic Meta-Learning (MAML) \cite{maml} aim to find a set of model parameters that are sensitive to changes in the task, allowing for quick adaptation to new tasks with just a few examples. Variants of MAML, such as First-Order MAML (FOMAML) and Reptile \cite{reptile}, reduce the computational complexity of the original algorithm while maintaining competitive performance.

Metric-based Approaches: Metric-based meta-learning methods, such as Matching Networks \cite{relationnet} and Prototypical Networks \cite{protonet}, learn an embedding space where similar tasks are closer together. These models perform classification by comparing the distance between new examples and a few labeled instances (support set) in this learned space, achieving remarkable results in few-shot classification tasks.

Memory-augmented Networks: Another line of research in meta-learning explores the use of external memory structures to facilitate rapid adaptation. Santoro et al introduced Memory-Augmented Neural Networks (MANNs) \cite{mann} that use an external memory to store and retrieve information about past tasks, enabling the model to perform well even in tasks with highly variable distributions.

Gradient-based Meta-learning: Beyond MAML, other gradient-based methods such as Meta-SGD \cite{Meta-sgd} and Learning to Learn with Gradient Descent have been proposed. These methods modify the way gradients are used during the training of the model, either by learning the initial parameters (as in MAML) or by learning the learning rates for different parameters, allowing for more efficient adaptation.

Bayesian Meta-learning: Bayesian approaches to meta-learning, such as Bayesian MAML \cite{cnap}, offer a probabilistic framework for capturing uncertainty and improving generalization to new tasks. These methods have been particularly useful in scenarios where task distributions are diverse, and the model needs to account for uncertainty in task inference.

Meta-learning for Reinforcement Learning: Meta-learning has also been successfully applied in the domain of reinforcement learning (RL). Methods such as Meta-RL \cite{yu2020meta} aim to train agents that can quickly adapt to new environments by leveraging the experience gained in previous tasks. These approaches have shown promise in enabling RL agents to solve tasks with minimal exploration, a crucial aspect for real-world applications where exploration can be costly or risky.

In summary, meta-learning has rapidly evolved as a versatile framework that enhances the ability of models to adapt quickly to new tasks, and operate efficiently in dynamic environments. Compared meta-learning with the proposed GeSSL, we can see that the main difference between them located in the way to model discriminability and generalizability. For more details, please refer to the last paragraph of 
Section \ref{sec:4.3}.

\subsection{Broader Impacts and Limitations}
In this subsection, we briefly illustrate the broader impacts and limitations of this work.

\paragraph{Broader Impacts} This work advances SSL by explicitly modeling ``universality'', which refers to the capacity of representations to discriminate, generalize, and transfer, through a unified bi-level optimization that balances task-specific adaptation with cross-task consistency. By deriving a theoretical generalization bound, we provide formal guarantees that GeSSL’s learned features will perform robustly on unseen tasks. Empirically, GeSSL delivers SOTA results across diverse benchmarks, demonstrating its advantages across various settings. This work benefits the field of SSL and machine learning, and also opens up exciting new avenues for future research.

\paragraph{Limitations} This work includes analyses under a variety of settings and presents extensive empirical evidence of its effectiveness. However, it does not offer a dedicated examination of multi-modal scenarios, despite demonstrating the proposed GeSSL’s performance on different modalities, such as images and text. We will investigate additional case studies to extend this work in the future.

\end{document}